\documentclass[twoside,11pt]{article} 
\usepackage{isomath}

\usepackage{jmlr2e}

\usepackage{enumerate}
\usepackage{amsmath}
\usepackage{amssymb}
\usepackage{amsbsy}
\usepackage{dsfont}
\usepackage{theorem}
\usepackage[ruled,vlined]{algorithm2e}
\usepackage{algorithmic}
\usepackage{mathrsfs}
\usepackage{paralist}
\usepackage{epsfig}
\usepackage{makeidx}
\usepackage{xcolor}
\usepackage{color}
\usepackage{times}
\usepackage{tikz}

\usepackage{textcomp}
\usepackage{float}
\usepackage{pgfplots}
\usetikzlibrary{pgfplots.groupplots}
\usepgfplotslibrary{patchplots,colormaps}

\newcommand \E {\mathop{\mbox{\ensuremath{\mathbb{E}}}}\nolimits}

\newcommand\Reals {{\mathds{R}}}

\newcommand \CA {{\mathcal{A}}}

\newcommand \CD {{\mathcal{D}}}

\newcommand \CF {{\mathcal{F}}}

\newcommand \CM {{\mathcal{M}}}

\newcommand \CR {{\mathcal{R}}}
\newcommand \CS {{\mathcal{S}}}

\newcommand \defn {\mathrel{\triangleq}}

\newcommand \argmax{\mathop{\rm arg\,max}}
\newcommand \argmin{\mathop{\rm arg\,min}}

\DeclareMathAlphabet{\mathpzc}{OT1}{pzc}{m}{it}

\newcommand \Normal {\mathop{\mathpzc{N}}\nolimits}

\newcommand \pol {\pi}

\newcommand \mdp {\mu}

\newcommand \eye {\matrixsym{I}}

\newcommand \vs {\vectorsym{s}}

\newcommand \vm {\vectorsym{m}}
\newcommand \vb {\vectorsym{b}}

\newcommand \basis {\phi}

\newcommand \disc {\gamma}

\newcommand \param {\vectorsym{\theta}}

\newcommand \Data {\CD}

\def\clap#1{\hbox to 0pt{\hss#1\hss}}

\usepackage{xcolor}

\usepackage{graphicx}
\usepackage{caption}
\usepackage{subcaption}
\usepackage{enumitem}

\SetKwInput{Initialization}{Initialization}

\ShortHeadings{Randomised Bayesian Least-Squares Policy Iteration}{Tziortziotis, Dimitrakakis, and Vazirgiannis}
\firstpageno{1}

\ewrlheading{14}{2018}{October 2018, Lille, France}{}

\begin{document}

\title{Randomised Bayesian Least-Squares Policy Iteration}

\author{\name Nikolaos Tziortziotis \email ntziorzi@gmail.com \\
       \addr Computer Science Laboratory (LIX)\\
       \'{E}cole Polytechnique, France
       \AND
       \name Christos Dimitrakakis \email christos.dimitrakakis@gmail.com\\
       \addr Department of Informatics\\
       University of Oslo, Norway
       \AND
       \name Michalis Vazirgiannis \email mvazirg@lix.polytechnique.fr \\
       \addr LIX, \'{E}cole Polytechnique, France \\
       \addr Athens University of Economics and Business, Greece}


\maketitle

\begin{abstract}%
  We introduce Bayesian least-squares policy iteration (BLSPI), an off-policy, model-free, policy iteration algorithm that uses the Bayesian least-squares temporal-difference (BLSTD) learning algorithm to evaluate policies. An online variant of BLSPI has been also proposed, called randomised BLSPI (RBLSPI), that improves its policy based on an incomplete policy evaluation step. In online setting, the exploration-exploitation dilemma should be addressed as we try to discover the optimal policy by using samples collected by ourselves. RBLSPI exploits the advantage of BLSTD to quantify our uncertainty about the value function. Inspired by Thompson sampling, RBLSPI first samples a value function from a posterior distribution over value functions, and then selects actions based on the sampled value function. The effectiveness and the exploration abilities of RBLSPI are demonstrated experimentally in several environments.
\end{abstract}
\begin{keywords}
  Reinforcement learning, Bayesian reinforcement learning, Bayesian least-squares temporal-difference, Bayesian least-squares policy-iteration, exploration
\end{keywords}

\section{Introduction}
\label{sec:introduction}

In artificial intelligence (AI), our primary objective is to design intelligent agents able to discover autonomously optimal policies (behaviors) by interacting with their (usually unknown) environment. The reinforcement learning (RL) problem \citep{Sutton+Barto:1998} is a special case of this general setting. Least-squares policy iteration (LSPI) \citep{lagoudakis2003least} is a model-free RL algorithm that is known for its efficient use of training samples and has succeeded in many challenging control problems. It belongs to the family of policy iteration algorithms, using a variant of the least-squares temporal difference (LSTD) algorithm \citep{bradtke1996linear, boyan2002technical} for policy evaluation.

In its original form, LSPI is an offline algorithm that is based on a batch of samples provided beforehand and have been collected through the interaction of the decision maker with his environment. Nevertheless, most of the problems that encountered on real world are online. An online version of LSPI has been proposed by \citet{Busoniu10OnlineLSPI}, that performs policy improvements once every few transitions have been completed, and updates its policy based on an \emph{incomplete} evaluation of its current policy. Efficient exploration is one of the main challenges in online learning, as at each time step we have to decide if we will exploit our current knowledge (estimation), or we will explore our world trying to gain more information about poorly understood/visited states and actions. This is well-known as the \emph{exploration-exploitation dilemma}. In online LSPI, authors have adopted the simple $\epsilon-$greedy exploration strategy \citep{Sutton+Barto:1998}. Before \citet{Busoniu10OnlineLSPI}, \citet{Li:2009:OEL:1558109.1558113} have introduced \textsc{Rmax}-LSPI algorithm that incorporates the \textsc{Rmax} exploration technique \citep{Brafman:2003:RGP:944919.944928} into LSPI.  In contrast to online LSPI that discards the collected transition samples, \textsc{Rmax}-LSPI algorithm updates its policy based on the whole set of visited sample transitions (LSPI is fully executed at the end of an episode). 


In this work, we propose the Bayesian LSPI (BLSPI) algorithm that constitutes a Bayesian version of the the standard LSPI algorithm. Instead of the LSTD algorithm used by LSPI for policy evaluation, BLSPI uses the Bayesian LSTD (BLSTD) algorithm proposed recently by \citet{Tziortziotis:BLSTD17}.
BLSTD adopts a probabilistic model for the empirical Bellman operator and introduces a prior distribution over the model's parameters. 
This gives us the advantage of probabilistic predictions that quantify our uncertainty over the estimated value function after each policy evaluation step.
We also extend the BLSPI algorithm in the online setting by introducing the Randomised Bayesian LSPI (RBLSPI) algorithm. Like online LSPI, RBLSPI updates its policies based on incomplete policy evaluations. To address the exploration problem, RBLSPI exploits the ability of BLSTD to capture  our uncertainty of the value estimates by using Bayesian inference. Instead of selecting random actions, RBLSPI explores its environment by sampling randomly value functions from a posterior distribution over value functions.
More specifically, RBLSPI samples a value function after each policy improvement step, that defines the \emph{behavioral policy} of the agent. Thereafter, the \emph{behavioral policy} is followed greedily up to the next policy improvement step that is performed once every few state transition samples.


This kind of exploration is based on the simple idea of Thompson sampling \citep{thompson1933lou} that has been been shown to perform very well in Bayesian reinforcement learning \citep{strens2000bayesian,Ghavamzadeh:2015:BRL}. In model-based Bayesian RL \citep{NIPS2013_5185,ijcai:lbrl,Tziortziotis:2014:CTB}, the agent starts by considering a prior belief over the unknown environment model. Then, a model from the posterior distribution is randomly sampled, and an optimal policy is calculated w.r.t. the sampled model. This policy is greedily followed thereafter, with the observed samples to be used to update the posterior distribution over models. Recently, \citet{pmlr-v48-osband16} introduced the idea of randomised value functions by proposing the randomised least-squares value iteration (RLSVI) algorithm. RLSVI uses an exploration method that is similar to the one of RBLSPI: it estimates a Bayesian linear regression model for the value function at the end of each episode, from the observations made up to this point. This mean that we need to keep in memory the transitions observed through the time, in order to estimate value functions at each episode by building a ``new'' Bayesian regression model. Then, a sample from the posterior of the bayesian model over value functions is obtained, and the greedy policy with respect to it is followed at the next episode. Due to our different modelling choices, RBLSPI can use all of the previous episodes data history without the need to keep the observed transitions in memory, and is so more data efficient.
The idea of randomised value functions has been also adopted to deep RL showing that it is an efficient exploration strategy \citep{osband0WR17,2018Touati,Azizzadenesheli18}.

\section{Preliminaries}
\label{sec:preliminaries}

We formulate the underlying control problem as a discrete-time $\gamma$-discounted \emph{Markov decision process} (MDP), $\mu \in \CM$, defined by the tuple $\{\CS, \CA, P, \CR, \disc\}$,  where $\CS$ is the set of states; $\CA$ is the set of available actions; $P(\cdot| \vs, a)$ is a transition kernel, specifying the probability of transition from state $\vs \in \CS$ to next states by taking action $\vectorsym{a} \in \CA$; $\CR: \CS \times \CA \rightarrow \Reals$ is a reward function and $\disc \in [0,1)$ is a constant discount factor. 

We assume that the agent selects its actions based on a  deterministic policy, $\pol: \CS \rightarrow \CA$, which is a mapping from states to actions; $\pol(\vs) \in \CA$ denotes the action returned by policy $\pol$ at state $\vs$.
The utility of the agent is the discounted sum of future reward, $U \defn \sum_{t=0}^{\infty} \disc^{t} r_t$, where $r_t = \CR(\vs_t, a_t)$ is the reward received after executing action $a_t$ at state $\vs_t$.
Given MDP $\mdp$, the agent's objective is to discover an optimal policy $\pol^{*}$ that maximises its expected utility for each possible state $\vs$: $ V^{\pol}(\vs) \defn \E_\mdp^\pol \left[ U | \vs_0 = \vs \right]$,
where the expectation is getting w.r.t. the agent's policy $\pol$ and environment $\mdp$.
In the control problem and especially in the case where the model of the environment is unknown, it is more preferable to consider the action-value function, $Q^{\pol}: \CS \times \CA \rightarrow \Reals$, which given a policy $\pol$ indicates the expected return obtained by executing action $a$ at state $\vs$, and following the policy $\pol$ thereafter: $ Q^{\pol}(\vs,a) \defn \E_\mdp^{\pol}  \left[ U | \vs_0 = \vs, a_0 = a\right]$.

It is well-known that the action-value function of a policy $\pol$ is the unique fixed-point of the \emph{Bellman operator} \citep{Puterman:MDP:1994}, i.e. $Q^{\pol} = T^{\pol} Q^{\pol}$, with the operator $T^{\pol}: (\CS \times \CA) \rightarrow (\CS \times \CA)$ to be defined as:
\begin{equation}
  T^{\pol}Q(\vs,a) \defn r(\vs,a) + \disc \int_{\CS}  Q(\vs',\pol(\vs')) dP(\vs' | \vs, a),
  \label{Eq:BellmanOperator}
\end{equation}
or in a vector form as:
 $ T^{\pol}Q \defn \CR + \disc P^{\pol} Q,$
where $\CR \in \Reals^{|\CS||\CA|}$ is the reward vector, and $P^{\pol} \in \Reals^{|\CS||\CA| \times |\CS||\CA|}$ is the transition matrix induced by the selection of an action and policy $\pol$ right after.

The optimal action-value function is defined as $Q^{*}(\vs,a) \defn \sup_{\pol} Q^{\pol}(\vs,a)$, for each $(s,a) \in \CS \times \CA$.
Given $Q$, a policy is called greedy when $\pol(\vs) = \argmax_{a\in\CA} Q(\vs,a),  \forall \vs \in \CS$.
The greedy policy w.r.t. the optimal action-value function $Q^*$ is optimal, and is denoted by $\pol^*$.
Thus, we need to determine the optimal action-value function in order to find an optimal policy.


\paragraph{Policy iteration \citep{howard:dp}} Policy iteration  is an iterative procedure able to discover the optimal solution for a given MDP.
By starting from an arbitrary policy $\pol_0$ (i.e., randomly selected), it generates a sequence of monotonically improving policies along with their corresponding approximate action-value functions: $Q^{\pol_0} \rightarrow \pol_1 \rightarrow Q^{\pol_1} \rightarrow \pol_2 \rightarrow \ldots$.
Each iteration $k$ consists of two successive steps: \emph{policy evaluation} and \emph{policy improvement}.
In \emph{policy evaluation} step the action-value function $Q^{\pol_k}$ of current policy $\pol_k$ is computed by solving the linear system of the Bellman equations: $Q^{\pol} = (\eye - \disc P^\pol)^{-1} \CR$.
Afterwards, \emph{policy improvement} step generates an improved greedy policy with respect to the action-value function $Q^{\pol_k}$, i.e., $\pol_{k+1} = \argmax_{a\in\CA} Q^{\pol_k}(\vs,a)$.
The whole procedure terminates when no further improvements is possible (i.e., $\pol_k = \pol_{k-1}$), with the policy iteration algorithm to have converge to an optimal policy, $\pol^* = \pol_k$.
In general, policy iteration converges after a small number of iterations.

\paragraph{Approximate policy iteration (API)} Despite its merits, policy iteration requires the exact computation and representation of the action-value function.
Nevertheless, in general the state and/or the action spaces is large or infinite (e.g., continuous spaces), making the explicit representation of the action-value function infeasible.
To tackle this problem, a function approximation scheme is usually employed in the policy evaluation step.
This kind of policy iteration is widely known as \emph{approximate policy iteration} (API) (see \citet{bertsekas2011approximate} for a survey).
In RL, it is common to approximate the action-value function $Q$ by considering linear approximation architectures, i.e. a linear combination of basis functions: $Q^{\pol}_{\param}(\vs,a) = \vectorsym{\basis}(\vs,a)^{\top} \param$,
where $\param \in\Reals^{k}$ is a parameter vector and  $\vectorsym{\phi}: \CS \times \CA \rightarrow \Reals^{k}$ is a function that maps state-action pairs to a feature vector of k components, $\vectorsym{\phi}(\cdot) = \left(\phi_1(\cdot),\ldots,\phi_k(\cdot)\right)^{\top}$.
In that way, the number of parameters ($k << |\CS||\CA| $) that need to be estimated is much less compared to these required in the case of an exact representation.
We also denote by $\CF$ the linear function space spanned by the features $\phi_i$, i.e., $ \CF = \{f_{\param} | f_{\param}(\cdot)  = \vectorsym{\phi}(\cdot)^{\top} \param \}$.
Actually, $\CF$ contains all the action-value functions that can be represented by the features. 

\subsection*{Least Squares Policy Iteration}

One of the most well-known approximate policy iteration algorithms is that of least-squares policy iteration (LSPI) \citep{lagoudakis2003least}.
It is a model-free, off-policy algorithm that uses the least-squares temporal difference (LSTD) \citep{bradtke1996linear} algorithm at the policy evaluation phase.
In practice, LSTD returns the parameters vector $\param$ that minimises the \emph{mean-squared projected bellman error} (MSPBE): 
\begin{equation*}
  \text{MSPBE}(\param) \defn \| Q^{\pol}_{\param} - \Pi T^{\pol} Q^{\pol}_{\param}\|_{\Xi}^2, \label{Eq:MSPBE}
\end{equation*}
where $\Xi$ is a diagonal matrix whose diagonal elements indicate the distribution over $\CS \times \CA$, and $\Pi$ is the projection operator onto feature space $\CF$.
Actually, operator $\Pi$ projects any value function $\vectorsym{u}$ to its nearest value function onto $\CF$, such that $\Pi \vectorsym{u} = Q^{\pol}_{\param}$ where the corresponding parameters are the solution on the next least-squares problem: $\param = \argmin_{\param} \| Q^{\pol}_{\param} - \vectorsym{u}\|^2_{\Xi}$ \citep{sutton2009}.
As the parameterisation is linear, we can show that the projection operator is linear and independent of the parameters $\param$ and given by: $ \Pi = \Phi C^{-1}\Phi^{\top}\Xi$,
where $\Phi \in \Reals^{|\CS||\CA|\times k}$ is a matrix whose rows contain the feature vector $\vectorsym{\phi}(\vs,a)^{\top}, \forall (\vs,a) \in \CS \times \CA$ and $C = \Phi^{\top}\Xi\Phi$ is the Gram matrix.

In practice, the dynamics of the environment $\mdp$ are unknown and the full feature matrices $\Phi$ cannot be formed explicitly in continuous environments. For that purpose, LSTD relies on a batch of transition samples, which is assuming to be  available at our disposal and have been collected through the interaction of the agent with the generative model: $\Data = \{ (\vs_i, a_i, r_i, \vs_i')\}_{i=1}^{N}$, where $\vs_i' \sim P(\vs_i,a_i)$. Given dictionary $\Data$, let us define as $\tilde{\Phi} = [\phi(\vs_1,a_1)^{\top}; \ldots; \phi(\vs_N,a_N)^{\top}]$ and $\tilde{\Phi}' = [\phi(\vs_1',\pol(\vs_1'))^{\top}; \ldots; \phi(\vs_N',\pol(\vs_N'))^{\top}]$ the sampled feature matrices, and as $\CR = [r_i, \ldots, r_N]^{\top}$ the sampled reward vector. Therefore, the empirical MSPBE can be expressed as a standard weighted least-squares problem:
\begin{equation*}
  E_{\Data}(\param) \defn \| A\param - \vb \|_{\tilde{C}^{-1}}^2, \label{Eq:MSPBE1}
\end{equation*}
where $A \defn \tilde{\Phi}^{\top} (\tilde{\Phi} - \disc \tilde{\Phi'})$, $\vectorsym{b} \defn \tilde{\Phi}^{\top} \CR$, \text{and} $\tilde{C} \defn \tilde{\Phi}^{\top}\tilde{\Phi}$.
Minimising the empirical MSPBE we get that the optimal solution is given as:
\begin{equation}
  \param^{*} = A^{-1} \vb.
  \label{eq:lstdsolution}
\end{equation}
It has been shown \citep{bradtke1996linear,lazaric2010finite,Nedic2003} that the LSTD solution $ \tilde{\Phi} \param^{*}$ converges to the fixed-point of $\hat{\Pi} T^{\pol}$ as $N \rightarrow \infty$.
For the rest of the paper, we denote as $\hat{\Pi}$ the sample based feature space projector, called empirical projection.
A variant of the LSTD algorithm that can be considered to have some resemblances to BLSTD \citep{Tziortziotis:BLSTD17} adopted in our work is the slLSTD proposed by \citet{5676598}.
The slLSTD algorithm is a statistical linearisation-based generalisation of LSTD that  allows considering nonlinear parameterisations, i.e., neural networks. 


After having approximated the action-value function $Q_{\param}^{\pol_{k-1}}$ of policy $\pol_{k-1}$ (\emph{policy evaluation} phase), an improved greedy policy $\pol_k$ is generated (\emph{policy improvement} phase). These two steps are repeated successively at each iteration, until we converge at an optimal policy.

\section{Bayesian Least-Squares Policy Iteration}
\label{sec:BLSTD}

In this section, we introduce a Bayesian variant of LSPI algorithm, called BLSPI. Like LSPI, BLSPI is an offline, model-free and off-policy algorithm that belongs to the family of API algorithms and can be applied to control problems. In contrast to LSPI, that uses the standard LSTD algorithm at the policy evaluation step, BLSPI evaluates policies by using a Bayesian version of the LSTD algorithm, called BLSTD \citep{Tziortziotis:BLSTD17}. 
BLSTD has the advantage of probabilistic predictions as it quantifies our uncertainty about the value function instead of having only a point estimate over the unknown value function parameters like in the case of LSTD. 
In this section, we present a modified version of the BLSTD algorithm that learns the approximate action-value function $Q^\pol$ of a given policy $\pol$ instead of the state value function, $V$. In this way, we are able to select actions without the knowledge of the environment model. 

Similarly to \citet{Tziortziotis:BLSTD17}, we consider the \emph{empirical Bellman operator} that is given by the standard Bellman operator \eqref{Eq:BellmanOperator} plus some additive noise $\epsilon =\Normal(0, \beta^{-1})$:
\begin{equation*}
  \hat{T}^{\pol}Q^{\pol}_{\param} = \CR + \disc P^{\pol} Q^{\pol}_{\param} + N, \quad N \sim \Normal(0, \beta^{-1}\eye).
\end{equation*} 
Given a set of transitions $\Data$, BLSTD seeks the action-value function parameters $\param$ that are invariant w.r.t. the composed operator $\hat{\Pi}\hat{T}^{\pol}$:
\begin{equation*}
  Q^{\pol}_{\param} = \hat{\Pi}\hat{T}^{\pol}Q^{\pol}_{\param} \Leftrightarrow  
  \tilde{\Phi}^{\top} \CR = \tilde{\Phi}^{\top}(\tilde{\Phi} - \disc\tilde{\Phi'})\vectorsym{\theta} + \tilde{\Phi}^{\top} N.
\end{equation*}
It allows us to treat the empirical MSPBE as a linear regression model:
\begin{equation*}
  \vb = A \param +  \tilde{\Phi}^{\top} N,
\end{equation*}
with its \textbf{log likelihood function} to be:
\begin{equation*}
  \ln p(\vb| \vectorsym{\theta}, \beta) =  \frac{k}{2}\ln(\beta) - \frac{1}{2} \ln (|\tilde{C}|)  - \frac{k}{2}\ln(2\pi) - \frac{\beta}{2} E_{\Data}(\vectorsym{\theta}).
\end{equation*}
It can be easily verified that by setting the gradient of the log likelihood with respect to model's parameters $\param$ equal to zero, we get the batch LSTD solution \eqref{eq:lstdsolution} (\emph{maximum likelihood} solution). 

A zero-mean isotropic Gaussian conjugate \textbf{prior distribution} over the model parameters $\param$ has been also considered, to model the \emph{parametric} uncertainty \citep{icml/MannorSST04}:
  $p(\param | \alpha) =  \Normal(\param | 0,\alpha^{-1} \eye)$, 
with the log posterior distribution to given as $ \ln p(\param | \Data) \propto - \frac{\beta}{2} E_{\Data}(\param) - \frac{\alpha}{2}\param^{\top}\param$.

Therefore, maximising the posterior distribution w.r.t. $\vectorsym{\theta}$ is equivalent to the minimisation of the MSPBE plus an $\ell_2$-penalty ($\lambda = \alpha / \beta$).
Thus, if we set hyperparameter $\alpha$ to a large value, the total squared length of the parameter vector $\vectorsym{\theta}$ will be encouraged to be small.
Completing the squares of the log of the posterior distribution,
we get that the {\bf posterior distribution} is also Gaussian,
\begin{equation}
  p(\param | \Data) = \Normal\left(\param | \vm \defn \beta S A^{\top} \tilde{C}^{-1} \vb, S  \defn (\alpha \eye + \beta A^{\top} \tilde{C}^{-1} A)^{-1}\right),
  \label{eq:posterior}
\end{equation}
where matrix $\Sigma \defn A^{\top} \tilde{C}^{-1} A$ is always positive definite. 
The {\bf predictive distribution} of the action-value function over a new state-action pair, $(\vs^*, a^*)$, is again Gaussian and it is given by: 
\begin{equation*}
  p(Q^{\pol}_{\param}(\vs^*,a^*) | \vs^*, a^*, \Data) = \Normal(Q^{\pol}_{\param}(\vs^*,a^*) | \vectorsym{\phi}(\vs^*,a^*)^{\top}\vm, \vectorsym{\phi}(\vs^*,a^*)^{\top} S \vectorsym{\phi}(\vs^*,a^*)).
\end{equation*}

Finally, the generic bound of approximate policy iteration \citep{BertsekasTsitsiklis:NDP} holds also for the proposed BLSPI algorithm like in LSPI. Therefore, the performance of the sequence of policies produced by BLSPI is at most a constant multiple of $\delta$ (positive scalar that bounds the policy evaluation errors over all iterations) away from the optimal one.

\section{Randomised Bayesian Least-Squares Policy Iteration}
\label{sec:onlineblspi}

One of the most challenging tasks in reinforcement learning is the development of an agent able to adapt its behavior online by interacting with the environment.
In this section, we introduce an online variant of Bayesian LSPI, called  randomised BLSPI (RBLSPI).
RBLSPI updates the agent's policy once every few transition samples based on the samples collected by itself up to this point.
Similarly to online LSPI introduced by \citet{Busoniu10OnlineLSPI}, RBLSPI carries out policy updates based on an incomplete evaluation of the current policy.
This is a variation of policy iteration named as \emph{optimistic policy iteration} \citep{Tsitsiklis:2003:COP}. 
\begin{algorithm}[t!]
  \small
  \KwIn{Basis $\basis$, $\disc$, $\alpha$, $\beta$, $K$}
  \Initialization{$A \leftarrow \mathbf{0}_{k,k}, \tilde{C} \leftarrow \vectorsym{0}_{k,k}, \vb \leftarrow \mathbf{0}_k$, $\vm \sim \Normal(\mathbf{0}_{k}, \eye)$, $\tilde{\param} = \vm$}
  \DontPrintSemicolon 
  \Begin{
    Observe $\vs_{0}$ \;
    \For{$t=0,\dots$} {
      $a_{t} \in \argmax_{a \in \CA} \basis(\vs_{t},a) \tilde{\param}$ \tcp*{Take action}
      Observe $r_t$, $\vs_{t+1}$ \;
      $A \leftarrow A +\basis(\vs_{t}, a_{t})(\basis(\vs_{t}, a_{t}) - \disc\basis(\vs_{t+1}, \argmax_{a \in \CA} \basis(\vs_{t+1},a) \vm))^\top$\;
      $\tilde{C} \leftarrow \tilde{C} + \basis(\vs_{t}, a_{t})\basis(\vs_{t}, a_{t})^\top$\;
      $\vb \leftarrow \vb + \basis(\vs_{t}, a_{t})r_{t}$\;
      \If{$(t \mod K == 0)$}{
        $S = (\alpha \eye + \beta A^{\top} \tilde{C}^{-1} A)^{-1}$\;
        $\vm = \beta S A^{\top} \tilde{C}^{-1} \vb$\;
        $\tilde{\param} \sim \Normal(\vm, S)$ \tcp*{Sample Gaussian posterior (Eq.\ref{eq:posterior})}
      }  
    }
  }
  \caption{Randomised BLSPI (RBLSPI)}
  \label{alg:RBLSPI}
\end{algorithm}
In contrast to the standard offline BLSPI algorithm, in the case of RBLSPI algorithm two critical parameters should be considered: i) the number $K \geq 1$ of collected transitions between successive policy improvements, and ii) the exploration strategy that should be followed. 
In the case where the parameter $K$ has been selected to be too large, a bad policy is highly possible to be used for a long period that could affect the general performance of the algorithm.
On the other hand, if policy is improved after each transition ($K=1$), the running time of the RBLSPI will be increased making that non applicable in some cases.
An extensive analysis about the impact of the free parameter $K$ in the performance of RBLSPI is presented in appendix. 

The design of an efficient exploration strategy constitutes the core of an online agent that updates its belief based on samples collected by itself.
In practice, the agent has to make sure that the collected samples cover the state and action spaces sufficiently.
In any other case, the value function of the unvisited state-action pairs will be poorly estimated leading to imprecise policy improvements.
One of the simplest exploration strategies is that of $\epsilon-$greedy: the agent selects a random action with probability $\epsilon \in [0,1]$ and the greedy action w.r.t. to its value function otherwise. Nevertheless, it has been shown that it could lead to highly inefficient learning \citep{pmlr-v48-osband16}. 

Motivated by Thompson sampling and taking advantage of the BLSTD algorithm that considers our uncertainty over the estimated action-value function, RBLSPI algorithm explores the environment by  sampling the value function randomly instead of selecting random actions. More specifically, a value function is sampled right after each policy evaluation step and actions are selected greedily based on the sampled value function thereafter. RBLSPI samples a value function by sampling the posterior distribution (given by Eq.~\eqref{eq:posterior}) over model's parameters.
Therefore, RBLSPI tends to explore more non frequently selected actions with a highly uncertain value.
As learning process evolves through time, our knowledge about the environment will be enriched as the number of visited state-action pairs will be increased.
Therefore, our confidence about the estimated value functions will be increased, driving us to act more greedily with the passing of time.
The main advantage of the specific exploration strategy over other strategies, i.e., $\epsilon-$greedy, boltzmann, etc., is its ability to identify if a region has sufficiently explored by acting randomly mainly only on the areas that have not explored adequately yet.
Finally, it should be noted that in RBLSPI the free parameter $K$ has a dual role, as it determines how often we update our policy and select a new exploration policy.
Randomised BLSPI is presented in detail in Algorithm~\ref{alg:RBLSPI}.

\section{Experiments}
\label{sec:experiments}

In this section, we formally present the results about the performance of the online RBLSPI algorithm. 
Through our analysis,  we examine the exploration efficiency of the proposed Randomised Bayesian LSPI (RBLSPI) algorithm in four well-known continuous state, discrete-action, episodic domains.
Due to space limitations, only a part of our empirical results are presented here.
Specifically, we present only results about the mountain car environment and a sparse reward version of it that requires {\it deep exploration}.
The full set of our results along with the experimental set-up is presented in detail in appendix.

In mountain car \citep{Moore90,Sutton+Barto:1998}, our objective is to drive an underpowered car up a steep road to the right hilltop ($p \geq 0.5$) within at most $500$ steps.
The car state in this domain is described by two continuous variables, its position ($p$) and its velocity ($v$).
At each time step, the agent can select between three possible actions: forward, reverse and zero throttle.
In our experiments we have considered two different reward signals:
\begin{itemize}[noitemsep, topsep=0pt]
\item In standard mountain car the immediate reward at each step is equal to $-1$ except in the case where the agent reaches the goal (zero reward). 
\item The immediate reward is equal to $1$ when the episode terminates and zero otherwise \citep{ijcai2018-376}. We call this version sparse Mountain car.
\end{itemize}
In both cases, the car is positioned motionless ($v_0=0$) to a uniformly randomly selected position close to the valley ($-0.6 \leq p_0 \leq -0.4$) at the beginning of each episode.
The discount factor is set to $0.99$. 
An equidistant $8 \times 8$ grid of RBFs over the state space plus a constant term is selected.
This set of basis functions is replicated for each action, giving a total of $195$ RBFs.
\begin{figure*}[t]
  \centering
  \begin{subfigure}[t]{.497\textwidth}
    \begin{tabular}{cc}
      \scalebox{.219}{\includegraphics{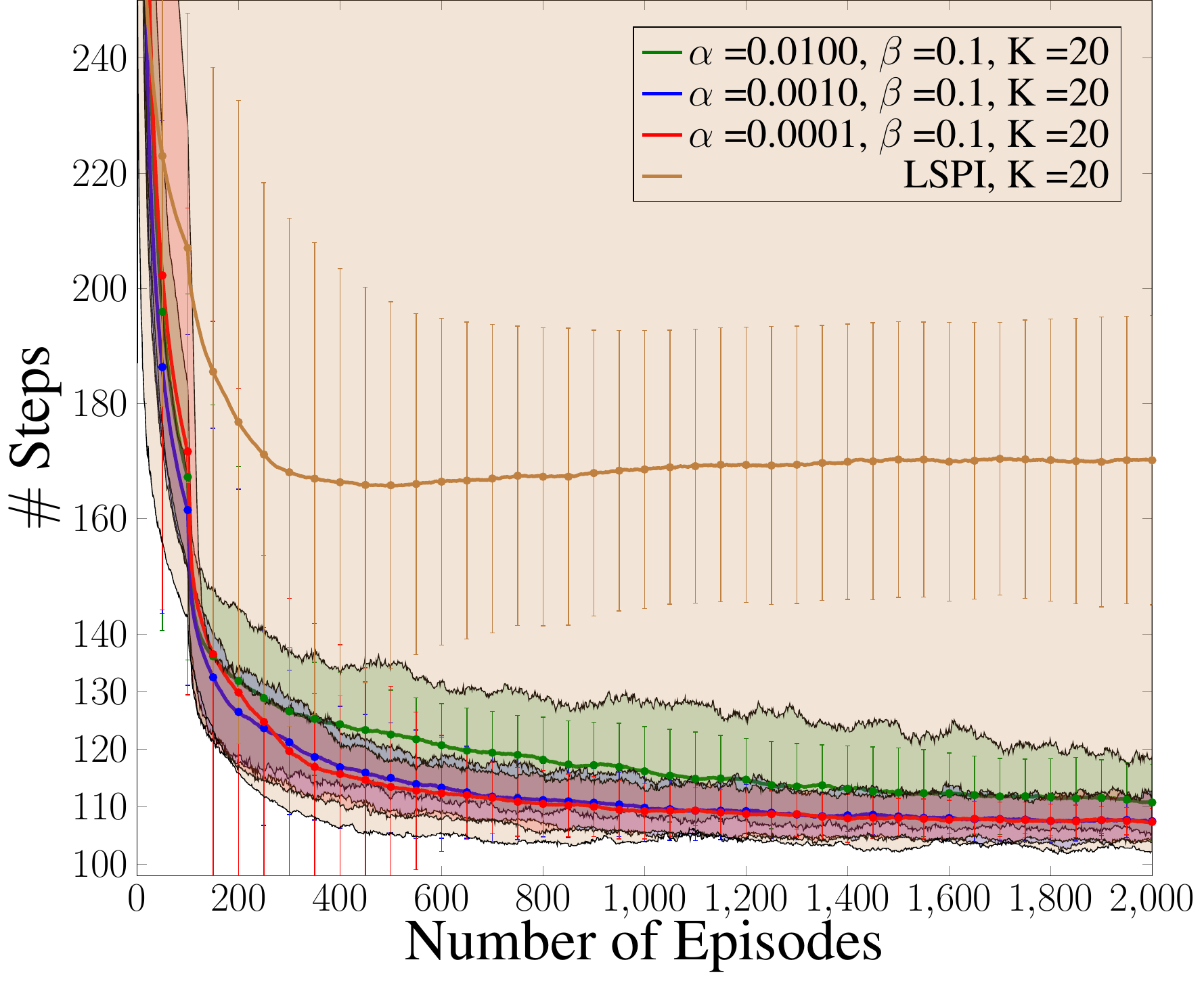}}
      \scalebox{.219}{\includegraphics{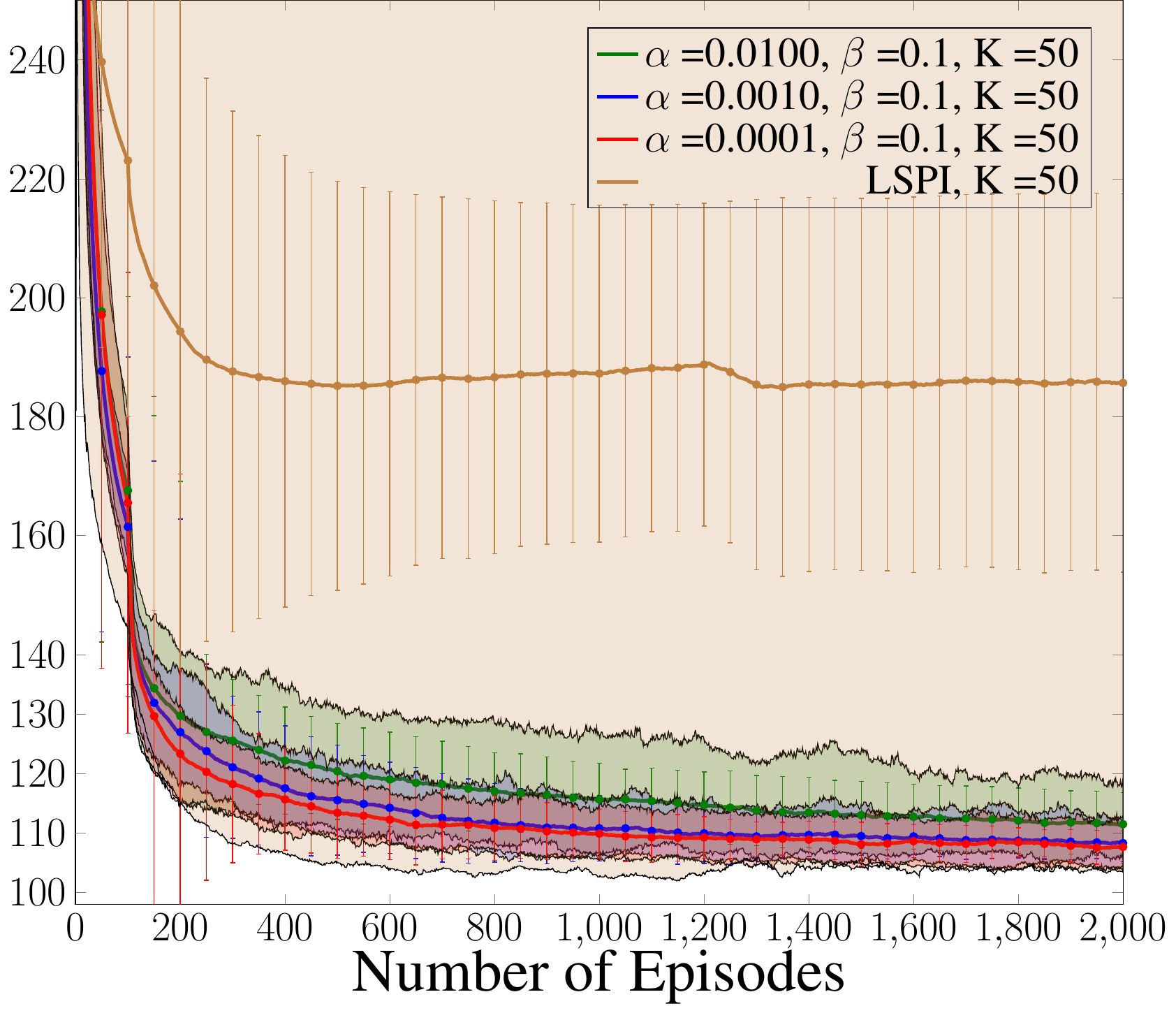}}\\
      \scalebox{.219}{\includegraphics{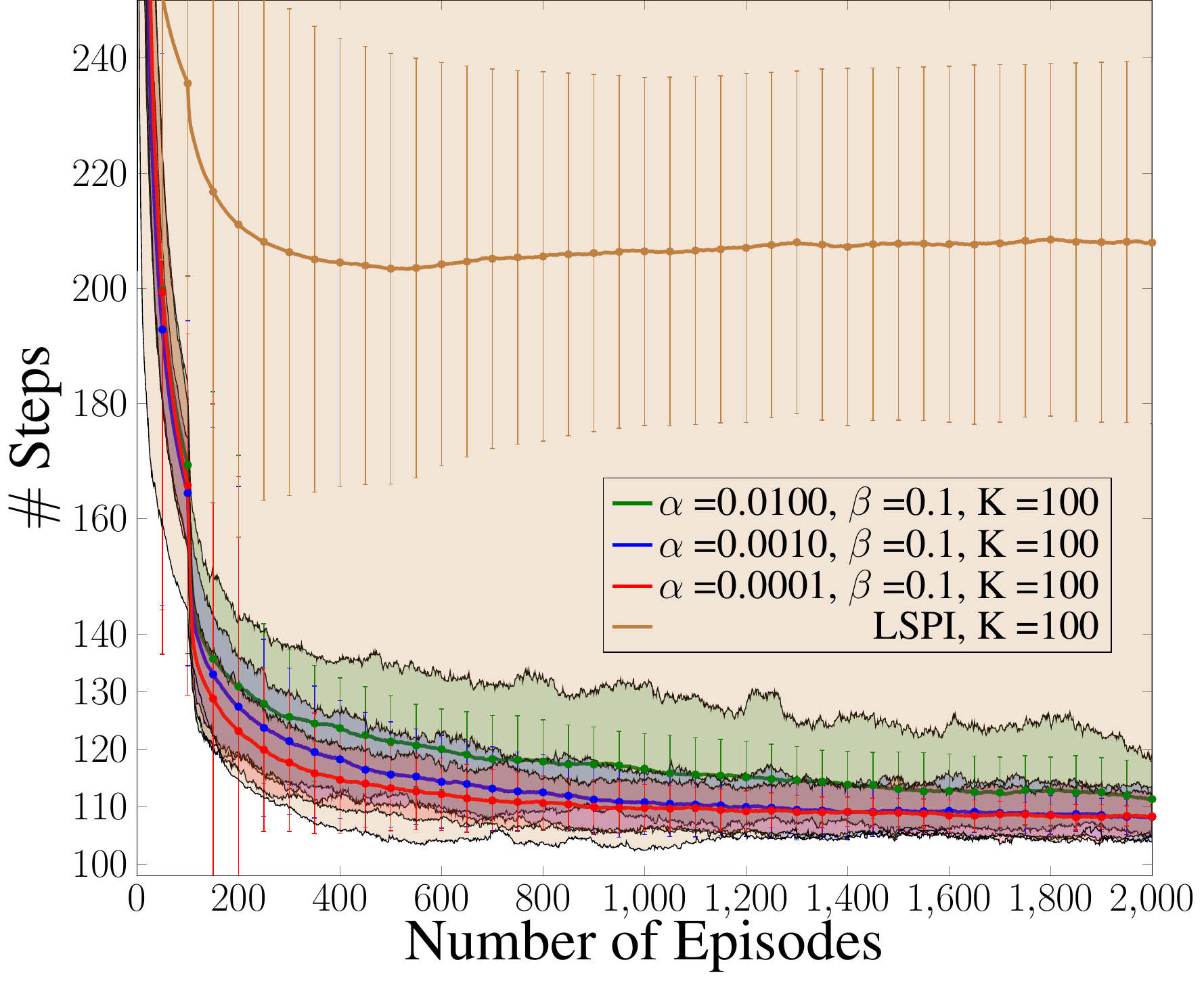}}
      \scalebox{.219}{\includegraphics{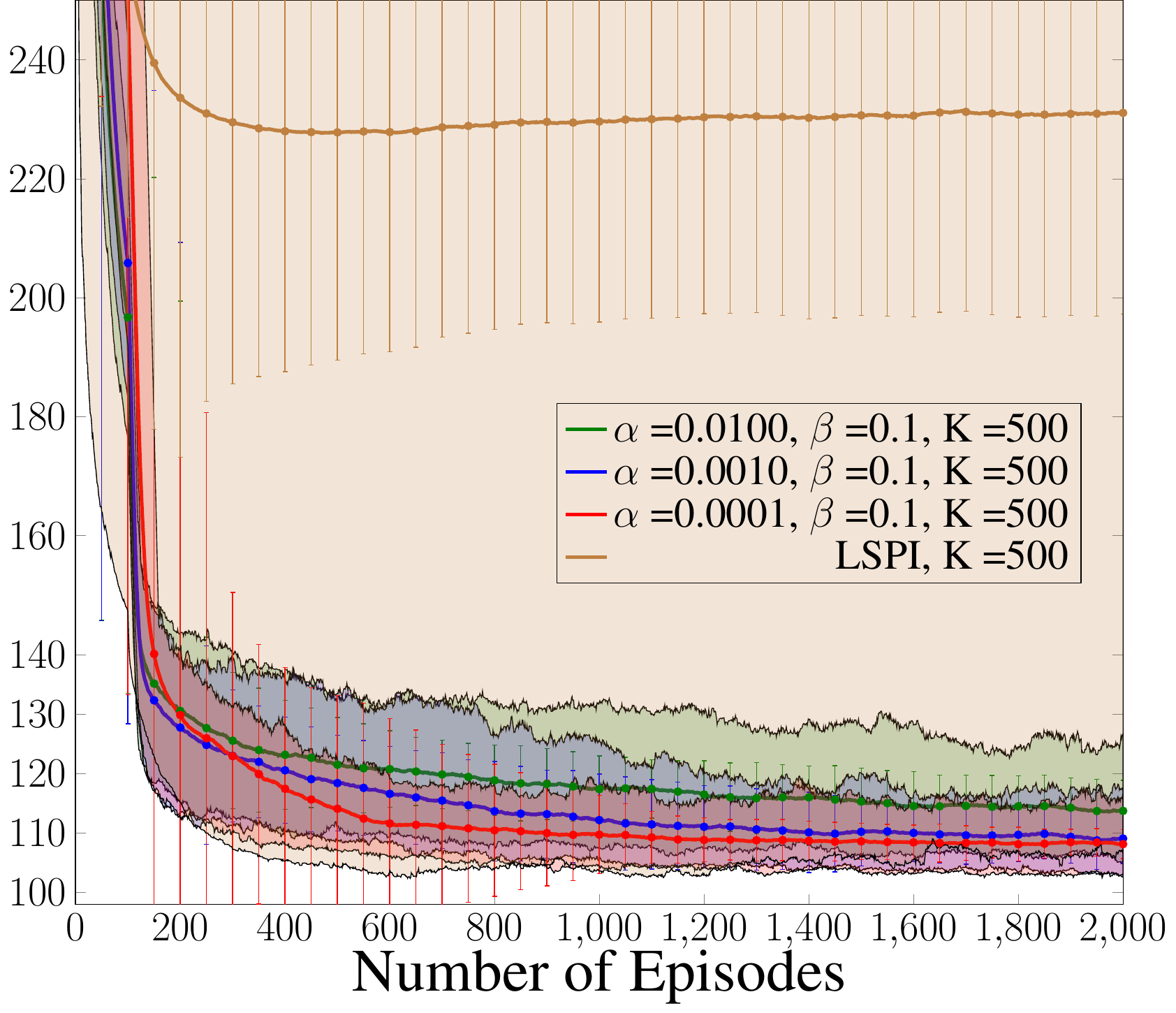}}
    \end{tabular}
    \caption{Mountain Car}
    \label{fig:mountain}
  \end{subfigure}
  \begin{subfigure}[t]{0.497\textwidth}
    \begin{tabular}{cc}
      \scalebox{.219}{\includegraphics{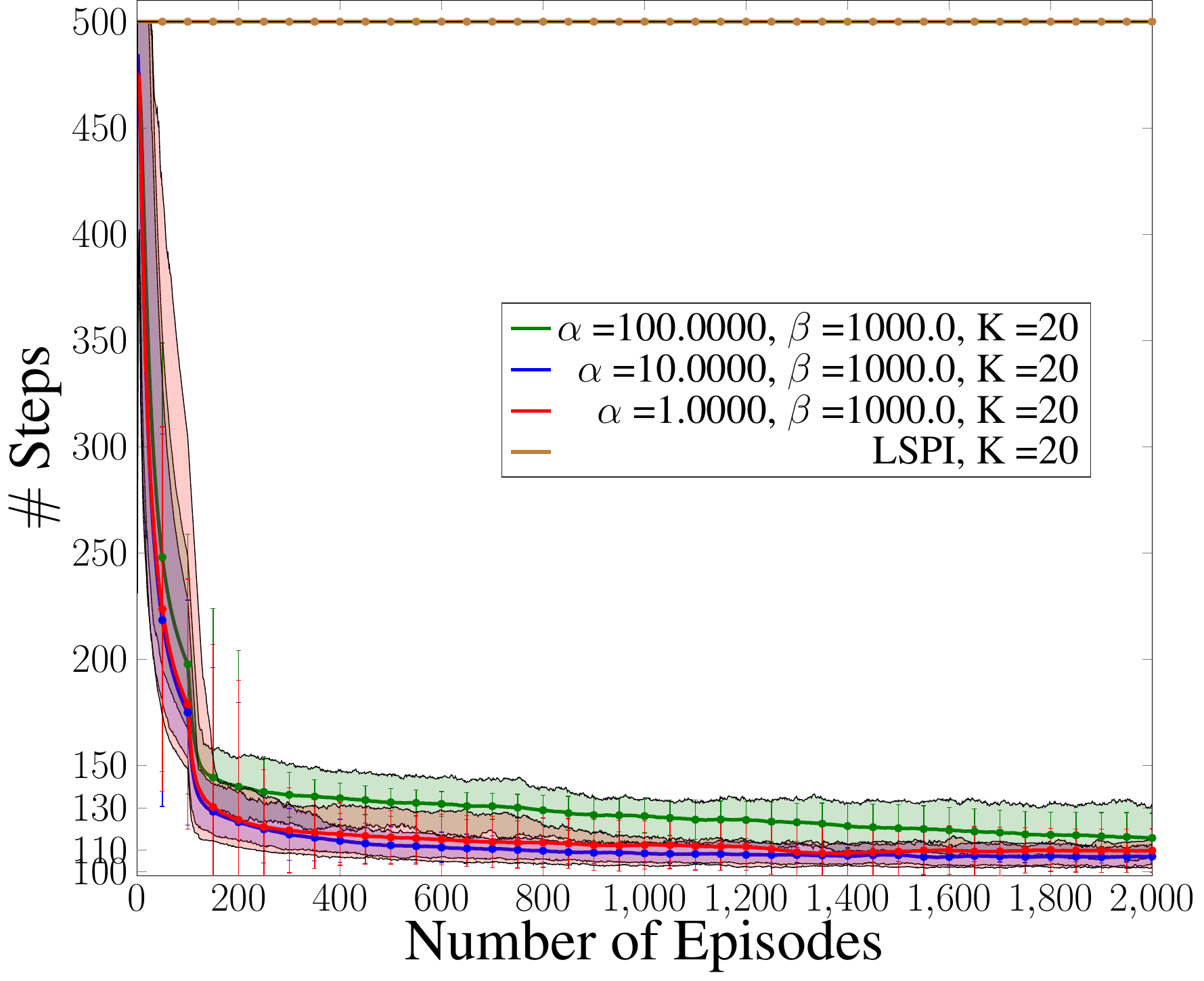}}
      \scalebox{.219}{\includegraphics{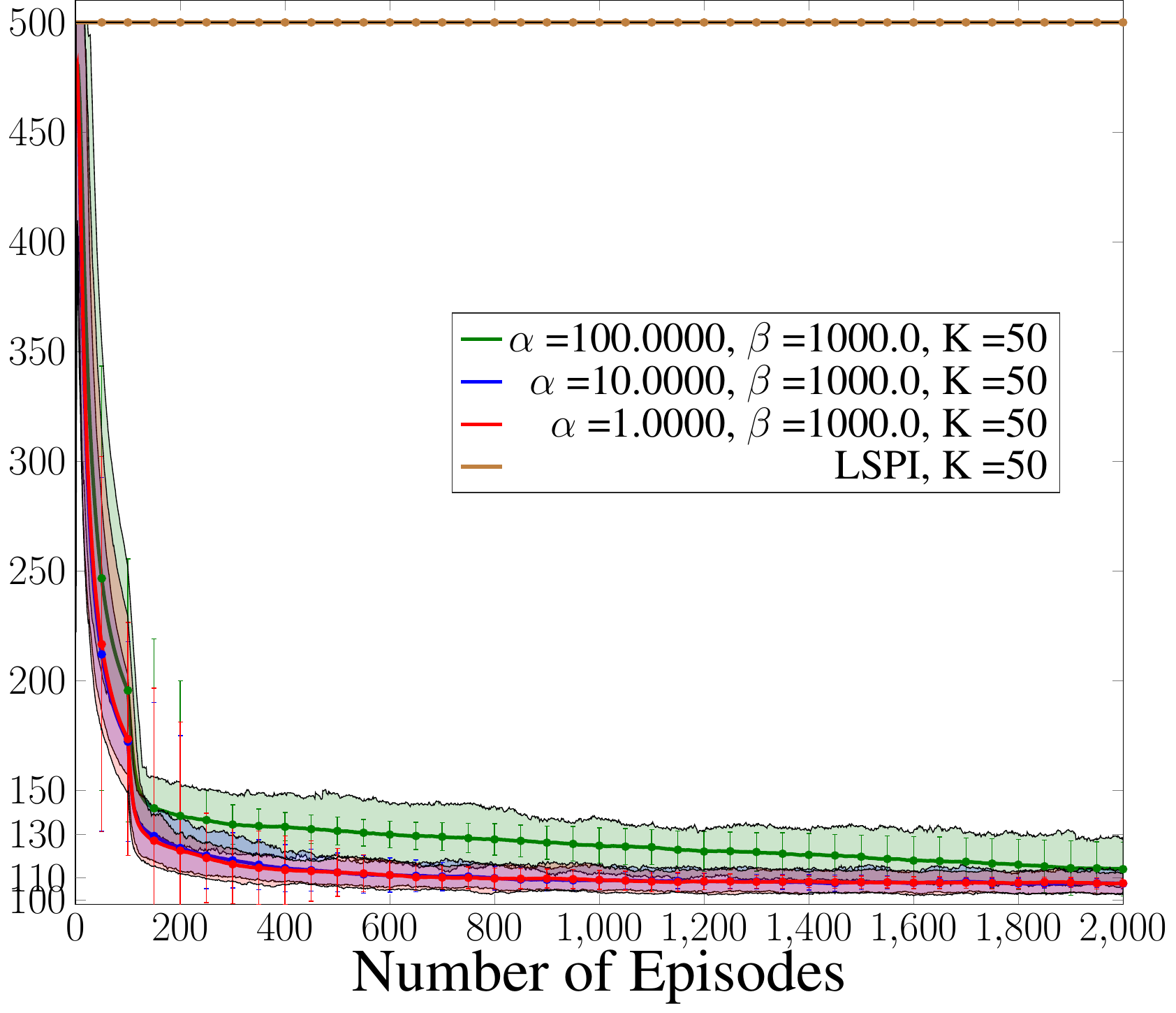}}\\
      \scalebox{.219}{\includegraphics{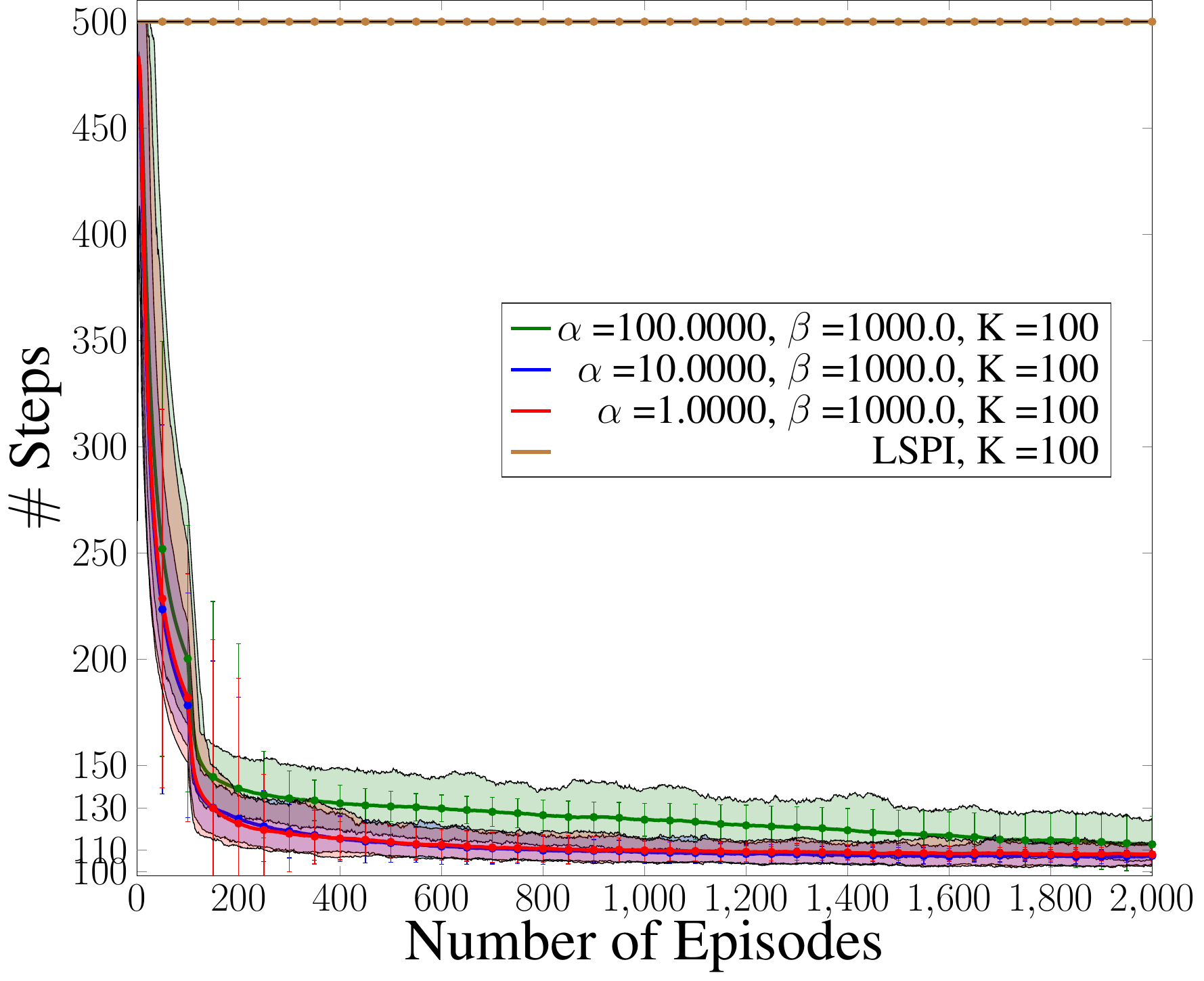}}
      \scalebox{.219}{\includegraphics{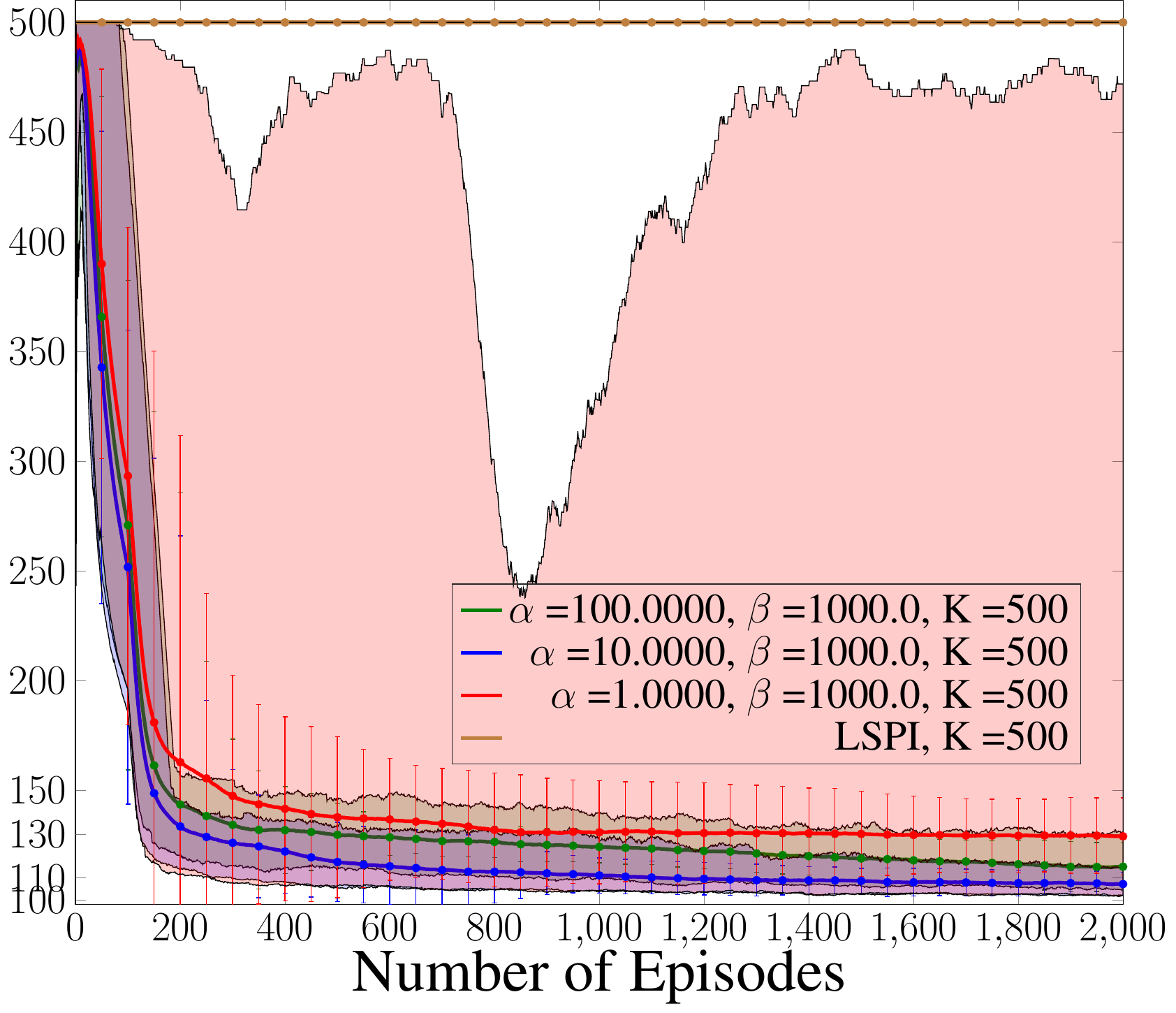}}
    \end{tabular}
    \caption{Sparse Mountain Car}
    \label{fig:mountainsparse}
  \end{subfigure}
  \caption{Performance of RBLSPI and online LSPI on the environments of Mountain Car and Sparse Mountain Car for varying parameter $K$.}
  \vspace{-2.em}
\end{figure*}
In the task of mountain car the usage of an efficient exploration strategy is of high importance due to the fact that the goal cannot be easily reached by the agent.
The necessity of an effective (deep) exploration scheme becomes more crucial in the case of the sparse reward mountain car environment.
The empirical results of the RBLSPI algorithm on the the standard and sparse mountain car environments are presented in Figs~\ref{fig:mountain} and \ref{fig:mountainsparse}, respectively.
Comparisons have been conducted with an onpolicy variant of online LSPI algorithm that uses the simple $\epsilon-$greedy exploration scheme.
In the case of RBLSPI, we start by using a randomly selected policy (see Alg.~\ref{alg:RBLSPI}) at the beginning of each run. 
For each experiment, we report the mean performance (average number of steps over $100$ episodes) across $100$ independent runs.
For each average, we also plot the $95\%$ confidence interval for the accuracy of the mean estimate with error bars. Additionally, we show the $90\%$ percentile region of the runs.
In the case of RBLSPI algorithm we set precision $\beta$ equal to $0.1$ and $1000$ for the mountain car and sparse mountain car, respectively, and vary the precision hyperparameter $\alpha$.

Let us first to examine the impact of free parameter $K$ in the performances of both algorithmic schemes.
It should be reminded that parameter $K$ defines the number of transitions that should be conducted between successive policy improvement steps (see Algorithm~\ref{alg:RBLSPI} for details).
The following values have been examined for $K$: $20$, $50$, $100$ and $500$ (end of episode).
As it was expected, the performance of both algorithms is better when $K$ is set to a low value, i.e., $K=20$.
It should be also mentioned that even if we set $K$ equal to $500$, RBLSPI is able to reach the goal in both environments. 
We have also examined the impact of precision parameter $\alpha$ on the performance of RBLSPI.
Our experiments show that it is more preferable to be set to a lower value if the parameter $K$ is not equal to $500$ (policy updates are executed only at the end of each episode).
Nevertheless, it can be easily verified that RBLSPI achieves to discover good policies independent of the selection of the value of the precision $\alpha$.
Our results also validates our claims about the efficiency of the exploration mechanism of RBLSPI. Actually, RBLSPI outperforms online LSPI in both environments.
More specifically, RBLSPI discovers a near-optimal policy in both environments as our RBLSPI agent is able to reach the goal in less than $110$ steps on average.
On the other hand, the online LSPI stuck in the valley in the case of sparse mountain car, and it doesn't reach the goal in none of the runs.

\section{Conclusion}
\label{sec:conclusion}
In the present work, we have introduced a fully Bayesian version of the widely used least-squares policy iteration, called BLSPI. It is achieved by adopted BLSTD in the policy evaluation step of policy iteration. We further extended BLSPI on an online setting by proposing the RBLSPI algorithm. Taking advantage of the BLSTD algorithm to estimate our uncertainty over the value function estimations, RBLSPI explores efficiently its environment by sampling randomly value functions instead of selecting random actions. 
The efficiency of the proposed exploration strategy has been demonstrated experimentally in four continuous state-space environments, where comparisons have been conducted with the online LSPI algorithm that uses the simple $\epsilon$-greedy exploration strategy.

\bibliography{misc}

\newpage

\section*{Appendix: Empirical Analysis}

In this section, we present the full set of our empirical results about the two proposed RL algorithmic schemes introduced in our manuscript, BLSPI (Sec.~\ref{sec:BLSTD}) and RBLSPI (Sec.~\ref{sec:onlineblspi}).
Actually, our empirical analysis is divided in two main parts:
\begin{itemize}[noitemsep]
\item First, we examine the ability of the Bayesian LSPI algorithm to discover the same (or a close) policy returned by the original LSPI algorithm. For this purpose, we consider the simple chain walk domain proposed by \citet{lagoudakis2003least}. 
\item Second, we examine the exploration efficiency of the proposed Randomised Bayesian LSPI (RBLSPI) algorithm on four well-known continuous state, discrete-action, episodic domains. Comparisons have been conducted with the online LSPI algorithm that follows the simple $\epsilon-$greedy strategy \citep{Busoniu10OnlineLSPI}.
\end{itemize}

\subsection*{Bayesian LSPI Performance}
\label{sec:blspi_performance}


As aforementioned, the simple chain walk \citep{lagoudakis2003least} environment has been considered to examine how close the policies returned by the Bayesian LSPI algorithm are to those returned by the vanilla LSPI algorithm.
This is a discrete $20-$state chain\footnote{In chain walk domain, we have used the LSPI code which is freely available at: \url{https://www2.cs.duke.edu/research/AI/LSPI/}.} with its boundaries to be dead-ends.
There are two noisy available actions,  i.e., ``left'' (L) and ``right'' (R).
The probability of an action to fail is equal to $0.1$, moving the agent in the opposite direction. 
A reward of $+1$ is given only at the boundaries of the chain (states $1$ and $20$). 
The discount factor is set to $0.9$.
Similarly to \citet{lagoudakis2003least}, we have used a polynomial of degree $4$ for approximating the value function for each one of the two actions ($10$ basis functions in total have been used).
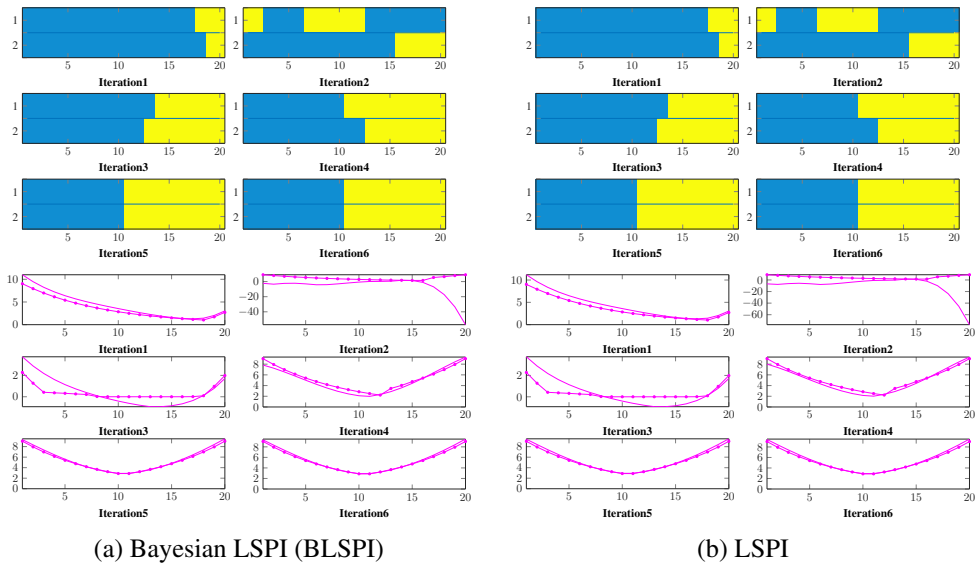
\begin{figure}[h]
  \begin{center}
    \begin{tabular}[c]{cc} 
      \hspace{-3.5mm}
      \scalebox{.4}{
%
%
%
\definecolor{mycolor1}{rgb}{0,0.447,0.741}%
\begin{tikzpicture}

\begin{axis}[%
width=2.64938446969697in,
height=0.653655181751278in,
axis on top,
scale only axis,
separate axis lines,
every outer x axis line/.append style={darkgray!60!black},
every x tick label/.append style={font=\color{darkgray!60!black}},
xmin=0.5,
xmax=20.5,
xlabel={\bf Iteration3},
every outer y axis line/.append style={darkgray!60!black},
every y tick label/.append style={font=\color{darkgray!60!black}},
y dir=reverse,
ymin=0.5,
ymax=2.5,
yticklabels={,,1,,2},
name=plot3
]
\addplot graphics [xmin=0.5,xmax=20.5,ymin=0.5,ymax=2.5] {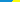};
\addplot [
color=mycolor1,
solid,
forget plot
]
table[row sep=crcr]{
1 1.5\\
20 1.5\\
};
\end{axis}

\begin{axis}[%
width=2.64938446969697in,
height=0.653655181751278in,
axis on top,
scale only axis,
separate axis lines,
every outer x axis line/.append style={darkgray!60!black},
every x tick label/.append style={font=\color{darkgray!60!black}},
xmin=0.5,
xmax=20.5,
xlabel={\bf Iteration1},
every outer y axis line/.append style={darkgray!60!black},
every y tick label/.append style={font=\color{darkgray!60!black}},
y dir=reverse,
ymin=0.5,
ymax=2.5,
name=plot1,
yticklabels={,,1,,2},
at=(plot3.above north west),
anchor=below south west
]
\addplot graphics [xmin=0.5,xmax=20.5,ymin=0.5,ymax=2.5] {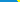};
\addplot [
color=mycolor1,
solid,
forget plot
]
table[row sep=crcr]{
1 1.5\\
20 1.5\\
};
\end{axis}

\begin{axis}[%
width=2.64938446969697in,
height=0.653655181751278in,
axis on top,
scale only axis,
separate axis lines,
every outer x axis line/.append style={darkgray!60!black},
every x tick label/.append style={font=\color{darkgray!60!black}},
xmin=0.5,
xmax=20.5,
xlabel={\bf Iteration2},
every outer y axis line/.append style={darkgray!60!black},
every y tick label/.append style={font=\color{darkgray!60!black}},
y dir=reverse,
ymin=0.5,
ymax=2.5,
name=plot2,
yticklabels={,,1,,2},
at=(plot1.right of south east),
anchor=left of south west
]
\addplot graphics [xmin=0.5,xmax=20.5,ymin=0.5,ymax=2.5] {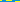};
\addplot [
color=mycolor1,
solid,
forget plot
]
table[row sep=crcr]{
1 1.5\\
20 1.5\\
};
\end{axis}

\begin{axis}[%
width=2.64938446969697in,
height=0.653655181751278in,
axis on top,
scale only axis,
separate axis lines,
every outer x axis line/.append style={darkgray!60!black},
every x tick label/.append style={font=\color{darkgray!60!black}},
xmin=0.5,
xmax=20.5,
xlabel={\bf Iteration4},
every outer y axis line/.append style={darkgray!60!black},
every y tick label/.append style={font=\color{darkgray!60!black}},
y dir=reverse,
ymin=0.5,
ymax=2.5,
name=plot4,
yticklabels={,,1,,2},
at=(plot2.below south west),
anchor=above north west
]
\addplot graphics [xmin=0.5,xmax=20.5,ymin=0.5,ymax=2.5] {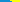};
\addplot [
color=mycolor1,
solid,
forget plot
]
table[row sep=crcr]{
1 1.5\\
20 1.5\\
};
\end{axis}

\begin{axis}[%
width=2.64938446969697in,
height=0.653655181751278in,
axis on top,
scale only axis,
separate axis lines,
every outer x axis line/.append style={darkgray!60!black},
every x tick label/.append style={font=\color{darkgray!60!black}},
xmin=0.5,
xmax=20.5,
xlabel={\bf Iteration6},
every outer y axis line/.append style={darkgray!60!black},
every y tick label/.append style={font=\color{darkgray!60!black}},
y dir=reverse,
ymin=0.5,
ymax=2.5,
yticklabels={,,1,,2},
name=plot6,
at=(plot4.below south west),
anchor=above north west
]
\addplot graphics [xmin=0.5,xmax=20.5,ymin=0.5,ymax=2.5] {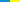};
\addplot [
color=mycolor1,
solid,
forget plot
]
table[row sep=crcr]{
1 1.5\\
20 1.5\\
};
\end{axis}

\begin{axis}[%
width=2.64938446969697in,
height=0.653655181751278in,
axis on top,
scale only axis,
separate axis lines,
every outer x axis line/.append style={darkgray!60!black},
every x tick label/.append style={font=\color{darkgray!60!black}},
xmin=0.5,
xmax=20.5,
xlabel={\bf Iteration5},
every outer y axis line/.append style={darkgray!60!black},
every y tick label/.append style={font=\color{darkgray!60!black}},
y dir=reverse,
ymin=0.5,
ymax=2.5,
yticklabels={,,1,,2},
at=(plot6.left of south west),
anchor=right of south east
]
\addplot graphics [xmin=0.5,xmax=20.5,ymin=0.5,ymax=2.5] {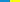};
\addplot [
color=mycolor1,
solid,
forget plot
]
table[row sep=crcr]{
1 1.5\\
20 1.5\\
};
\end{axis}
\end{tikzpicture}%
} &
                                                        \scalebox{.4}{
%
%
%
\definecolor{mycolor1}{rgb}{0,0.447,0.741}%
\begin{tikzpicture}

\begin{axis}[%
width=2.64938446969697in,
height=0.653655181751278in,
axis on top,
scale only axis,
separate axis lines,
every outer x axis line/.append style={darkgray!60!black},
every x tick label/.append style={font=\color{darkgray!60!black}},
xmin=0.5,
xmax=20.5,
xlabel={\bf Iteration3},
every outer y axis line/.append style={darkgray!60!black},
every y tick label/.append style={font=\color{darkgray!60!black}},
y dir=reverse,
ymin=0.5,
ymax=2.5,
        scale only axis,		
yticklabels={,,1,,2},
name=plot3
]
\addplot graphics [xmin=0.5,xmax=20.5,ymin=0.5,ymax=2.5] {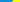};
\addplot [
color=mycolor1,
solid,
forget plot
]
table[row sep=crcr]{
1 1.5\\
20 1.5\\
};
\end{axis}

\begin{axis}[%
width=2.64938446969697in,
height=0.653655181751278in,
axis on top,
scale only axis,
separate axis lines,
every outer x axis line/.append style={darkgray!60!black},
every x tick label/.append style={font=\color{darkgray!60!black}},
xmin=0.5,
xmax=20.5,
xlabel={\bf Iteration1},
every outer y axis line/.append style={darkgray!60!black},
every y tick label/.append style={font=\color{darkgray!60!black}},
y dir=reverse,
ymin=0.5,
ymax=2.5,
name=plot1,
yticklabels={,,1,,2},
at=(plot3.above north west),
anchor=below south west
]
\addplot graphics [xmin=0.5,xmax=20.5,ymin=0.5,ymax=2.5] {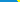};
\addplot [
color=mycolor1,
solid,
forget plot
]
table[row sep=crcr]{
1 1.5\\
20 1.5\\
};
\end{axis}

\begin{axis}[%
width=2.64938446969697in,
height=0.653655181751278in,
axis on top,
scale only axis,
separate axis lines,
every outer x axis line/.append style={darkgray!60!black},
every x tick label/.append style={font=\color{darkgray!60!black}},
xmin=0.5,
xmax=20.5,
xlabel={\bf Iteration2},
every outer y axis line/.append style={darkgray!60!black},
every y tick label/.append style={font=\color{darkgray!60!black}},
y dir=reverse,
ymin=0.5,
ymax=2.5,
name=plot2,
yticklabels={,,1,,2},
at=(plot1.right of south east),
anchor=left of south west
]
\addplot graphics [xmin=0.5,xmax=20.5,ymin=0.5,ymax=2.5] {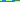};
\addplot [
color=mycolor1,
solid,
forget plot
]
table[row sep=crcr]{
1 1.5\\
20 1.5\\
};
\end{axis}

\begin{axis}[%
width=2.64938446969697in,
height=0.653655181751278in,
axis on top,
scale only axis,
separate axis lines,
every outer x axis line/.append style={darkgray!60!black},
every x tick label/.append style={font=\color{darkgray!60!black}},
xmin=0.5,
xmax=20.5,
xlabel={\bf Iteration4},
every outer y axis line/.append style={darkgray!60!black},
every y tick label/.append style={font=\color{darkgray!60!black}},
y dir=reverse,
ymin=0.5,
ymax=2.5,
name=plot4,
yticklabels={,,1,,2},
at=(plot2.below south west),
anchor=above north west
]
\addplot graphics [xmin=0.5,xmax=20.5,ymin=0.5,ymax=2.5] {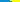};
\addplot [
color=mycolor1,
solid,
forget plot
]
table[row sep=crcr]{
1 1.5\\
20 1.5\\
};
\end{axis}

\begin{axis}[%
width=2.64938446969697in,
height=0.653655181751278in,
axis on top,
scale only axis,
separate axis lines,
every outer x axis line/.append style={darkgray!60!black},
every x tick label/.append style={font=\color{darkgray!60!black}},
xmin=0.5,
xmax=20.5,
xlabel={\bf Iteration6},
every outer y axis line/.append style={darkgray!60!black},
every y tick label/.append style={font=\color{darkgray!60!black}},
y dir=reverse,
ymin=0.5,
ymax=2.5,
name=plot6,
yticklabels={,,1,,2},
at=(plot4.below south west),
anchor=above north west
]
\addplot graphics [xmin=0.5,xmax=20.5,ymin=0.5,ymax=2.5] {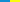};
\addplot [
color=mycolor1,
solid,
forget plot
]
table[row sep=crcr]{
1 1.5\\
20 1.5\\
};
\end{axis}

\begin{axis}[%
width=2.64938446969697in,
height=0.653655181751278in,
axis on top,
scale only axis,
separate axis lines,
every outer x axis line/.append style={darkgray!60!black},
every x tick label/.append style={font=\color{darkgray!60!black}},
xmin=0.5,
xmax=20.5,
xlabel={\bf Iteration5},
every outer y axis line/.append style={darkgray!60!black},
every y tick label/.append style={font=\color{darkgray!60!black}},
y dir=reverse,
ymin=0.5,
ymax=2.5,
yticklabels={,,1,,2},
at=(plot6.left of south west),
anchor=right of south east
]
\addplot graphics [xmin=0.5,xmax=20.5,ymin=0.5,ymax=2.5] {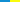};
\addplot [
color=mycolor1,
solid,
forget plot
]
table[row sep=crcr]{
1 1.5\\
20 1.5\\
};
\end{axis}
\end{tikzpicture}%
} \\
      \scalebox{.4}{
%
%
%
\definecolor{mycolor1}{rgb}{1,0,1}%
\begin{tikzpicture}

\begin{axis}[%
width=2.64938446969697in,
height=0.653655181751278in,
scale only axis,
separate axis lines,
every outer x axis line/.append style={darkgray!60!black},
every x tick label/.append style={font=\color{darkgray!60!black}},
xmin=1,
xmax=20,
xlabel={\bf Iteration3},
every outer y axis line/.append style={darkgray!60!black},
every y tick label/.append style={font=\color{darkgray!60!black}},
ymin=-0.932546653752473,
ymax=3.74458703134455,
name=plot3
]
\addplot [
color=mycolor1,
solid,
forget plot
]
table[row sep=crcr]{
1 3.74458703134455\\
2 2.88560234861128\\
3 2.18524868312691\\
4 1.61136644139244\\
5 1.13632034901155\\
6 0.736999450690583\\
7 0.394817110238581\\
8 0.0957110105672632\\
9 -0.169856846308968\\
10 -0.406900139273027\\
11 -0.615908228105153\\
12 -0.792846153482888\\
13 -0.929154636981103\\
14 -0.932546653752473\\
15 -0.846747154011435\\
16 -0.663750976612398\\
17 -0.354212848383778\\
18 0.116542022975406\\
19 0.788517948895489\\
20 1.70704875993619\\
};
\addplot [
color=mycolor1,
solid,
mark=*,
mark options={solid,scale=.5},
forget plot
]
table[row sep=crcr]{
1 2.27681216677763\\
2 1.27164523161921\\
3 0.419417619876139\\
4 0.368743681674129\\
5 0.322393439716186\\
6 0.263456195112679\\
7 0.213626078241462\\
8 0.0208797834403495\\
9 0.00204127950446654\\
10 0.00020012221584579\\
11 2.02556301775323e-05\\
12 2.7711490758163e-06\\
13 1.76981247744842e-06\\
14 9.51289556046877e-06\\
15 8.97705272637266e-05\\
16 0.000911834242886077\\
17 0.00932355684224954\\
18 0.0953885678390201\\
19 0.975960964408866\\
20 1.98551360510289\\
};
\end{axis}

\begin{axis}[%
width=2.64938446969697in,
height=0.653655181751278in,
scale only axis,
separate axis lines,
every outer x axis line/.append style={darkgray!60!black},
every x tick label/.append style={font=\color{darkgray!60!black}},
xmin=1,
xmax=20,
xlabel={\bf Iteration1},
every outer y axis line/.append style={darkgray!60!black},
every y tick label/.append style={font=\color{darkgray!60!black}},
ymin=0,
ymax=11.0027146363735,
name=plot1,
at=(plot3.above north west),
anchor=below south west
]
\addplot [
color=mycolor1,
solid,
forget plot
]
table[row sep=crcr]{
1 11.0027146363735\\
2 9.47840611358308\\
3 8.2387842877808\\
4 7.22550930400338\\
5 6.38748633988333\\
6 5.68086560564904\\
7 5.06904234412476\\
8 4.52265683073062\\
9 4.01959437348257\\
10 3.54498531299248\\
11 3.09120502246804\\
12 2.65787390771283\\
13 2.25185740712627\\
14 1.88726599170368\\
15 1.58545516503622\\
16 1.37502546331091\\
17 1.29182245531064\\
18 1.43393247073434\\
19 2.03024000857897\\
20 3.03482155330346\\
};
\addplot [
color=mycolor1,
solid,
mark=*,
mark options={solid,scale=.5},
forget plot
]
table[row sep=crcr]{
1 9.02262355864218\\
2 7.93664973491126\\
3 6.98138502678998\\
4 6.14109712790951\\
5 5.40194729121806\\
6 4.75176241790393\\
7 4.17983457796999\\
8 3.67674466075347\\
9 3.23420725108637\\
10 2.84493417640068\\
11 2.502514478008\\
12 2.20130883470494\\
13 1.93635675020519\\
14 1.70329548993542\\
15 1.49829469188014\\
16 1.31805938924945\\
17 1.16045209807259\\
18 1.03137769756152\\
19 1.71523038982963\\
20 2.71381789517182\\
};
\end{axis}

\begin{axis}[%
width=2.64938446969697in,
height=0.653655181751278in,
scale only axis,
separate axis lines,
every outer x axis line/.append style={darkgray!60!black},
every x tick label/.append style={font=\color{darkgray!60!black}},
xmin=1,
xmax=20,
xlabel={\bf Iteration2},
every outer y axis line/.append style={darkgray!60!black},
every y tick label/.append style={font=\color{darkgray!60!black}},
ymin=-57.0311147599823,
ymax=9.02262355864218,
name=plot2,
at=(plot1.right of south east),
anchor=left of south west
]
\addplot [
color=mycolor1,
solid,
forget plot
]
table[row sep=crcr]{
1 -2.33221251073817\\
2 -3.4492433889318\\
3 -2.37429486965755\\
4 -2.20277694101939\\
5 -3.03731103433027\\
6 -4.22035297800232\\
7 -4.09412276950767\\
8 -3.05128893796424\\
9 -1.84500638814393\\
10 -0.696229664508998\\
11 0.11228538430624\\
12 0.23598160549551\\
13 0.513880516162686\\
14 1.66581537339374\\
15 1.43143345954033\\
16 -1.07743360159049\\
17 -6.90352544096993\\
18 -17.2441529443473\\
19 -33.4511982522497\\
20 -57.0311147599823\\
};
\addplot [
color=mycolor1,
solid,
mark=*,
mark options={solid,scale=.5},
forget plot
]
table[row sep=crcr]{
1 9.02262355864218\\
2 7.93664973491127\\
3 6.98138502678999\\
4 6.14109712790955\\
5 5.40194729121847\\
6 4.7517624179081\\
7 4.17983457801273\\
8 3.67674466119069\\
9 3.23420725555978\\
10 2.84493422217029\\
11 2.50251494629848\\
12 2.20131362600606\\
13 1.93640577226991\\
14 1.70379705783332\\
15 1.50342646994105\\
16 1.51041472603057\\
17 5.41089112715811\\
18 6.52782749884204\\
19 7.87041716678223\\
20 8.99125023689685\\
};
\end{axis}

\begin{axis}[%
width=2.64938446969697in,
height=0.653655181751278in,
scale only axis,
separate axis lines,
every outer x axis line/.append style={darkgray!60!black},
every x tick label/.append style={font=\color{darkgray!60!black}},
xmin=1,
xmax=20,
xlabel={\bf Iteration4},
every outer y axis line/.append style={darkgray!60!black},
every y tick label/.append style={font=\color{darkgray!60!black}},
ymin=0,
ymax=9.32698030348594,
name=plot4,
at=(plot2.below south west),
anchor=above north west
]
\addplot [
color=mycolor1,
solid,
forget plot
]
table[row sep=crcr]{
1 7.83698792079317\\
2 7.24835603083627\\
3 6.54755895669173\\
4 5.78380859446514\\
5 5.00205435160988\\
6 4.24298314692716\\
7 3.54301941056598\\
8 2.93432508402315\\
9 2.44479962014331\\
10 2.09807998311889\\
11 1.98273862887251\\
12 2.3352425684735\\
13 2.87182929288024\\
14 3.57426322882066\\
15 4.41697955000722\\
16 5.36708417713693\\
17 6.38435377789133\\
18 7.42123576693648\\
19 8.42284830592297\\
20 9.32698030348594\\
};
\addplot [
color=mycolor1,
solid,
mark=*,
mark options={solid,scale=.5},
forget plot
]
table[row sep=crcr]{
1 9.02262355864349\\
2 7.93664973491404\\
3 6.98138502680897\\
4 6.14109712809552\\
5 5.40194729311399\\
6 4.75176243729575\\
7 4.17983477637128\\
8 3.67674669068579\\
9 3.23422802027838\\
10 2.84514667580992\\
11 2.50468865982703\\
12 2.22355391578877\\
13 3.46602193920892\\
14 4.03197788492338\\
15 4.73731084394382\\
16 5.40053436333422\\
17 6.14095850293738\\
18 6.98137087683617\\
19 7.93664766860716\\
20 9.02262257986655\\
};
\end{axis}

\begin{axis}[%
width=2.64938446969697in,
height=0.653655181751278in,
scale only axis,
separate axis lines,
every outer x axis line/.append style={darkgray!60!black},
every x tick label/.append style={font=\color{darkgray!60!black}},
xmin=1,
xmax=20,
xlabel={\bf Iteration6},
every outer y axis line/.append style={darkgray!60!black},
every y tick label/.append style={font=\color{darkgray!60!black}},
ymin=0,
ymax=9.47194572619271,
name=plot6,
at=(plot4.below south west),
anchor=above north west
]
\addplot [
color=mycolor1,
solid,
forget plot
]
table[row sep=crcr]{
1 9.31698397517046\\
2 8.32576086990222\\
3 7.36206933139203\\
4 6.44663314312024\\
5 5.59841219752332\\
6 4.83460249599387\\
7 4.17063614888056\\
8 3.62018137548819\\
9 3.19514250407768\\
10 2.90565997186603\\
11 2.91712180685402\\
12 3.19964397025015\\
13 3.61436354396407\\
14 4.15358511711775\\
15 4.80881512161955\\
16 5.57076183216427\\
17 6.42933536623311\\
18 7.3736476840937\\
19 8.39201258880007\\
20 9.47194572619271\\
};
\addplot [
color=mycolor1,
solid,
mark=*,
mark options={solid,scale=.5},
forget plot
]
table[row sep=crcr]{
1 9.02262355887143\\
2 7.93664973539525\\
3 6.98138503010426\\
4 6.14109716037897\\
5 5.40194762216131\\
6 4.75176580282609\\
7 4.17986920972696\\
8 3.67709899375368\\
9 3.23783259860929\\
10 2.88202681854233\\
11 2.88202681854233\\
12 3.23783259860929\\
13 3.67709899375368\\
14 4.17986920972696\\
15 4.75176580282609\\
16 5.40194762216131\\
17 6.14109716037897\\
18 6.98138503010426\\
19 7.93664973539525\\
20 9.02262355887144\\
};
\end{axis}

\begin{axis}[%
width=2.64938446969697in,
height=0.653655181751278in,
scale only axis,
separate axis lines,
every outer x axis line/.append style={darkgray!60!black},
every x tick label/.append style={font=\color{darkgray!60!black}},
xmin=1,
xmax=20,
xlabel={\bf Iteration5},
every outer y axis line/.append style={darkgray!60!black},
every y tick label/.append style={font=\color{darkgray!60!black}},
ymin=0,
ymax=9.47194572619271,
at=(plot3.below south west),
anchor=above north west
]
\addplot [
color=mycolor1,
solid,
forget plot
]
table[row sep=crcr]{
1 9.31698397517046\\
2 8.32576086990222\\
3 7.36206933139203\\
4 6.44663314312024\\
5 5.59841219752332\\
6 4.83460249599387\\
7 4.17063614888056\\
8 3.62018137548819\\
9 3.19514250407768\\
10 2.90565997186603\\
11 2.91712180685402\\
12 3.19964397025015\\
13 3.61436354396407\\
14 4.15358511711775\\
15 4.80881512161955\\
16 5.57076183216427\\
17 6.42933536623311\\
18 7.3736476840937\\
19 8.39201258880007\\
20 9.47194572619271\\
};
\addplot [
color=mycolor1,
solid,
mark=*,
mark options={solid,scale=.5},
forget plot
]
table[row sep=crcr]{
1 9.02262355887143\\
2 7.93664973539525\\
3 6.98138503010426\\
4 6.14109716037897\\
5 5.40194762216131\\
6 4.75176580282609\\
7 4.17986920972696\\
8 3.67709899375368\\
9 3.23783259860929\\
10 2.88202681854233\\
11 2.88202681854233\\
12 3.23783259860929\\
13 3.67709899375368\\
14 4.17986920972696\\
15 4.75176580282609\\
16 5.40194762216131\\
17 6.14109716037897\\
18 6.98138503010426\\
19 7.93664973539525\\
20 9.02262355887144\\
};
\end{axis}
\end{tikzpicture}%
} &
                                                         \scalebox{.4}{
%
%
%
\definecolor{mycolor1}{rgb}{1,0,1}%
\begin{tikzpicture}

\begin{axis}[%
width=2.64938446969697in,
height=0.653655181751278in,
scale only axis,
separate axis lines,
every outer x axis line/.append style={darkgray!60!black},
every x tick label/.append style={font=\color{darkgray!60!black}},
xmin=1,
xmax=20,
xlabel={\bf Iteration3},
every outer y axis line/.append style={darkgray!60!black},
every y tick label/.append style={font=\color{darkgray!60!black}},
ymin=-0.926632246993929,
ymax=3.7481527116622,
name=plot3
]
\addplot [
color=mycolor1,
solid,
forget plot
]
table[row sep=crcr]{
1 3.7481527116622\\
2 2.88649008619483\\
3 2.18480078902787\\
4 1.61061045482212\\
5 1.13600139458222\\
6 0.737612595656616\\
7 0.396639721737603\\
8 0.0988351128612925\\
9 -0.165492214592373\\
10 -0.401476567899623\\
11 -0.609695577992861\\
12 -0.786170199460662\\
13 -0.922364710547768\\
14 -0.926632246993929\\
15 -0.841873627032681\\
16 -0.65999730401828\\
17 -0.351462593641617\\
18 0.118644227073325\\
19 0.790609919769533\\
20 1.71009428475688\\
};
\addplot [
color=mycolor1,
solid,
mark=*,
mark options={solid,scale=.5},
forget plot
]
table[row sep=crcr]{
1 2.27681216677763\\
2 1.27164523161921\\
3 0.419417619876139\\
4 0.368743681674129\\
5 0.322393439716186\\
6 0.263456195112679\\
7 0.213626078241462\\
8 0.0208797834403495\\
9 0.00204127950446654\\
10 0.00020012221584579\\
11 2.02556301775323e-05\\
12 2.7711490758163e-06\\
13 1.76981247744842e-06\\
14 9.51289556046877e-06\\
15 8.97705272637266e-05\\
16 0.000911834242886077\\
17 0.00932355684224954\\
18 0.0953885678390201\\
19 0.975960964408866\\
20 1.98551360510289\\
};
\end{axis}

\begin{axis}[%
width=2.64938446969697in,
height=0.653655181751278in,
scale only axis,
separate axis lines,
every outer x axis line/.append style={darkgray!60!black},
every x tick label/.append style={font=\color{darkgray!60!black}},
xmin=1,
xmax=20,
xlabel={\bf Iteration1},
every outer y axis line/.append style={darkgray!60!black},
every y tick label/.append style={font=\color{darkgray!60!black}},
ymin=0,
ymax=11.1827396934899,
name=plot1,
at=(plot3.above north west),
anchor=below south west
]
\addplot [
color=mycolor1,
solid,
forget plot
]
table[row sep=crcr]{
1 11.1827396934899\\
2 9.62796756385752\\
3 8.36462081532788\\
4 7.33306860515617\\
5 6.48105688561784\\
6 5.76370840400857\\
7 5.14352270264426\\
8 4.59037611886103\\
9 4.08152178501523\\
10 3.60158962848342\\
11 3.14258637166239\\
12 2.70389553196915\\
13 2.29227742184093\\
14 1.92186914873519\\
15 1.61418461512961\\
16 1.39811451852207\\
17 1.30992635143073\\
18 1.44476280732624\\
19 2.04076188662614\\
20 3.04809211705106\\
};
\addplot [
color=mycolor1,
solid,
mark=*,
mark options={solid,scale=.5},
forget plot
]
table[row sep=crcr]{
1 9.02262355864218\\
2 7.93664973491126\\
3 6.98138502678998\\
4 6.14109712790951\\
5 5.40194729121806\\
6 4.75176241790393\\
7 4.17983457796999\\
8 3.67674466075347\\
9 3.23420725108637\\
10 2.84493417640068\\
11 2.502514478008\\
12 2.20130883470494\\
13 1.93635675020519\\
14 1.70329548993542\\
15 1.49829469188014\\
16 1.31805938924945\\
17 1.16045209807259\\
18 1.03137769756152\\
19 1.71523038982963\\
20 2.71381789517182\\
};
\end{axis}

\begin{axis}[%
width=2.64938446969697in,
height=0.653655181751278in,
scale only axis,
separate axis lines,
every outer x axis line/.append style={darkgray!60!black},
every x tick label/.append style={font=\color{darkgray!60!black}},
xmin=1,
xmax=20,
xlabel={\bf Iteration2},
every outer y axis line/.append style={darkgray!60!black},
every y tick label/.append style={font=\color{darkgray!60!black}},
ymin=-77.1979250591496,
ymax=9.02262355864218,
name=plot2,
at=(plot1.right of south east),
anchor=left of south west
]
\addplot [
color=mycolor1,
solid,
forget plot
]
table[row sep=crcr]{
1 -6.7525905864209\\
2 -7.65555286929114\\
3 -6.31150724652624\\
4 -5.69187364232418\\
5 -6.48417114593872\\
6 -7.79471811071872\\
7 -7.02286912867414\\
8 -5.47941529898375\\
9 -3.7348459817014\\
10 -2.07695536084933\\
11 -0.879322525404376\\
12 -0.601311469297983\\
13 -0.272133294330729\\
14 1.37994342459791\\
15 1.15793553731794\\
16 -2.13313530172007\\
17 -9.89711343728618\\
18 -23.7467092133697\\
19 -45.503498973181\\
20 -77.1979250591496\\
};
\addplot [
color=mycolor1,
solid,
mark=*,
mark options={solid,scale=.5},
forget plot
]
table[row sep=crcr]{
1 9.02262355864218\\
2 7.93664973491127\\
3 6.98138502678999\\
4 6.14109712790955\\
5 5.40194729121847\\
6 4.7517624179081\\
7 4.17983457801273\\
8 3.67674466119069\\
9 3.23420725555978\\
10 2.84493422217029\\
11 2.50251494629848\\
12 2.20131362600606\\
13 1.93640577226991\\
14 1.70379705783332\\
15 1.50342646994105\\
16 1.51041472603057\\
17 5.41089112715811\\
18 6.52782749884204\\
19 7.87041716678223\\
20 8.99125023689685\\
};
\end{axis}

\begin{axis}[%
width=2.64938446969697in,
height=0.653655181751278in,
scale only axis,
separate axis lines,
every outer x axis line/.append style={darkgray!60!black},
every x tick label/.append style={font=\color{darkgray!60!black}},
xmin=1,
xmax=20,
xlabel={\bf Iteration4},
every outer y axis line/.append style={darkgray!60!black},
every y tick label/.append style={font=\color{darkgray!60!black}},
ymin=0,
ymax=9.32390504623943,
name=plot4,
at=(plot2.below south west),
anchor=above north west
]
\addplot [
color=mycolor1,
solid,
forget plot
]
table[row sep=crcr]{
1 7.9987753618843\\
2 7.37847151310364\\
3 6.65557720062306\\
4 5.87694040186297\\
5 5.0854241052149\\
6 4.31990631004148\\
7 3.61528002667649\\
8 3.00245327642478\\
9 2.50834909156236\\
10 2.15590551533633\\
11 2.01846451716294\\
12 2.36040521575027\\
13 2.88532948189682\\
14 3.57572079203955\\
15 4.40701463681346\\
16 5.3475985210516\\
17 6.35881196378509\\
18 7.39494649824309\\
19 8.40324567185277\\
20 9.32390504623943\\
};
\addplot [
color=mycolor1,
solid,
mark=*,
mark options={solid,scale=.5},
forget plot
]
table[row sep=crcr]{
1 9.02262355864349\\
2 7.93664973491404\\
3 6.98138502680897\\
4 6.14109712809552\\
5 5.40194729311399\\
6 4.75176243729575\\
7 4.17983477637128\\
8 3.67674669068579\\
9 3.23422802027838\\
10 2.84514667580992\\
11 2.50468865982703\\
12 2.22355391578877\\
13 3.46602193920892\\
14 4.03197788492338\\
15 4.73731084394382\\
16 5.40053436333422\\
17 6.14095850293738\\
18 6.98137087683617\\
19 7.93664766860716\\
20 9.02262257986655\\
};
\end{axis}

\begin{axis}[%
width=2.64938446969697in,
height=0.653655181751278in,
scale only axis,
separate axis lines,
every outer x axis line/.append style={darkgray!60!black},
every x tick label/.append style={font=\color{darkgray!60!black}},
xmin=1,
xmax=20,
xlabel={\bf Iteration6},
every outer y axis line/.append style={darkgray!60!black},
every y tick label/.append style={font=\color{darkgray!60!black}},
ymin=0,
ymax=9.49556318569697,
name=plot6,
at=(plot4.below south west),
anchor=above north west
]
\addplot [
color=mycolor1,
solid,
forget plot
]
table[row sep=crcr]{
1 9.41791597693521\\
2 8.40978519705381\\
3 7.4323134528241\\
4 6.50567291071943\\
5 5.64833322102011\\
6 4.87706151781344\\
7 4.20692241899373\\
8 3.65127802626222\\
9 3.22178792512718\\
10 2.92840918490382\\
11 2.93252724497834\\
12 3.21300647541906\\
13 3.62610549819652\\
14 4.16419201217553\\
15 4.81888390382398\\
16 5.58104924721281\\
17 6.44080630401597\\
18 7.38752352351051\\
19 8.40981954257646\\
20 9.49556318569697\\
};
\addplot [
color=mycolor1,
solid,
mark=*,
mark options={solid,scale=.5},
forget plot
]
table[row sep=crcr]{
1 9.02262355887143\\
2 7.93664973539525\\
3 6.98138503010426\\
4 6.14109716037897\\
5 5.40194762216131\\
6 4.75176580282609\\
7 4.17986920972696\\
8 3.67709899375368\\
9 3.23783259860929\\
10 2.88202681854233\\
11 2.88202681854233\\
12 3.23783259860929\\
13 3.67709899375368\\
14 4.17986920972696\\
15 4.75176580282609\\
16 5.40194762216131\\
17 6.14109716037897\\
18 6.98138503010426\\
19 7.93664973539525\\
20 9.02262355887144\\
};
\end{axis}

\begin{axis}[%
width=2.64938446969697in,
height=0.653655181751278in,
scale only axis,
separate axis lines,
every outer x axis line/.append style={darkgray!60!black},
every x tick label/.append style={font=\color{darkgray!60!black}},
xmin=1,
xmax=20,
xlabel={\bf Iteration5},
every outer y axis line/.append style={darkgray!60!black},
every y tick label/.append style={font=\color{darkgray!60!black}},
ymin=0,
ymax=9.49556318569697,
at=(plot3.below south west),
anchor=above north west
]
\addplot [
color=mycolor1,
solid,
forget plot
]
table[row sep=crcr]{
1 9.41791597693521\\
2 8.40978519705381\\
3 7.4323134528241\\
4 6.50567291071943\\
5 5.64833322102011\\
6 4.87706151781344\\
7 4.20692241899373\\
8 3.65127802626222\\
9 3.22178792512718\\
10 2.92840918490382\\
11 2.93252724497834\\
12 3.21300647541906\\
13 3.62610549819652\\
14 4.16419201217553\\
15 4.81888390382398\\
16 5.58104924721281\\
17 6.44080630401597\\
18 7.38752352351051\\
19 8.40981954257646\\
20 9.49556318569697\\
};
\addplot [
color=mycolor1,
solid,
mark=*,
mark options={solid,scale=.5},
forget plot
]
table[row sep=crcr]{
1 9.02262355887143\\
2 7.93664973539525\\
3 6.98138503010426\\
4 6.14109716037897\\
5 5.40194762216131\\
6 4.75176580282609\\
7 4.17986920972696\\
8 3.67709899375368\\
9 3.23783259860929\\
10 2.88202681854233\\
11 2.88202681854233\\
12 3.23783259860929\\
13 3.67709899375368\\
14 4.17986920972696\\
15 4.75176580282609\\
16 5.40194762216131\\
17 6.14109716037897\\
18 6.98138503010426\\
19 7.93664973539525\\
20 9.02262355887144\\
};
\end{axis}
\end{tikzpicture}%
} \\                                                  
      {\small(a) Bayesian LSPI (BLSPI)} &  {\small(b) LSPI}
    \end{tabular}
  \end{center}
  \caption{BLSPI and LSPI iterations on the $20-$state chain problem. Top: Improved policy after each iteration of LSPI and Bayesian LSPI, respectively (R action - yellow shade; L action - blue shade; BLSPI/LSPI - top stripe; exact - bottom stripe). Bottom:  State value function $V^{\pol}(\vs)$ of the policy being evaluated at each iteration (BLSPI/LSPI approximation - solid line; exact values - dotted line)}
  \label{fig:chain}
  \hspace{-4mm}
\end{figure}

Figure~\ref{fig:chain} illustrates the policies returned after a single run of BLSPI and LSPI on the chain walk problem, respectively.
In both cases, the initial policy was selected to be the policy that chooses left action at each state.
Furthermore, both algorithms use the same single set of samples ($5000$ in total) collected from a single episode with the actions to be chosen uniformly at random.
The optimal policy in this domain is to go left in states $1-10$ and right in states $11-20$.
Therefore, it can be easily verified that both algorithms need only $6$ iterations to discover the optimal policy.
Additionally, this example shows that the performance of BLSPI is quite close to that of LSPI. 

\subsection*{Randomised Bayesian LSPI Performance}
\label{sec:rblspi_performance}

Let us now present our results for the online RBLSPI algorithm that constitutes the main contribution of this work.
The main advantage of RBLSPI is its efficient exploration scheme that is based on the idea of the randomised value function \citep{pmlr-v48-osband16}.
To examine the exploration efficiency of the proposed Randomised Bayesian LSPI (RBLSPI) algorithm, we have considered four well-known continuous state, discrete-action, episodic domains: i) (sparse) Mountain Car, ii) Inverted Pendulum, iii) Cart Pole, and iv) Puddle World.
Comparisons have been conducted with an onpolicy variant of online LSPI algorithm that uses the simple $\epsilon-$greedy exploration scheme. For each experiment, we report the mean performance (average return or number of steps over $100$ episodes) across $100$ independent runs. For each average, we also plot the $95\%$ confidence interval for the accuracy of the mean estimate with error bars. Additionally, we show the $90\%$ percentile region of the runs.

\paragraph{Mountain Car \citep{Moore90, Sutton+Barto:1998}}
The goal in this task is to drive an underpowered car up a steep road to the right hilltop ($p \geq 0.5$) within at most $500$ steps.
The car state in this domain is described by two continuous variables, its position ($p$) and its velocity ($v$).
At each time step, the agent can select between three possible actions: forward, reverse and zero throttle.
In our experiments we have considered two different reward signals:
\begin{itemize}[noitemsep]
  \item In standard mountain car the immediate reward at each time step is equal to $-1$ except in the case where the agent reaches the goal (zero reward). 
  \item The immediate reward is equal to $1$ when the episode terminates and zero otherwise \citep{ijcai2018-376}. We call this version sparse Mountain car.
  \end{itemize}
In both cases, the car is positioned motionless ($v_0=0$) to a uniformly randomly selected position close to the valley ($-0.6 \leq p_0 \leq -0.4$) at the beginning of each episode.
The discount factor is set to $0.99$. 
An equidistant $8 \times 8$ grid of RBFs over the state space plus a constant term is selected.
This set of basis functions is replicated for each action, giving a total of $195$ RBFs.

\begin{figure*}[t]
  \centering
  \begin{subfigure}[t]{.497\textwidth}
    \begin{tabular}{cc}
      \scalebox{.21}{\includegraphics{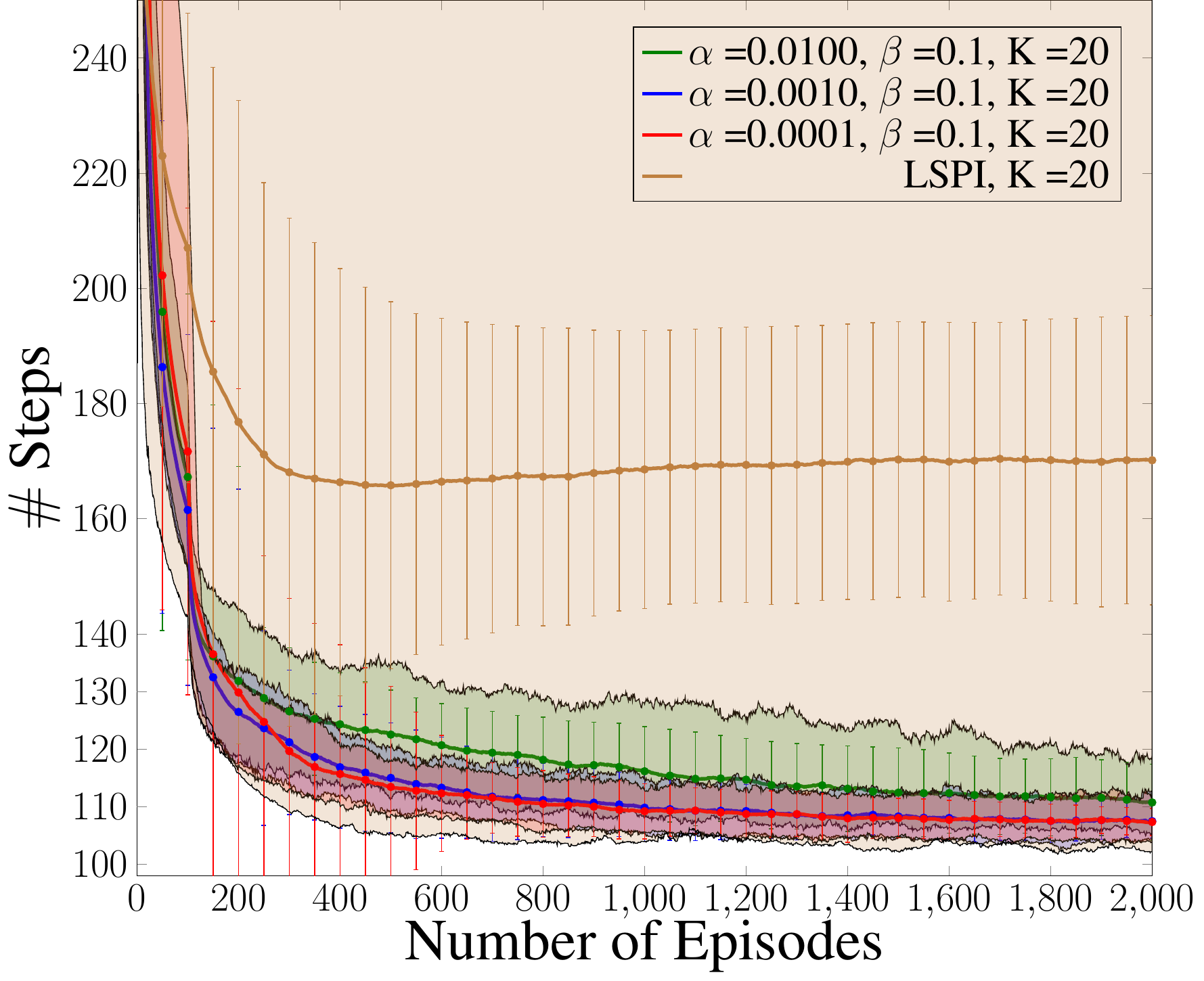}}
      \scalebox{.21}{\includegraphics{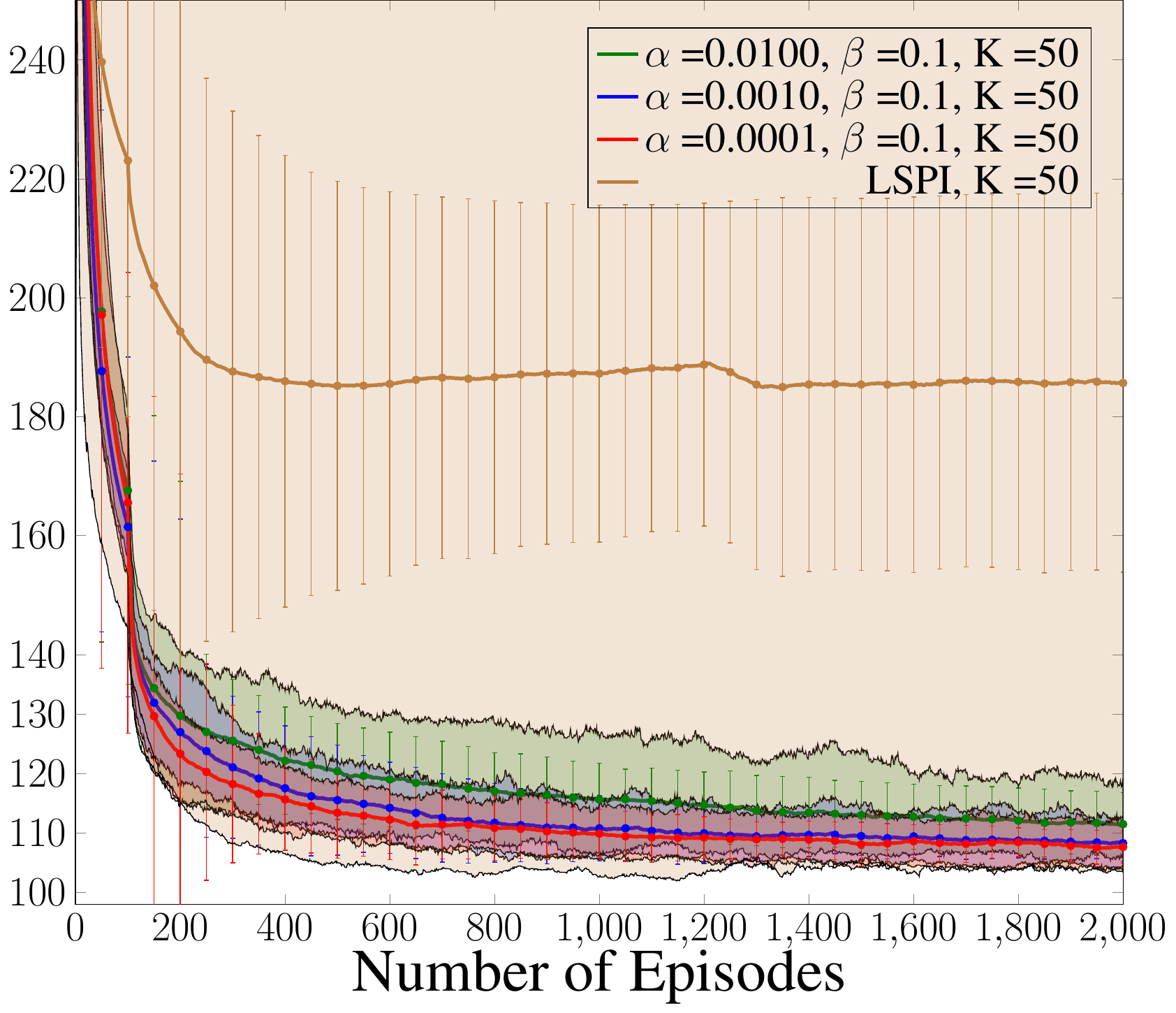}}\\
      \scalebox{.21}{\includegraphics{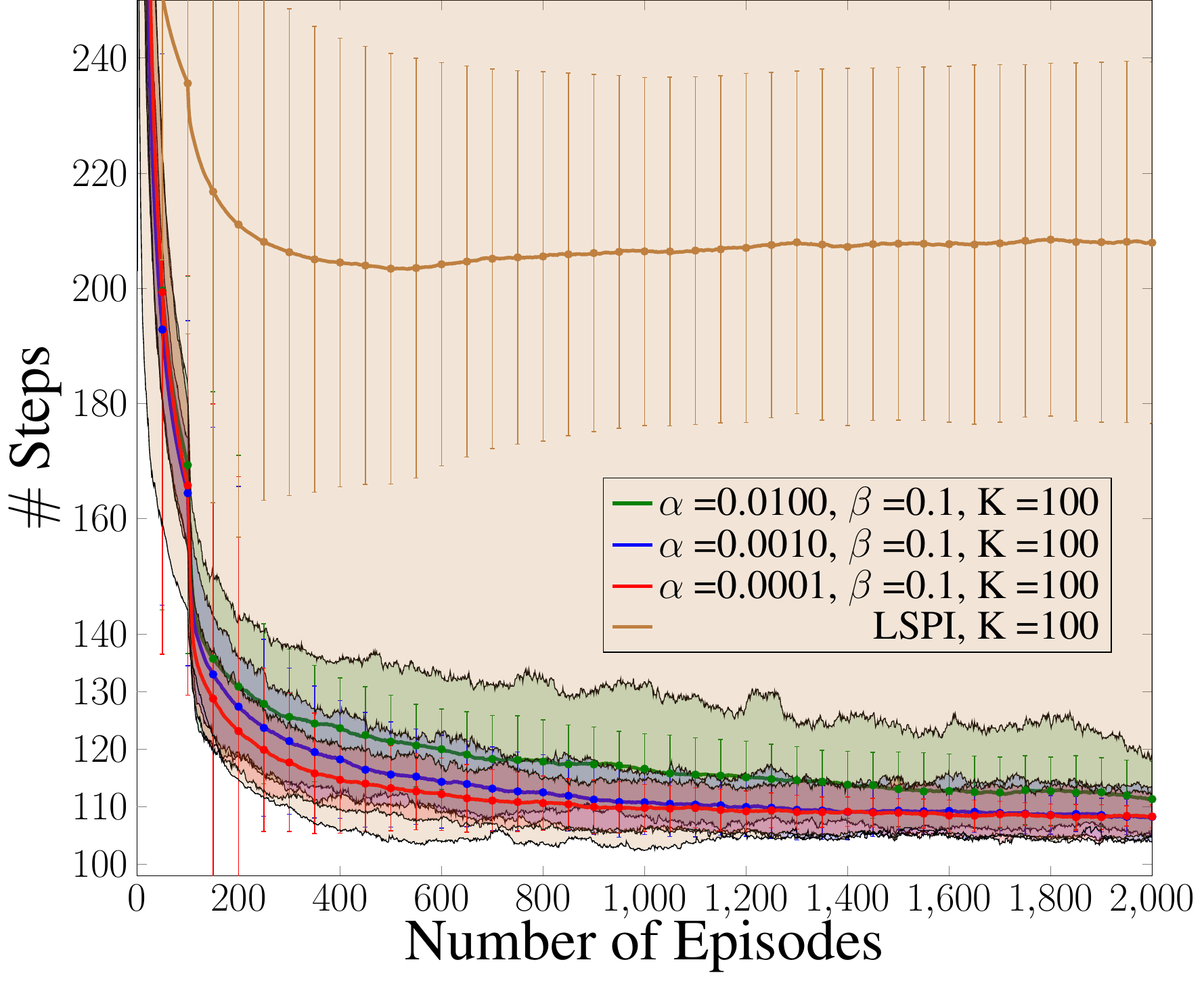}}
      \scalebox{.21}{\includegraphics{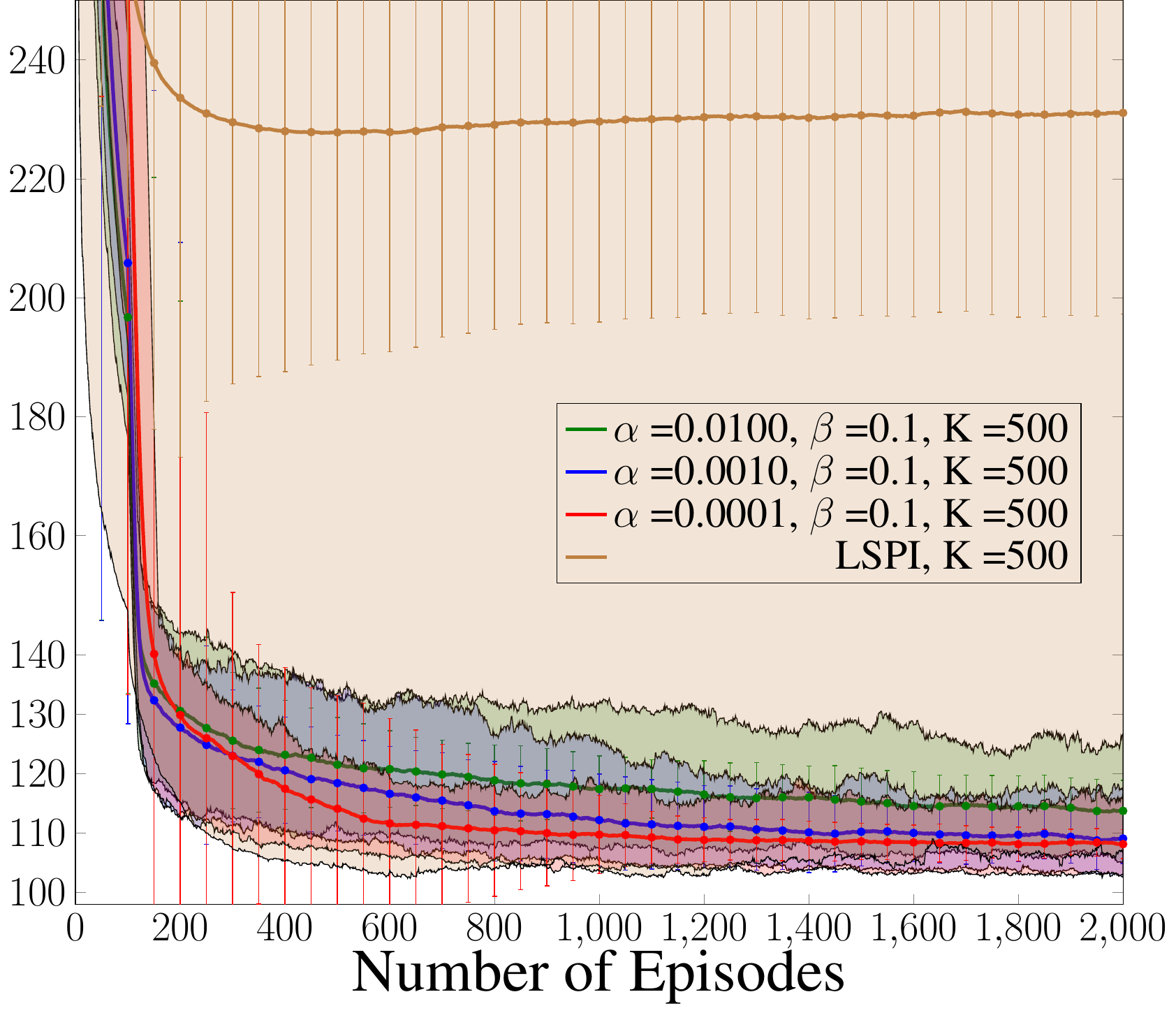}}
    \end{tabular}
    \caption{RBLSPI precision parameter $\beta = 0.1$}
    \label{fig:mountaincar01}
  \end{subfigure}
  \begin{subfigure}[t]{0.497\textwidth}
    \begin{tabular}{cc}
      \scalebox{.21}{\includegraphics{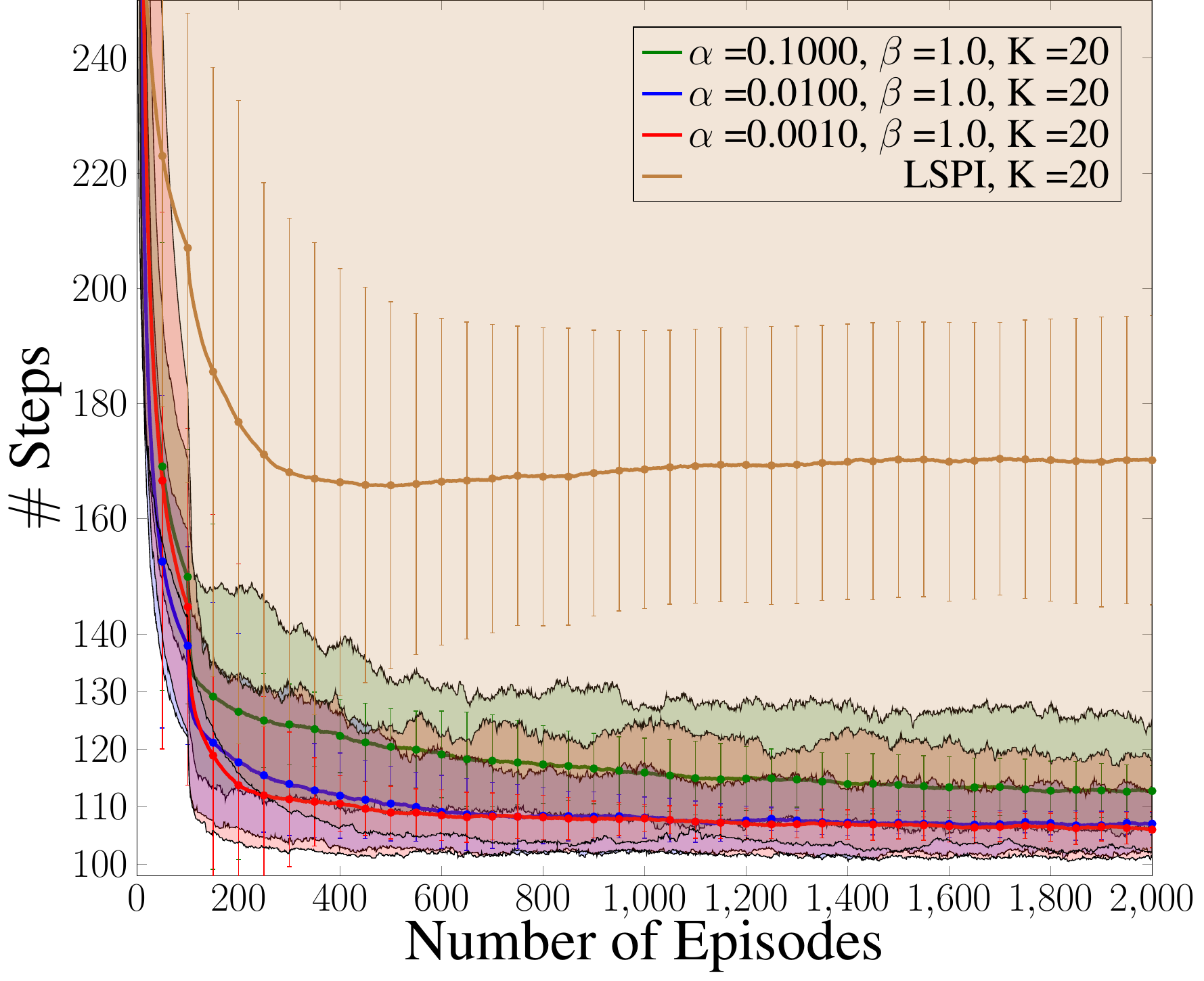}}
      \scalebox{.21}{\includegraphics{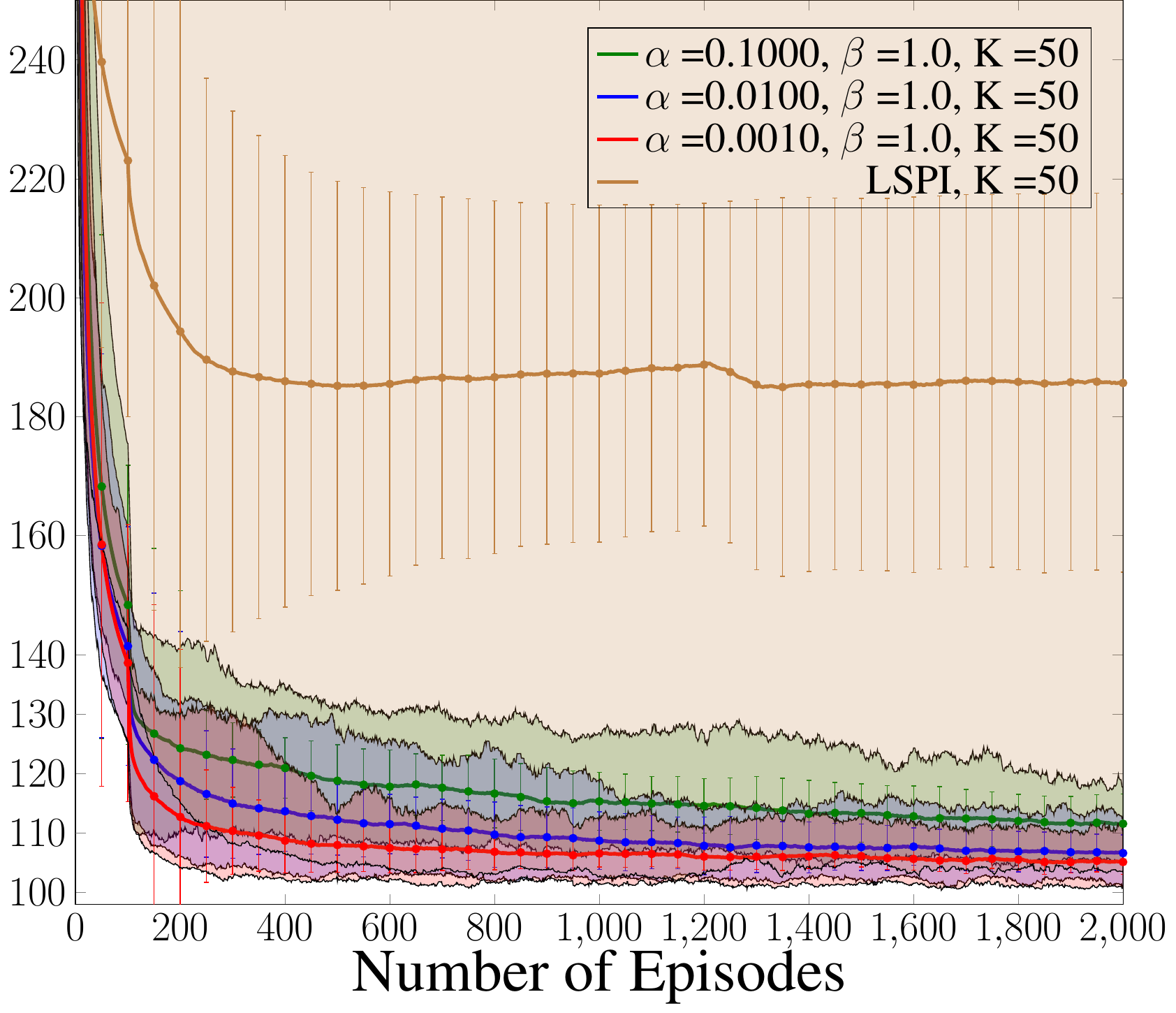}}\\
      \scalebox{.21}{\includegraphics{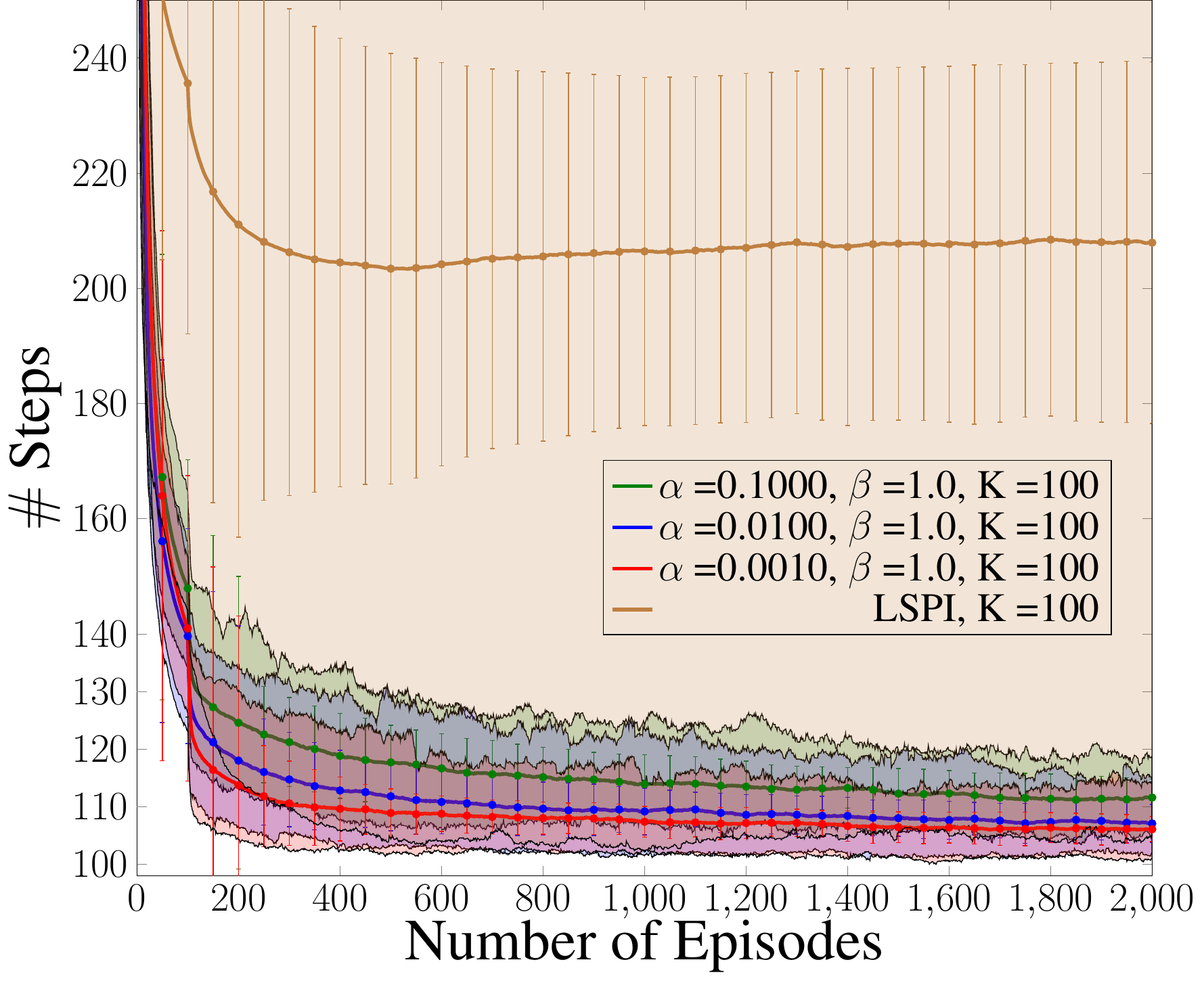}}
      \scalebox{.21}{\includegraphics{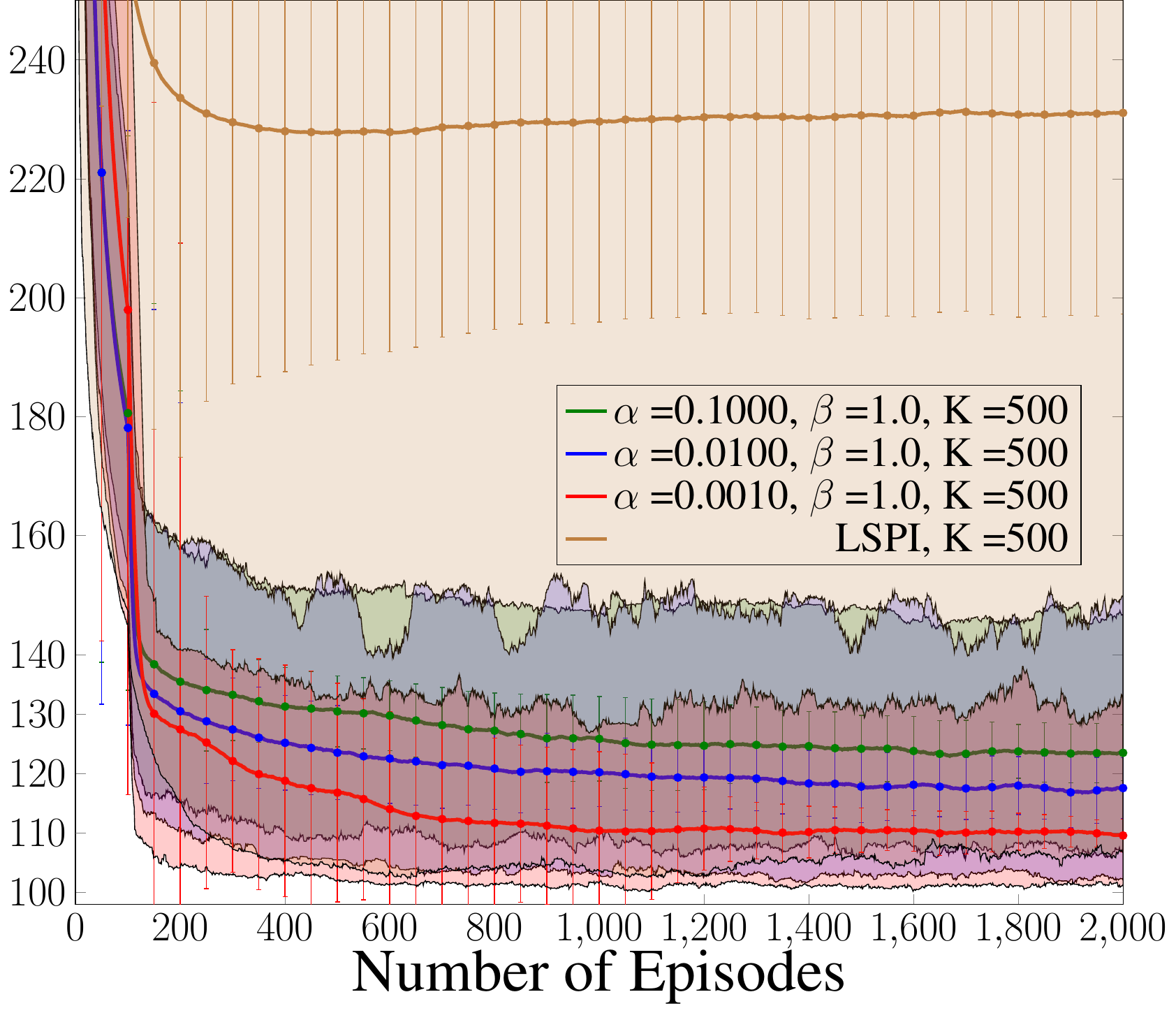}}
    \end{tabular}
    \caption{RBLSPI precision parameter $\beta = 1.0$}
    \label{fig:mountaincar1}
  \end{subfigure}
  \caption{Performance of RBLSPI and online LSPI on the {\bf mountain car} environment for varying parameter $K$. The performance of the RBLSPI algorithm is also presented for varying hyperparameters $\alpha$ and $\beta$.}
  \label{fig:mountaincar}
\end{figure*}
\begin{figure*}[t]
  \centering
  \begin{subfigure}[t]{.497\textwidth}
    \begin{tabular}{cc}
      \scalebox{.21}{\includegraphics{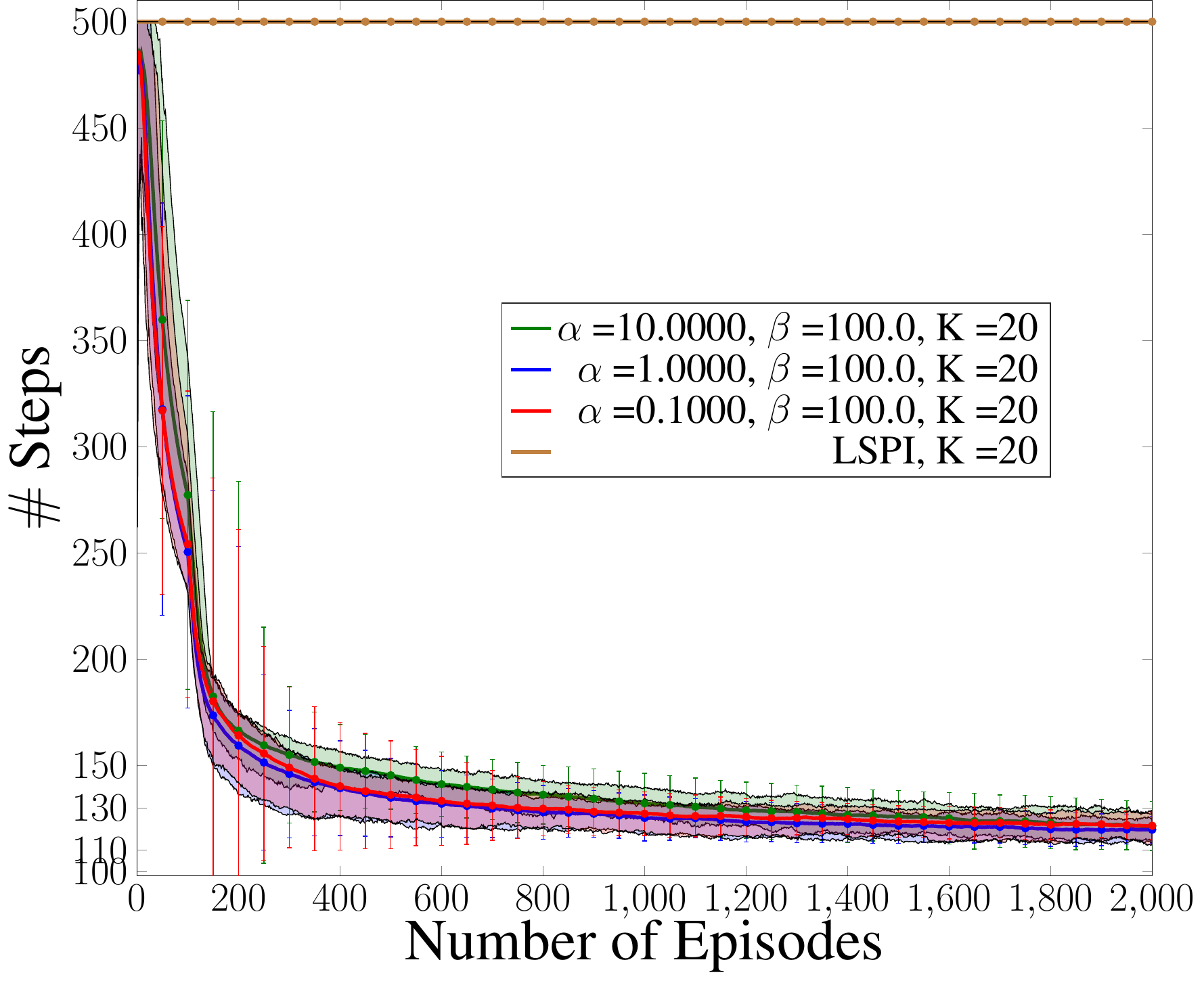}}
      \scalebox{.21}{\includegraphics{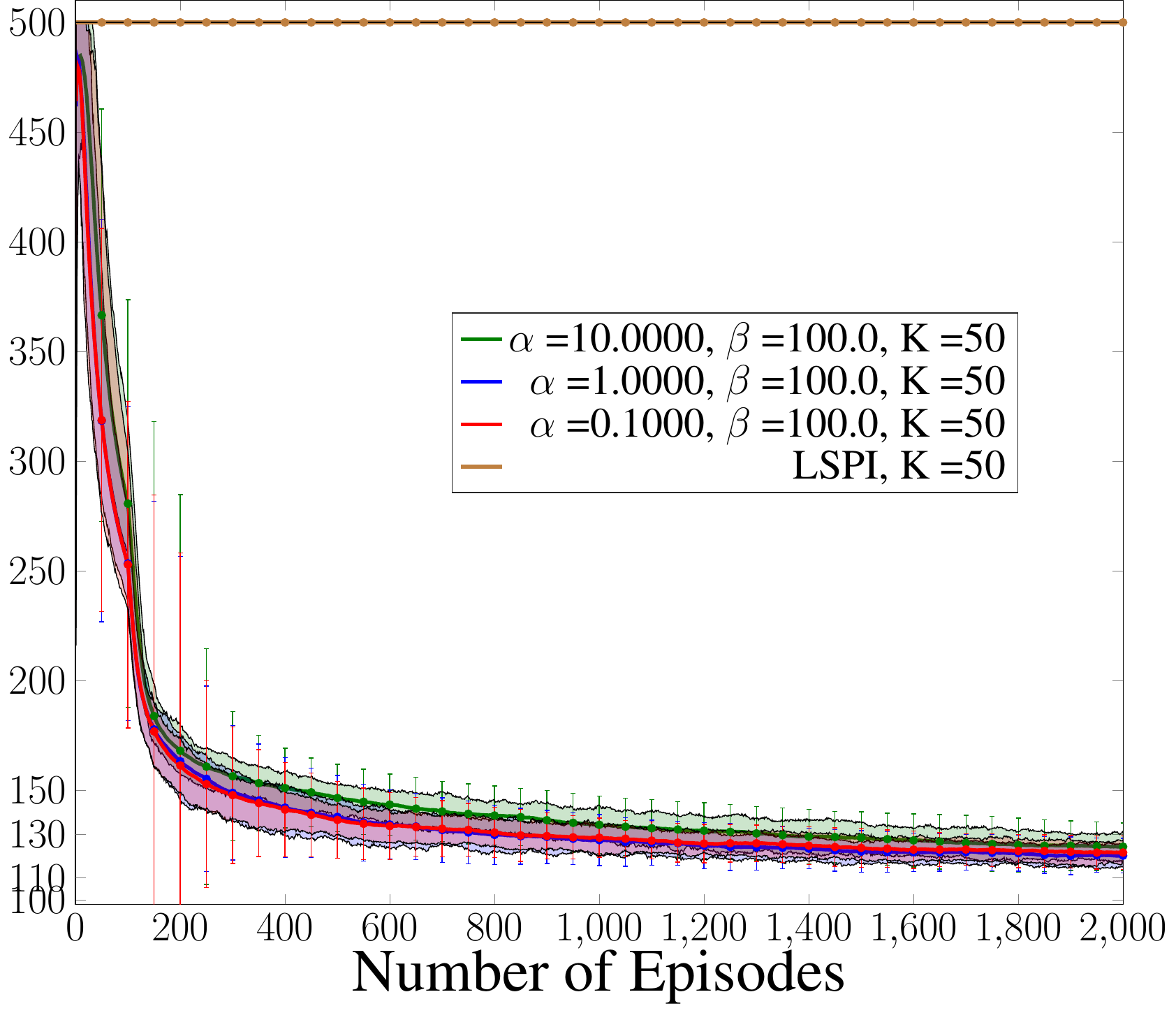}}\\
      \scalebox{.21}{\includegraphics{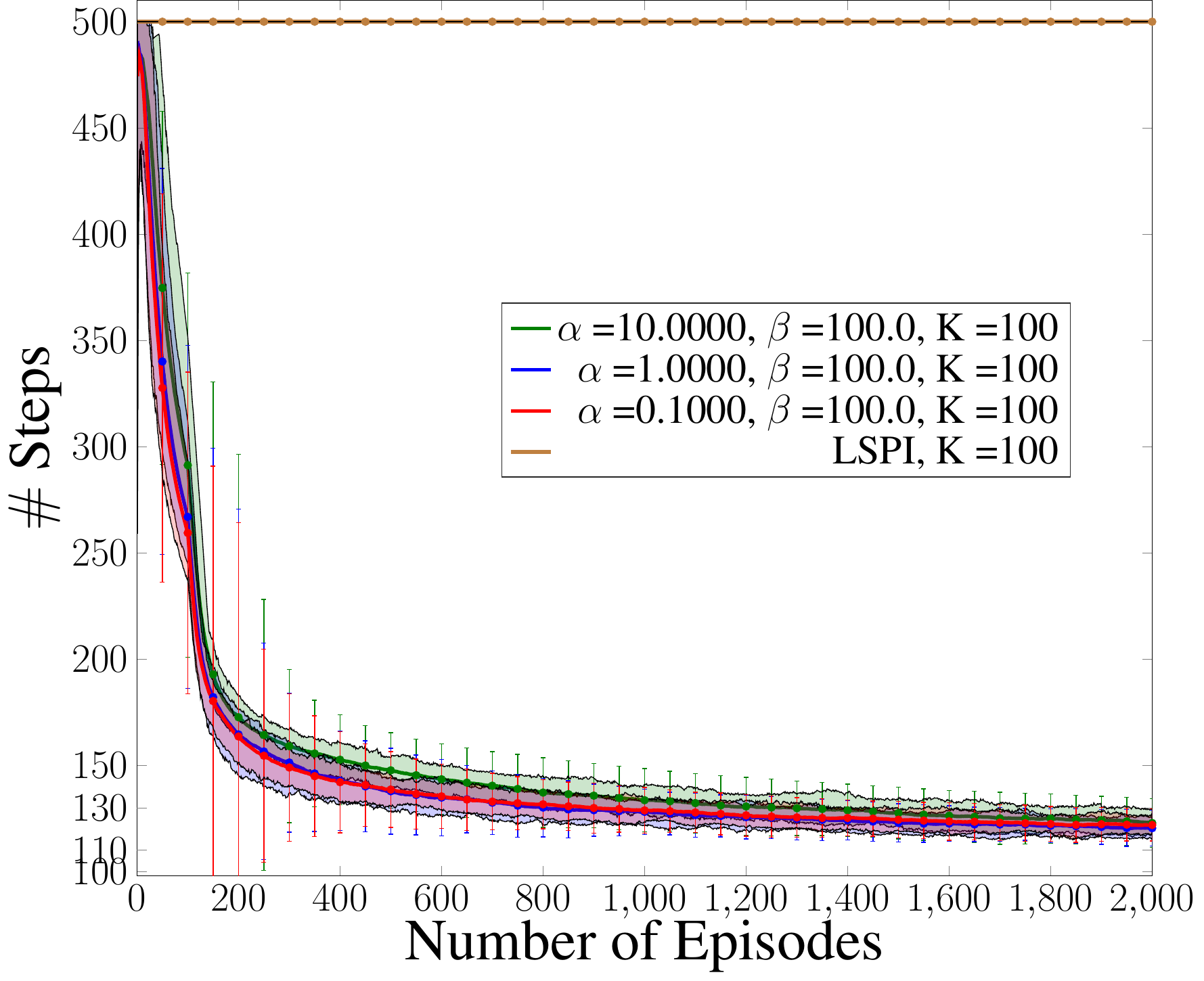}}
      \scalebox{.21}{\includegraphics{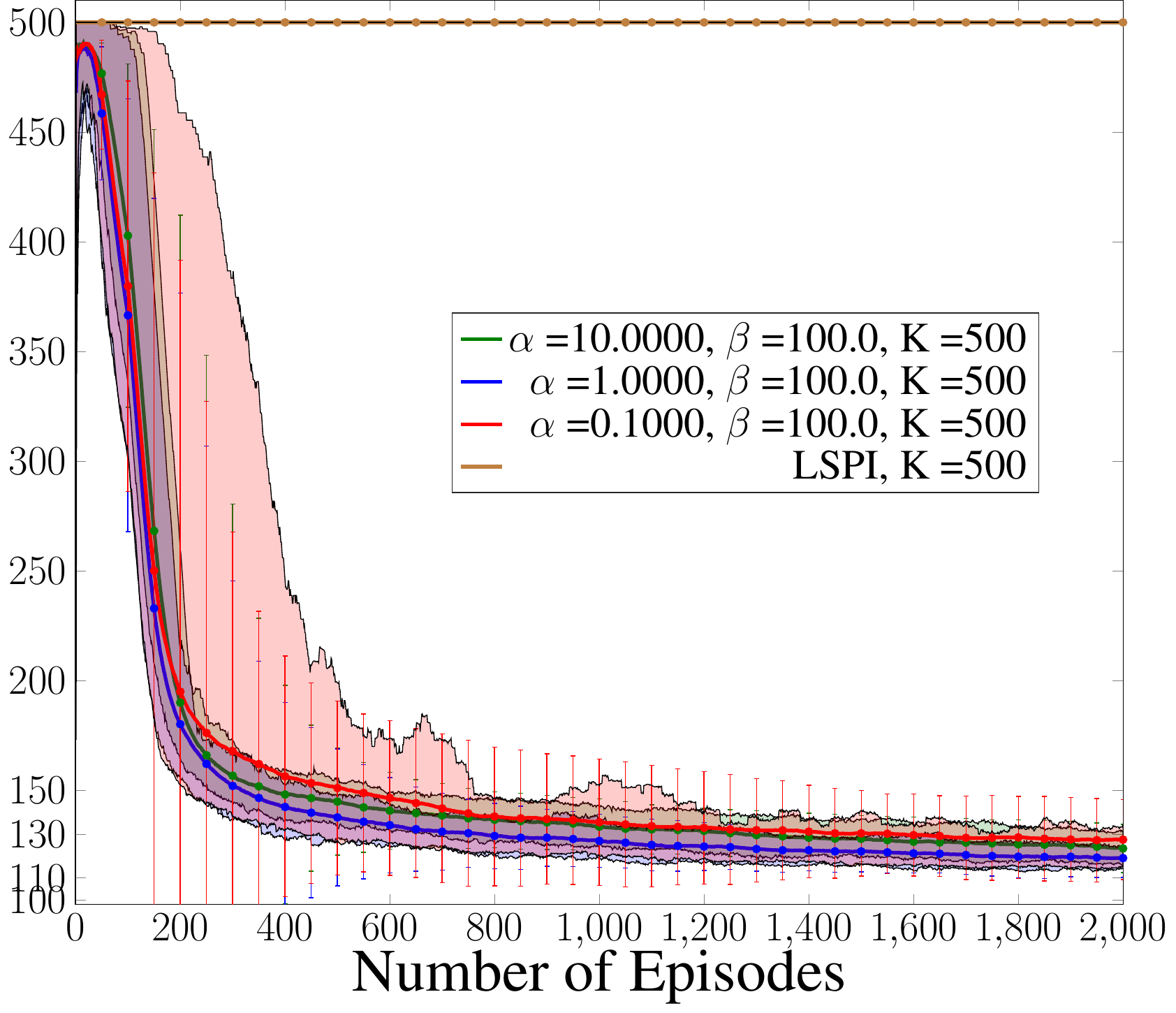}}
    \end{tabular}
    \caption{RBLSPI hyperparameter $\beta = 100$}
    \label{fig:mountaincarsparse100}
  \end{subfigure}
  \begin{subfigure}[t]{0.497\textwidth}
    \begin{tabular}{cc}
      \scalebox{.21}{\includegraphics{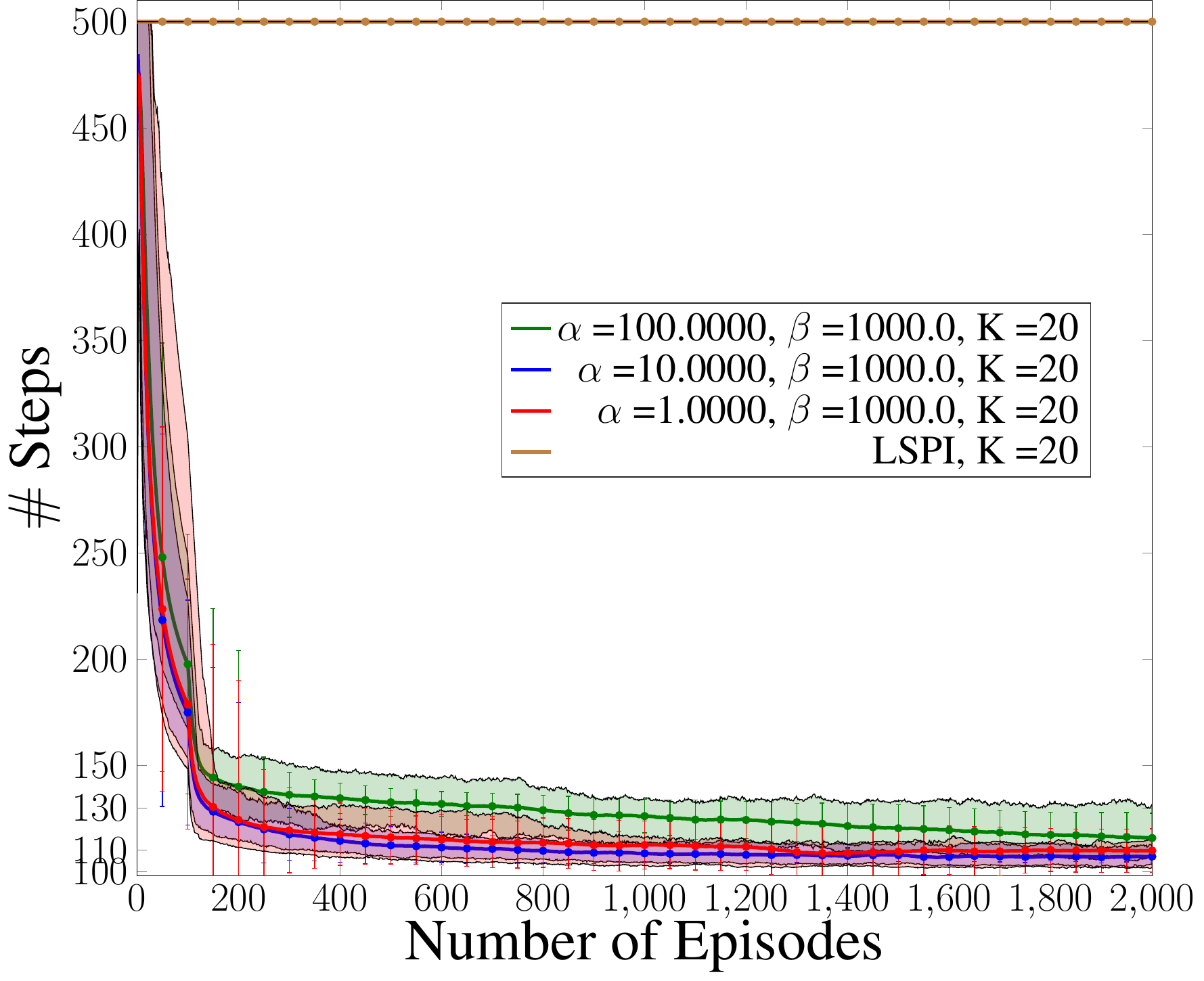}}
      \scalebox{.21}{\includegraphics{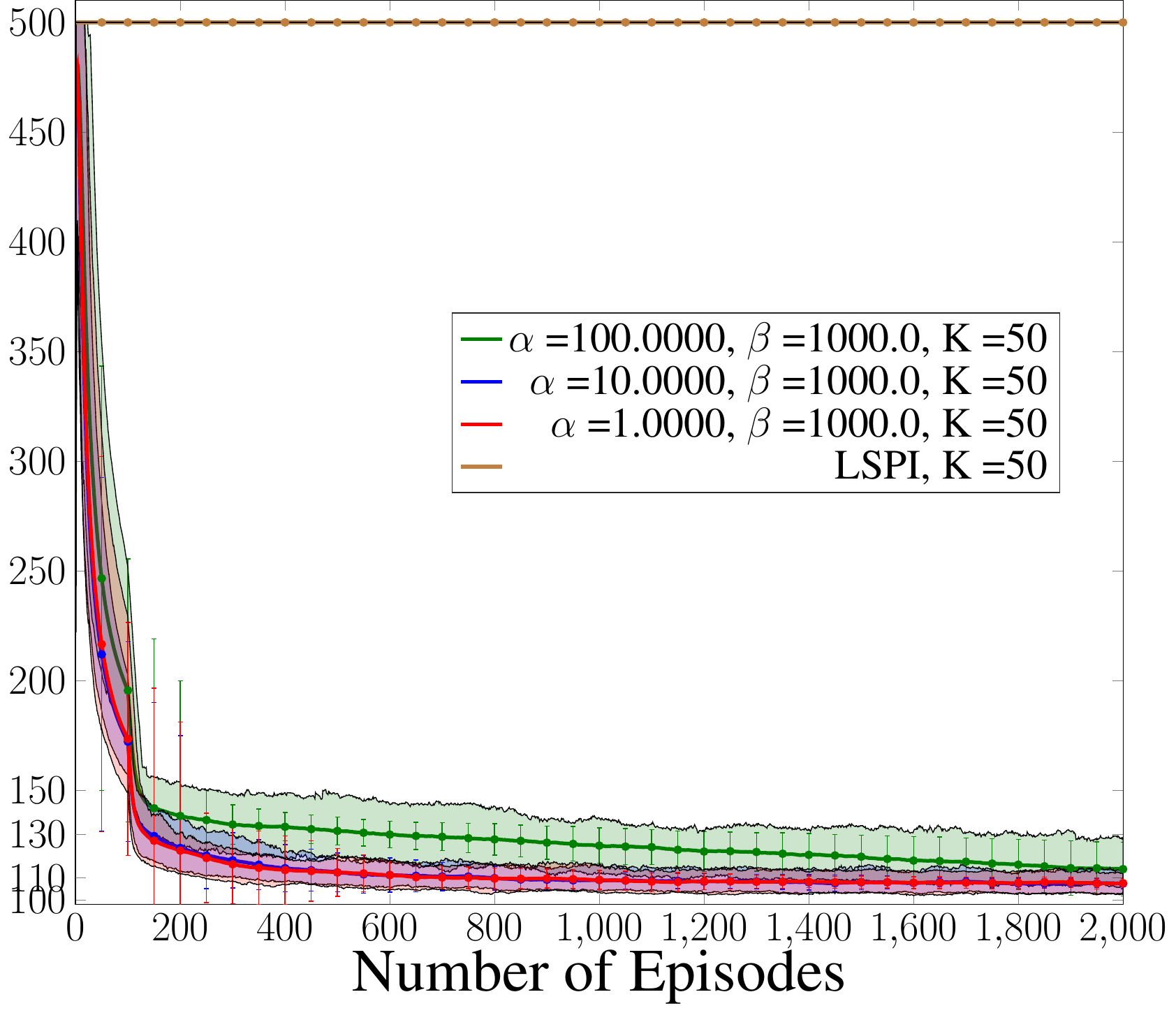}}\\
      \scalebox{.21}{\includegraphics{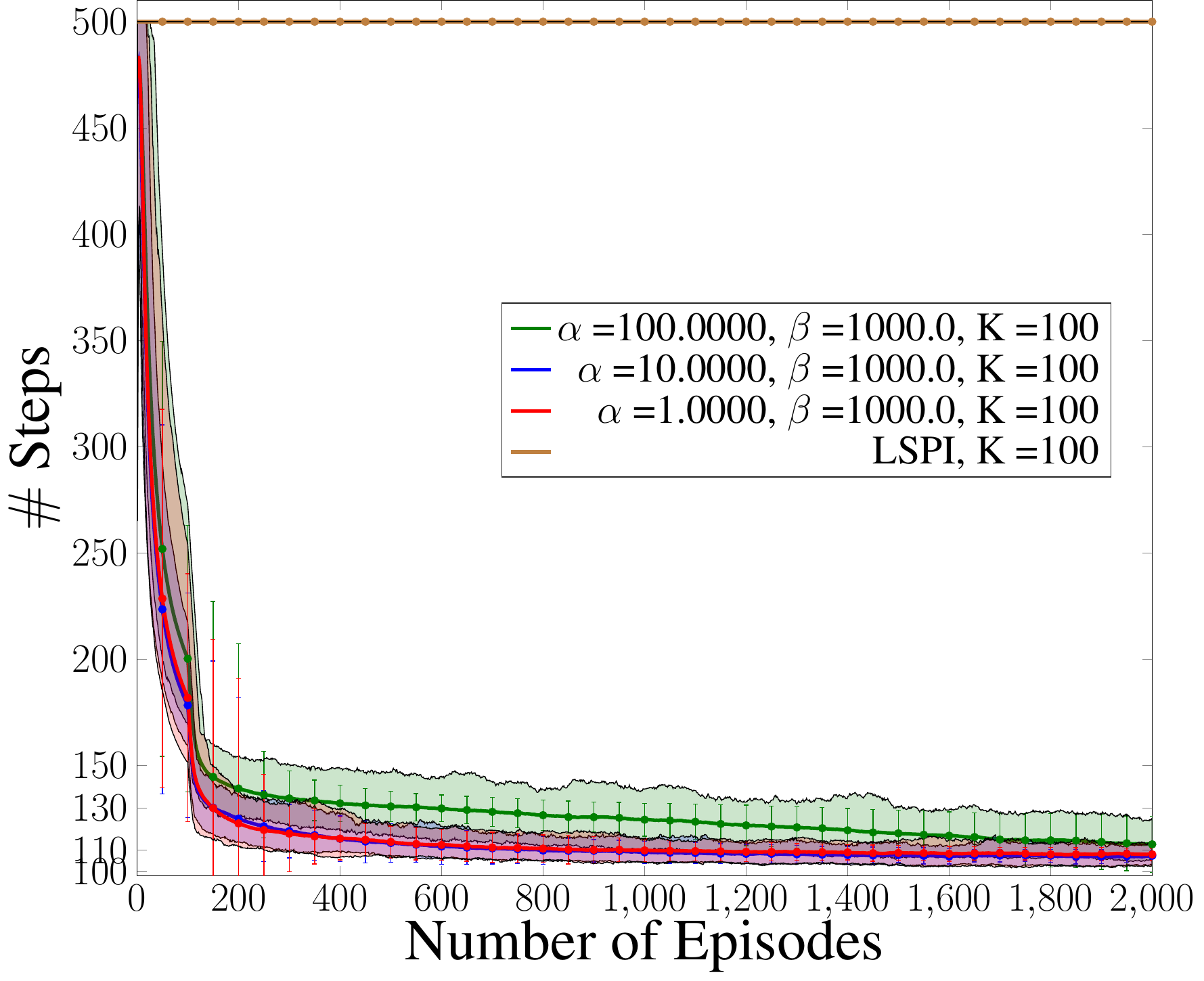}}
      \scalebox{.21}{\includegraphics{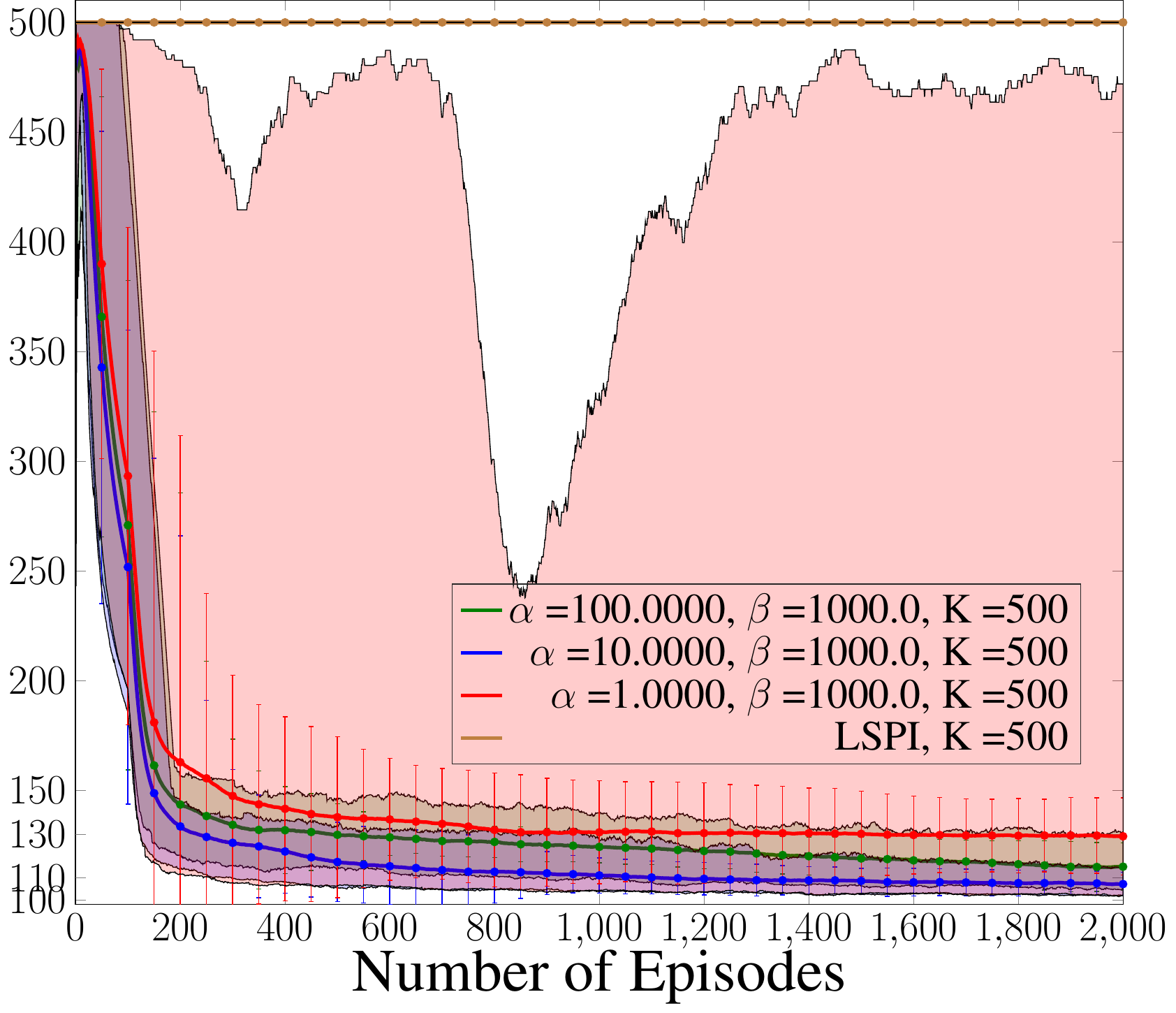}}
    \end{tabular}
    \caption{RBLSPI hyperparameter $\beta = 1000$}
    \label{fig:mountaincarsparse1000}
  \end{subfigure}
  \caption{Performance of RBLSPI and online LSPI on the {\bf sparse reward mountain car} environment for varying parameter $K$. The performance of the RBLSPI algorithm is also presented for varying hyperparameters $\alpha$ and $\beta$.}
  \label{fig:mountaincarsparse}
\end{figure*}

In the task of mountain car the usage of an efficient exploration strategy is of high importance due to the fact that the goal cannot be easily reached by the agent.
The necessity of an effective (deep) exploration scheme becomes more crucial in the case of the sparse reward mountain car environment.
The empirical results of the RBLSPI and online LSPI algorithms on the the standard and sparse mountain car environments are presented in Figures~\ref{fig:mountaincar} and \ref{fig:mountaincarsparse}, respectively. 
In both versions of mountain car environment, we have examined the effects in the performance of the RBLSPI algorithm  of varying the precision parameters $\alpha$ and $\beta$.
Let us first to examine the impact of free parameter $K$ in the performances of both algorithmic schemes.
It should be reminded that parameter $K$ defines the number of transitions that should be conducted between successive policy improvement steps (see Algorithm~\ref{alg:RBLSPI} for details).
For this purpose, the following values have been examined for $K$: $20$, $50$, $100$ and $500$ (end of episode).
As it was expected, the performance of both algorithms is better when $K$ is set equal to a low value, i.e., $K=20$.
It should be also mentioned that even if we set $K$ equal to $500$, the RBLSPI is able to reach the goal in both environments. 
We have also examined the impact of the precision parameter $\alpha$ on the performance of RBLSPI.
Our experiments show that it is more preferable to be set to a lower value if the parameter $K$ is not equal to $500$ (policy updates are executed only at the end of each episode). Nevertheless, it can be easily seen that RBLSPI achieves to discover good policies independent of the selection of the value of precision parameter $\alpha$.
Regarding the precision parameter $\beta$, it impacts a lot the performance of the RBLSPI.
More specifically, the setting of the precision parameter $\beta$ depends heavily on the reward signal of the environment.
For instance, on sparse mountain car we should set $\beta$ to a high value in contrast to the standard mountain car where the best performance is achieved when $\beta$ is set to $0.1$.
Our results also validates our claims about the efficiency of the exploration mechanism of RBLSPI. Actually, RBLSPI outperforms online LSPI in both environments.
More specifically, RBLSPI discovers a near-optimal policy in both environments as our RBLSPI agent is able to reach the goal in less than $110$ steps on average.
On the other hand, the online LSPI stuck in the valley in the case of sparse mountain car, and it doesn't reach the goal in none of the runs.




\begin{figure*}[t]
  \centering
  \begin{subfigure}[t]{.497\textwidth}
    \begin{tabular}{cc}
      \scalebox{.21}{\includegraphics{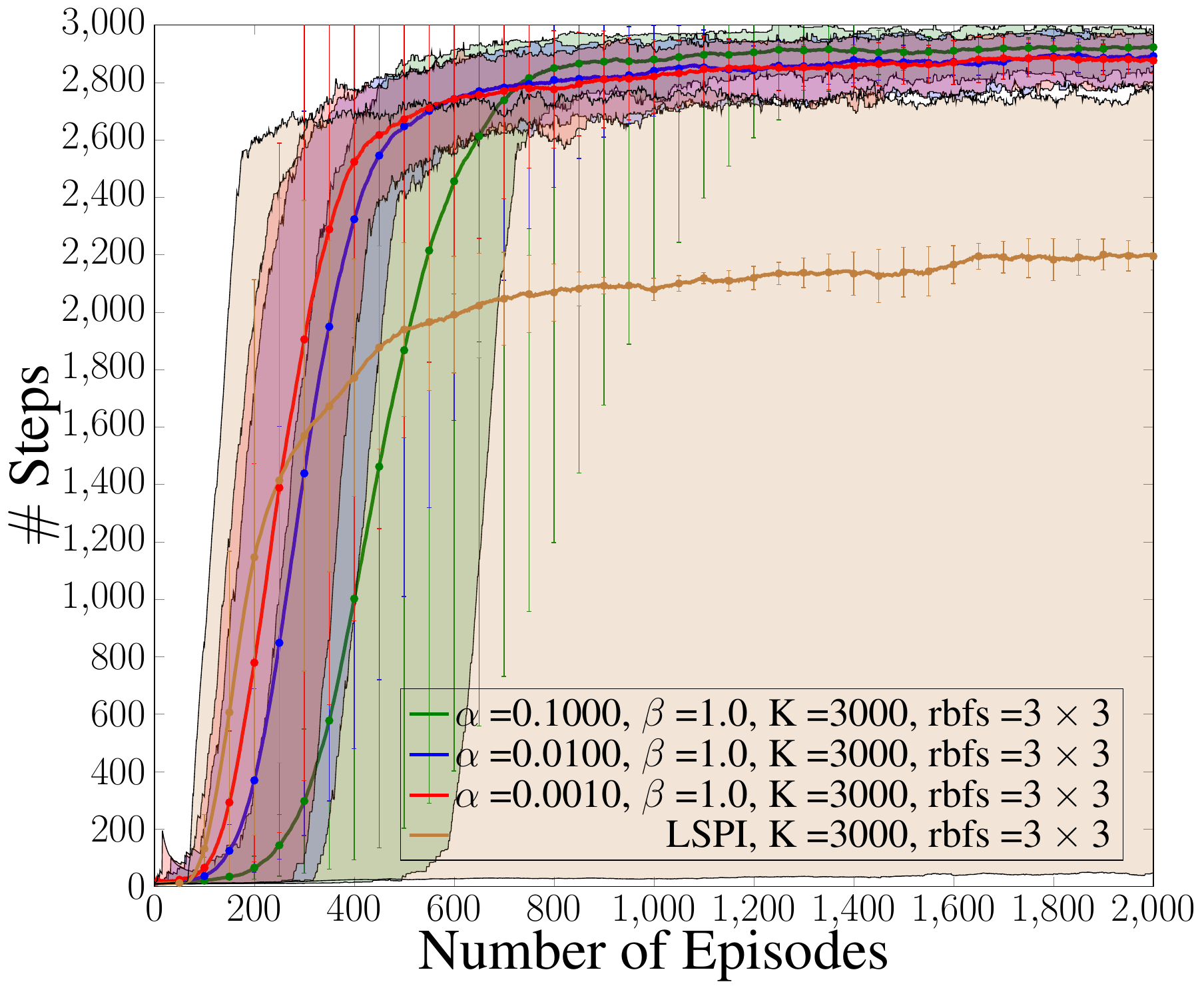}}
      \scalebox{.21}{\includegraphics{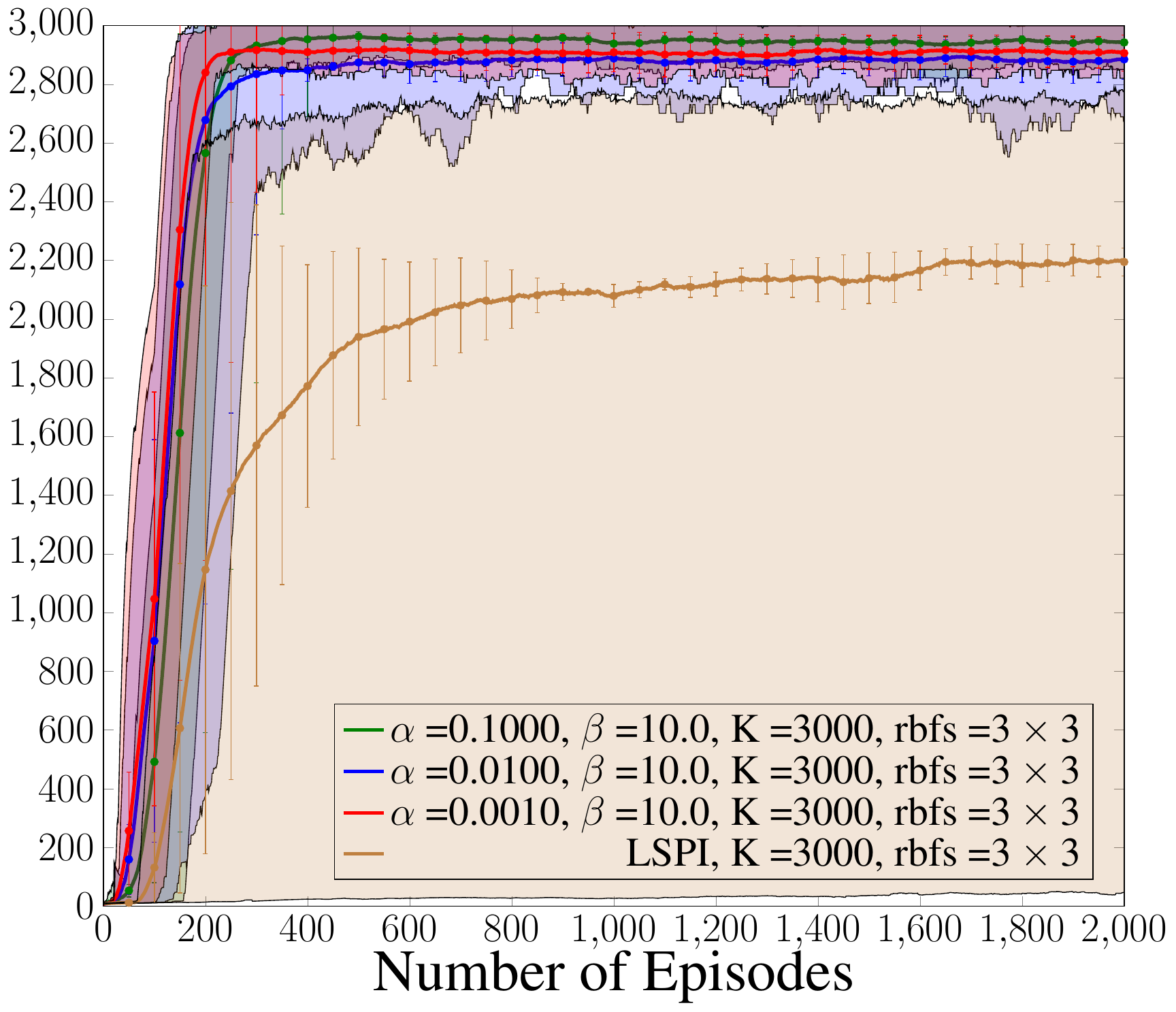}}
    \end{tabular}
    \caption{$3 \times 3$  equidistant grid of RBFs.}
    \label{fig:pendulum3}
  \end{subfigure}
  \begin{subfigure}[t]{0.497\textwidth}
    \begin{tabular}{cc}
      \scalebox{.21}{\includegraphics{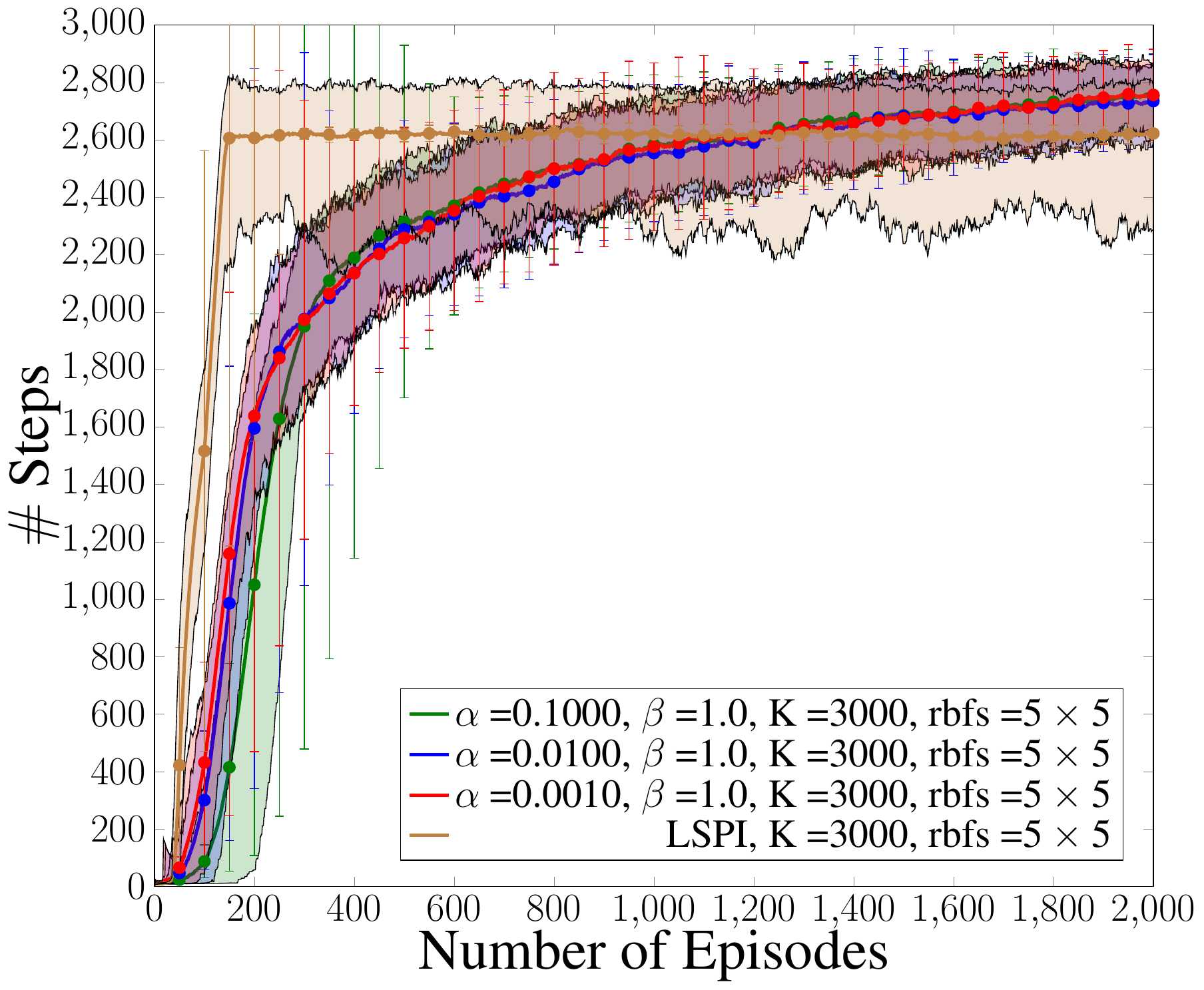}}
      \scalebox{.21}{\includegraphics{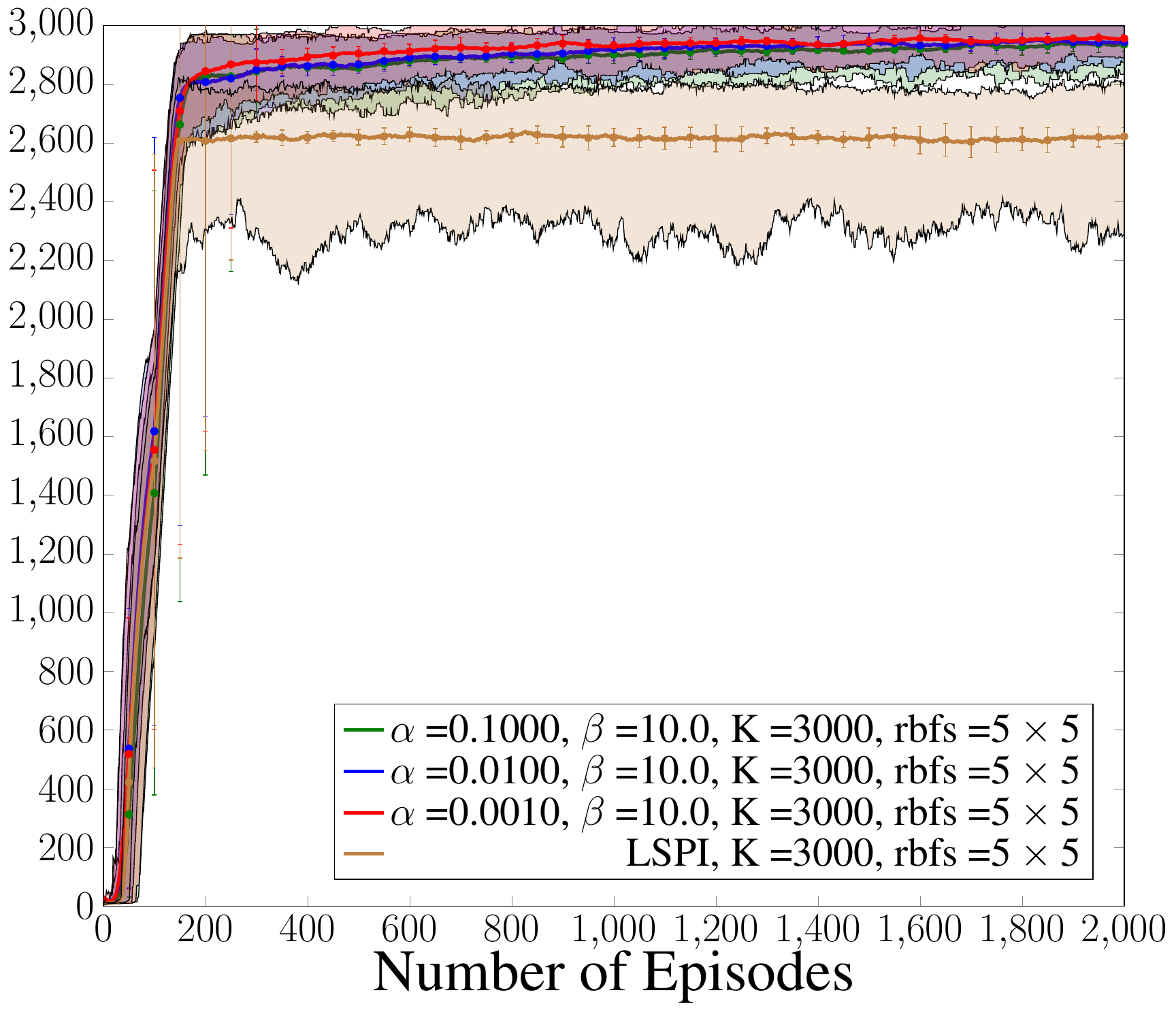}}
    \end{tabular}
    \caption{$5 \times 5$  equidistant grid of RBFs.}
    \label{fig:pendulum5}
  \end{subfigure}
  \caption{Performance of RBLSPI and online LSPI on the {\bf inverted pendulum} environment on two different grids of RBFs. The performance of the RBLSPI algorithm is also presented for varying hyperparameters $\alpha$ and $\beta$.}
  \label{fig:pendulum}
\end{figure*}

\paragraph{Inverted Pendulum \citep{lagoudakis2003least}}
In this domain our target is to keep a pendulum balanced for $3000$ steps by applying forces of a fixed magnitude ($50$ Newtons).
The state space is described by two continuous variables, the vertical angle ($\theta$) and the angular velocity ($\dot{\theta}$) of the pendulum.
There are three noisy actions where agent can apply to keep the pendulum balanced: no force, left force or right force.
A uniform noise in $[-10, 10]$ is applied to the chosen action.
A zero reward is received at each time step except in the case where the pendulum falls ($|\theta| \leq \pi/2$).
If pendulum falls ($|\theta| \geq \pi/2$) the episode ends and a penalty ($-1$) is received.
The discount factor of the process is set equal to $0.95$.
Two different grids of RBFs over the state space have been considered in our analysis, a $3 \times 3$ and a $5 \times 5$ equidistant grid.
Therefore, the total number of basis functions at each case is equal to $30$ and $78$, respectively.

Figure~\ref{fig:pendulum} illustrates the performance of the evolution of the policies discovered by RBLSPI and online LSPI algorithms in the environment of inverted pendulum.
More specifically, Figs.~\ref{fig:pendulum3} and ~\ref{fig:pendulum3} show the results of our analysis when a $3 \times 3$ and a $5 \times 5$ equidistant grid of RBFs is considered, respectively.
It seems that RBLSPI achieves to discover optimal or near-optimal policies independent of the hyperparameters setting. 
In the case where we set hyperparameter $\beta$ equal to $10.0$, our agent performs much better than in the case of $\beta = 1.0$.
Also, our analysis indicates that the agent's behavior is better by setting $\alpha$ to a high value, i.e. $\alpha = 0.1$.
It worths also to be noted that the RBLSPI outperforms the online LSPI algorithm in both cases ($3 \times 3$ grid (Fig.~\ref{fig:pendulum3})  and $5 \times 5$ grid (Fig.~\ref{fig:pendulum5})).
More specifically, the online LSPI doesn't achieve to discover the optimal policy even after $2000$ episodes.
On the contrary, the RBLSPI algorithm discover an optimal policy after around $200$ episodes if we set the precision parameter $\beta$ equal to $10.0$.

\paragraph{Cart Pole \citep{Barto83, Sutton+Barto:1998}}
Our objective in this task is to keep a pole hinged to a cart moving along a track from falling over.
The agent controls the system by applying a force of $+1$ or $-1$ to the cart. 
The state space is described by four continuous variables: cart position ($p$), cart velocity ($v$), pole angle ($\theta$) from vertical, and pole angular velocity ($\dot{\theta}$).
A failure occurs even if the pole falls ($|\theta| \geq \pi/6$) or if the cart runs out of the track ($|p| \geq 2.4$).
At each episode, the starting state of the agent is selected uniformly over $[-0.05, 0.05]^4$.
The reward is equal to $+1$ for every time step where the pole remains upright and the cart is kept inside the track.
The discount factor is $0.99$ and the maximum number of steps per episode is equal to $500$.
An equidistant $3 \times 3$ grid of RBFs over the state space plus a constant term is selected ($164$ basis functions in total).
\begin{figure}[t]
  \centering
  \scalebox{.25}{\includegraphics{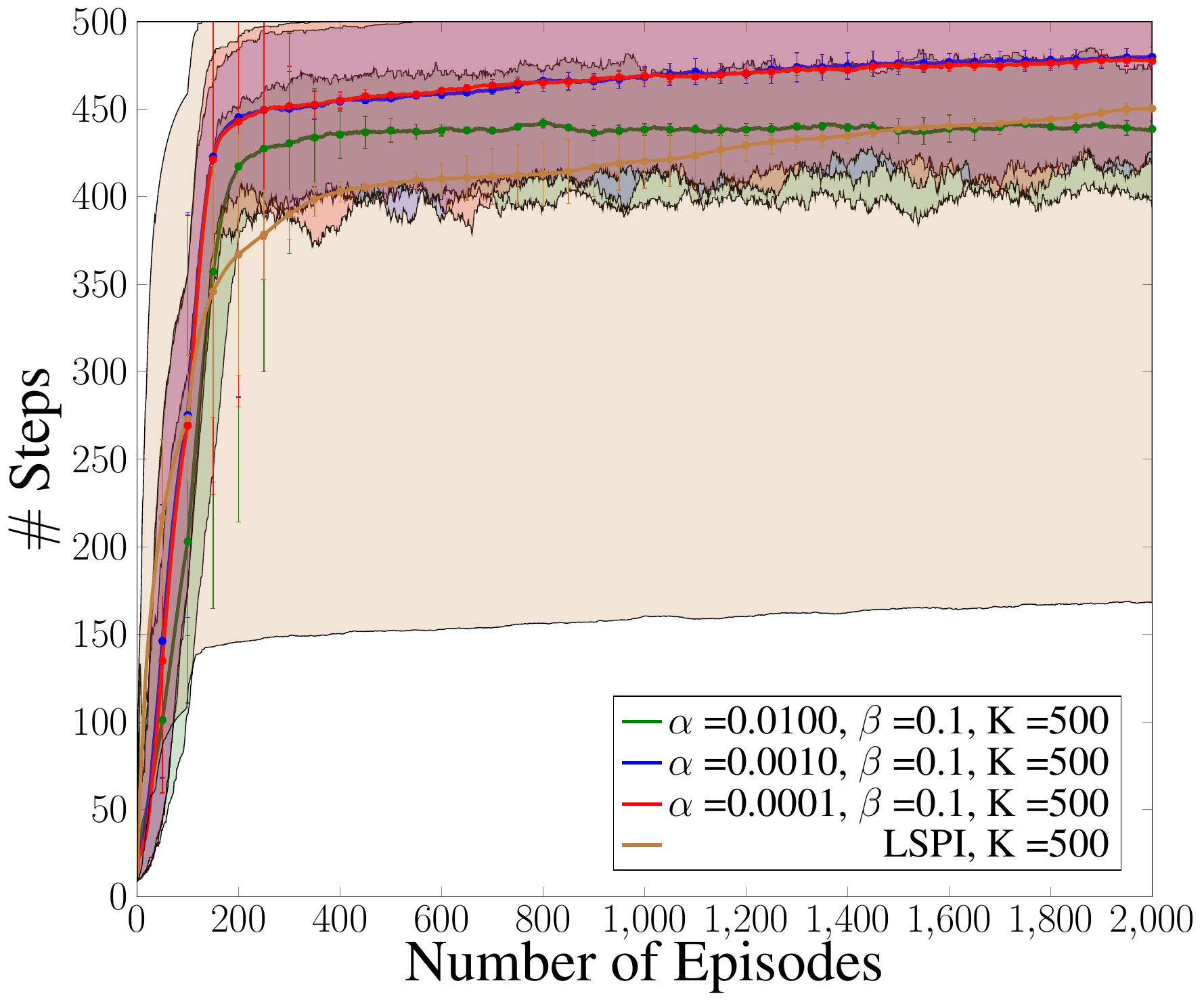}}
  \scalebox{.25}{\includegraphics{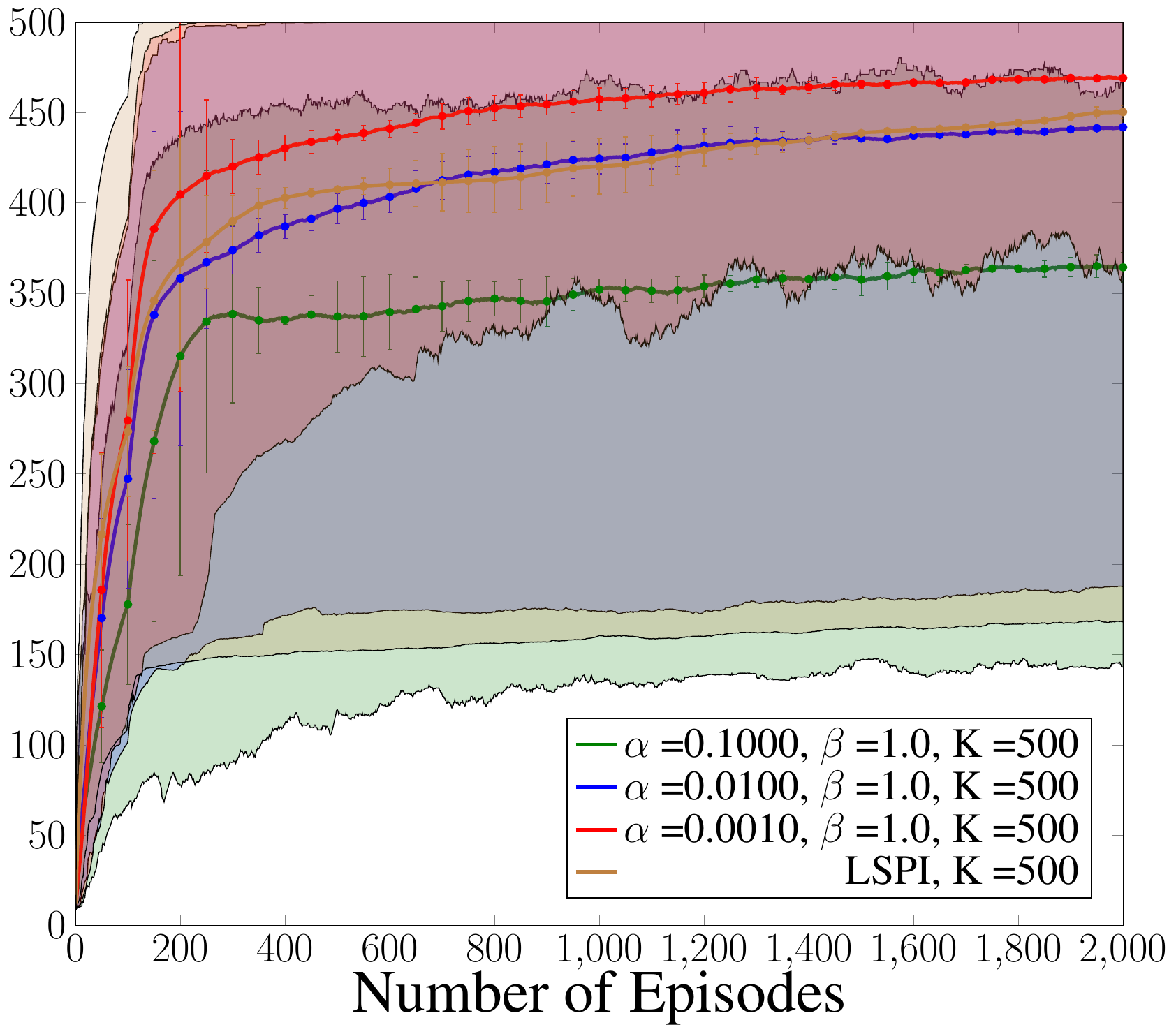}}
  \caption{Performance of RBLSPI and online LSPI on the {\bf cart pole} environment. The performance of the RBLSPI algorithm is also presented for varying hyperparameters $\alpha$ and $\beta$.}
  \label{fig:cartpole1}
\end{figure}

Figure~\ref{fig:cartpole1} illustrates the performance of the RBLSPI and the online LSPI algorithms in cart pole environment.
Actually, our experiments show that RBLSPI is able to discover near optimal policies as it is able to balance the pole for more than $480$ steps and to keep the cart on track.
Moreover, the performance of the RBLSPI is much better (especially if we set $\beta$ to $0.1$) compared to the one of the online LSPI. 
It should be also noted that the performances of both the RBLSPI and the online LSPI are highly correlated to the number of used RBFs.
In this way, we intend that the RBLSPI algorithm will be able to discover the optimal solution in the case where we increase the resolution of our feature space (i.e., by using a $5 \times 5$ equidistant grid of RBFs).

\paragraph{PuddleWorld \citep{nips:sutton:1996}}

The goal for the agent in this domain is to reach an area located at the upper right corner, avoiding two puddles.
Actually, the agent is located in a continuous $2-$dimensional terrain ($[0,1]^2$) that contains two oval puddles.
There are four discrete actions: up, down, left, and right.
By selecting an action, the agent moves by $0.05$ in the corresponding direction, up to the limits of the area.
A random white gaussian noise with $std = 0.01$ is also added to the motion along each dimension.
A negative reward ($-1$) is received at each time step plus a penalty between $0$ and $-40$ (depending on the proximity to the middle of the puddle) in a puddle is entered by the agent.
A new episode starts even if the goal is reached or after $500$ time steps.
Starting states are chosen uniformly over $[0,1]^2$ and are considered to be out of a puddle.
The discount factor is $\gamma = 0.99$.
An equidistant $8 \times 8$ grid of RBFs over the state space plus a constant term is selected ($260$ basis functions are used in total).

\begin{figure*}[t]
  \centering
  \begin{subfigure}[t]{0.497\textwidth}
    \begin{tabular}{cc}
      \scalebox{.21}{\includegraphics{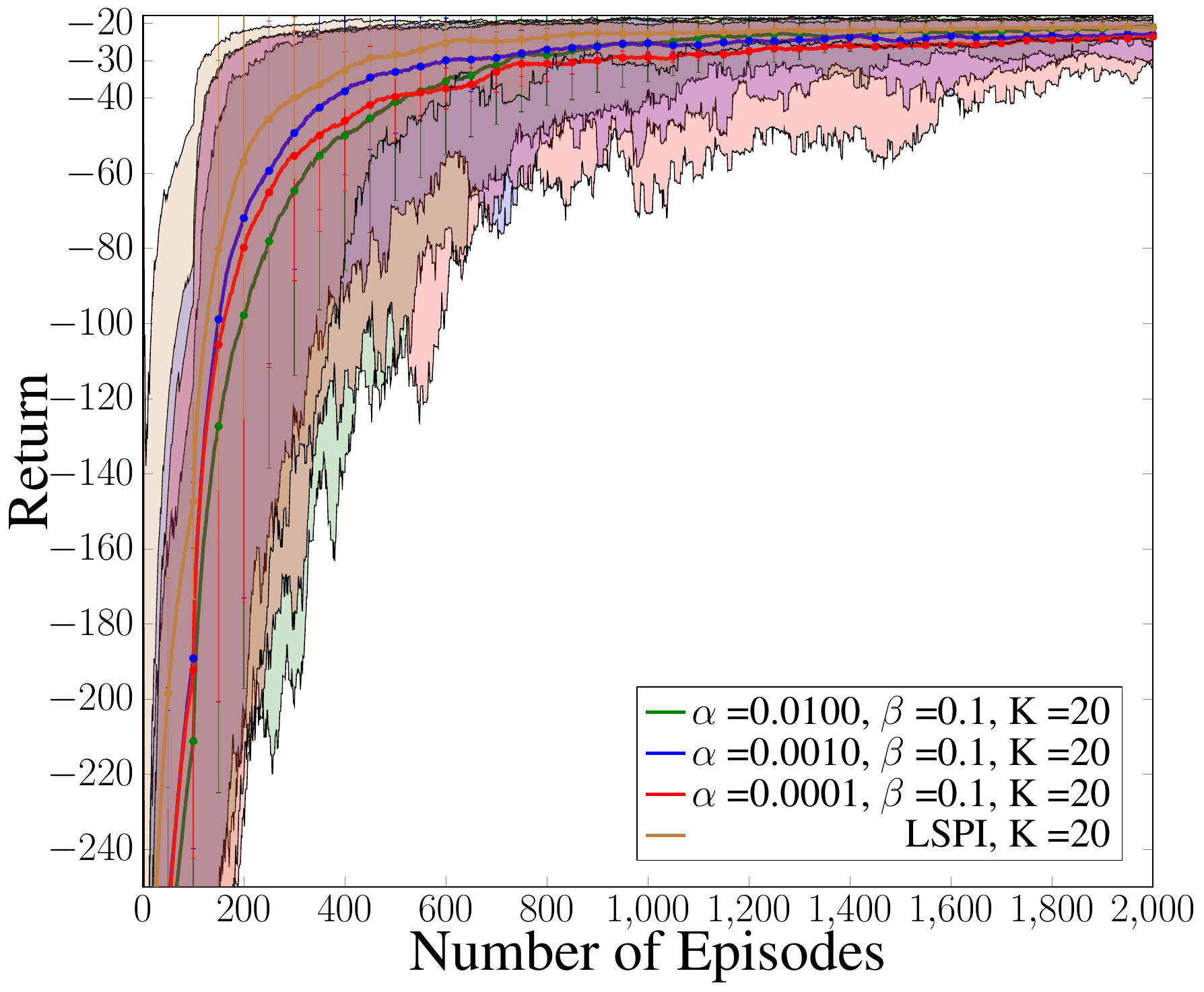}}
      \scalebox{.21}{\includegraphics{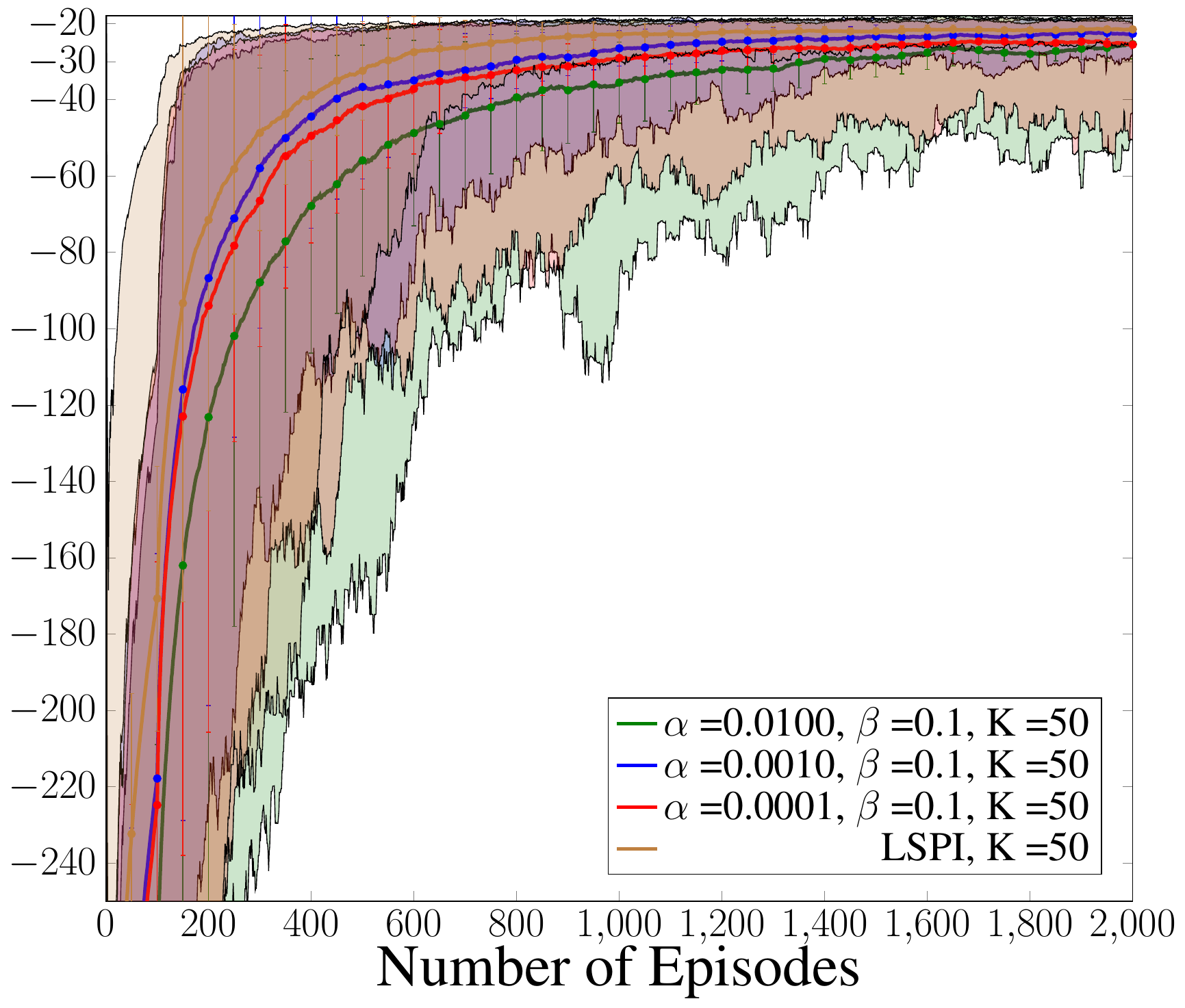}}\\
      \scalebox{.21}{\includegraphics{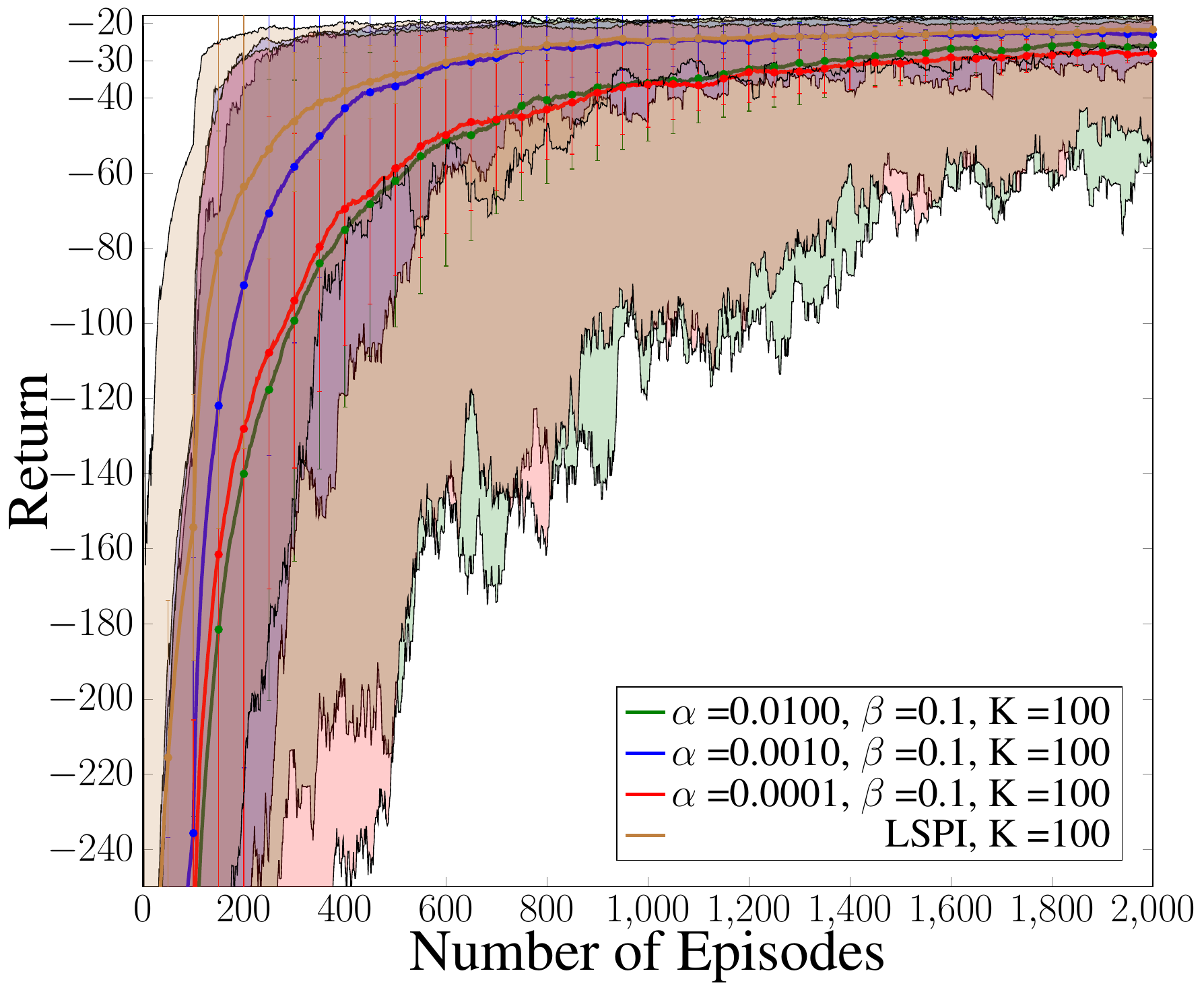}}
      \scalebox{.21}{\includegraphics{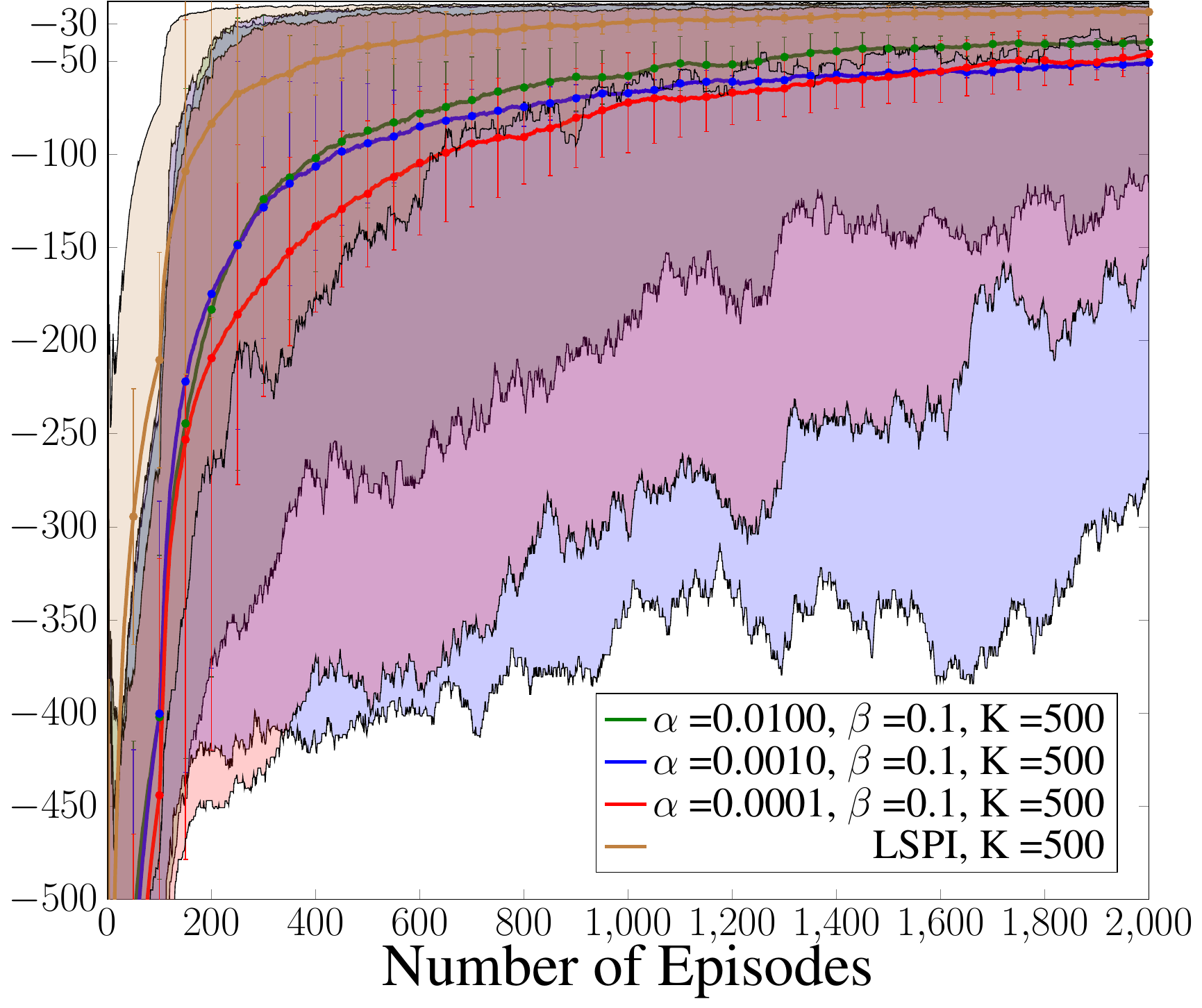}}
    \end{tabular}
    \caption{RBLSPI hyperparameter $\beta = 0.1$}
    \label{fig:puddleworld01}
  \end{subfigure}
  \begin{subfigure}[t]{.497\textwidth}
    \begin{tabular}{cc}
      \scalebox{.21}{\includegraphics{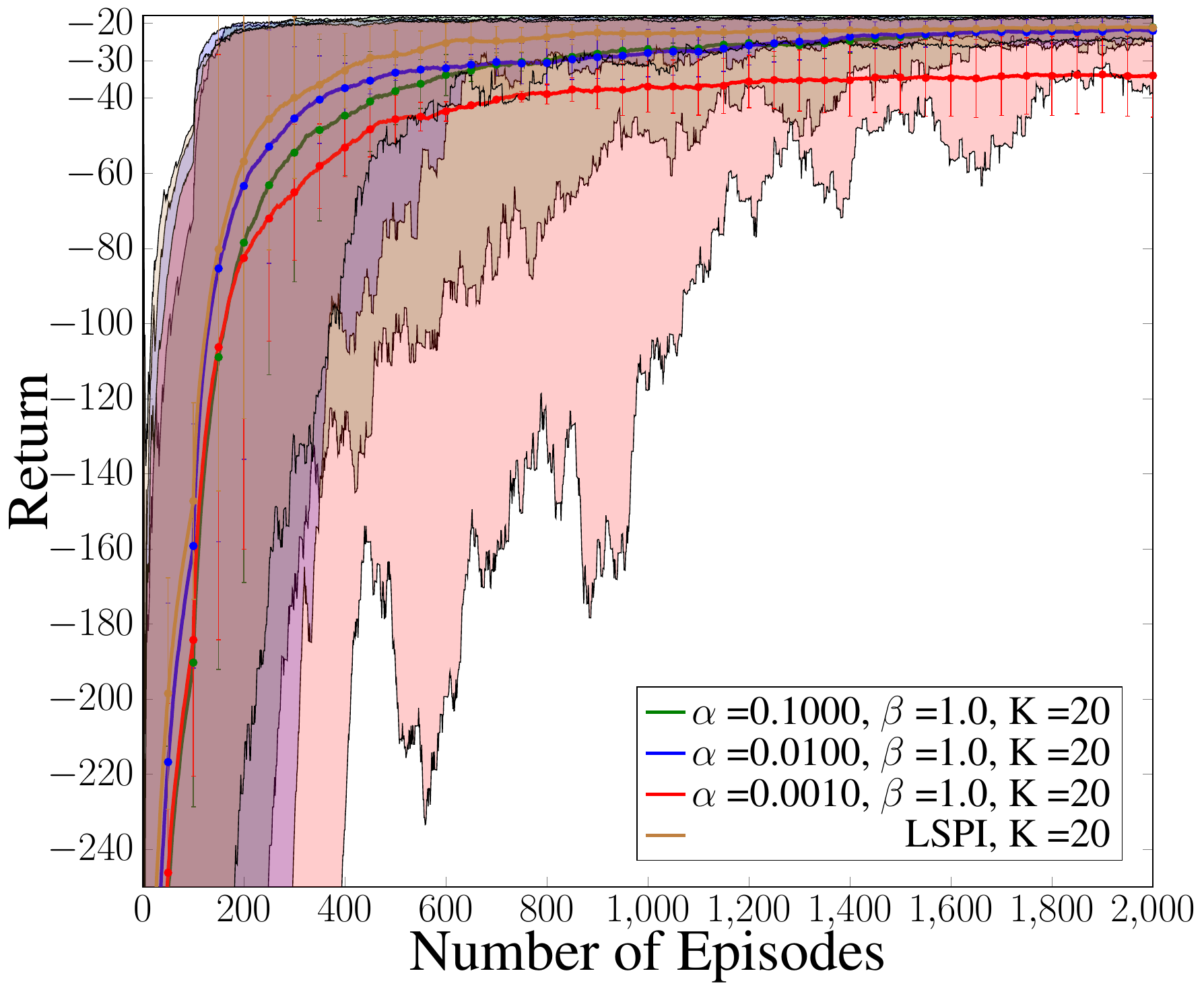}}
      \scalebox{.21}{\includegraphics{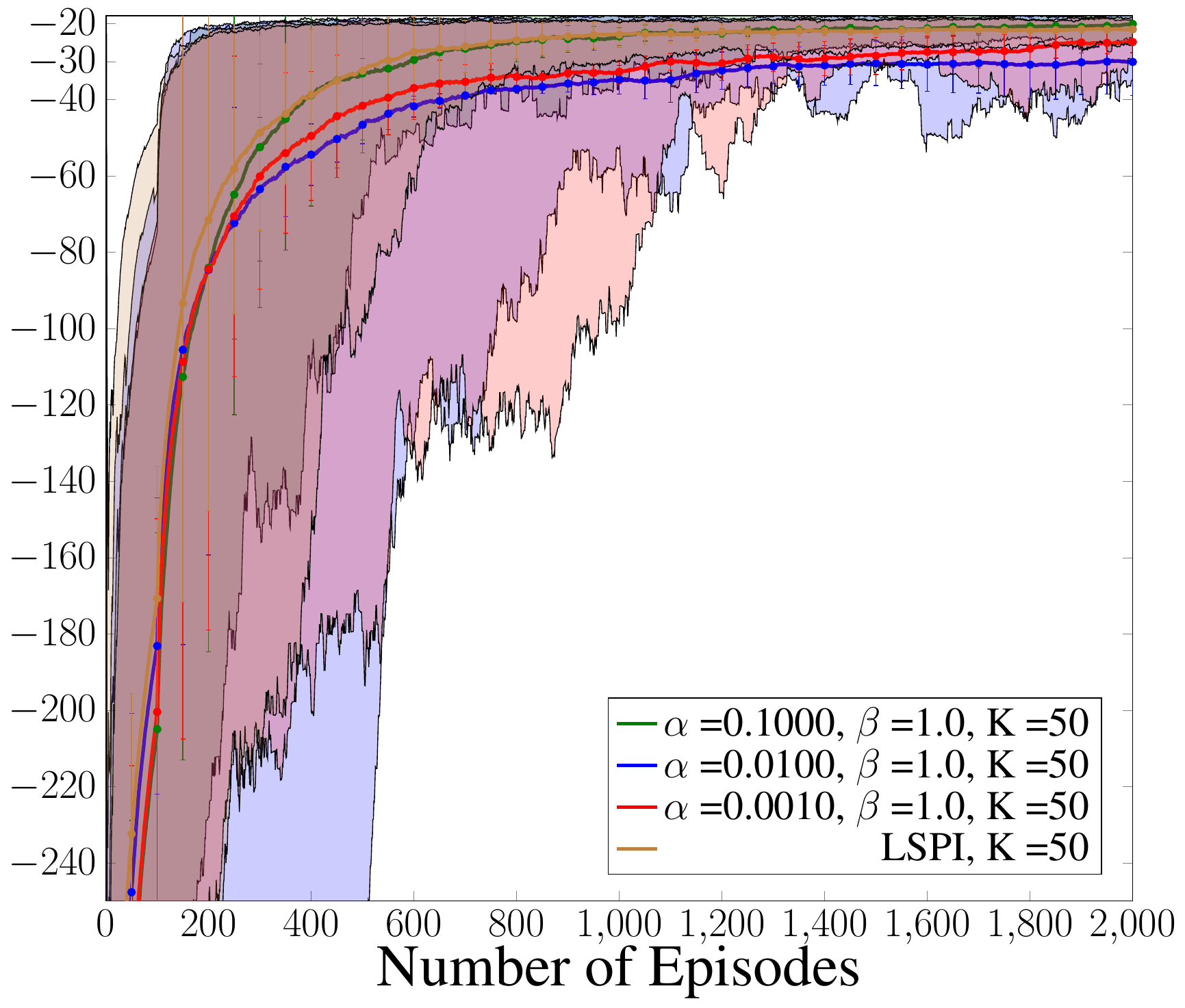}}\\
      \scalebox{.21}{\includegraphics{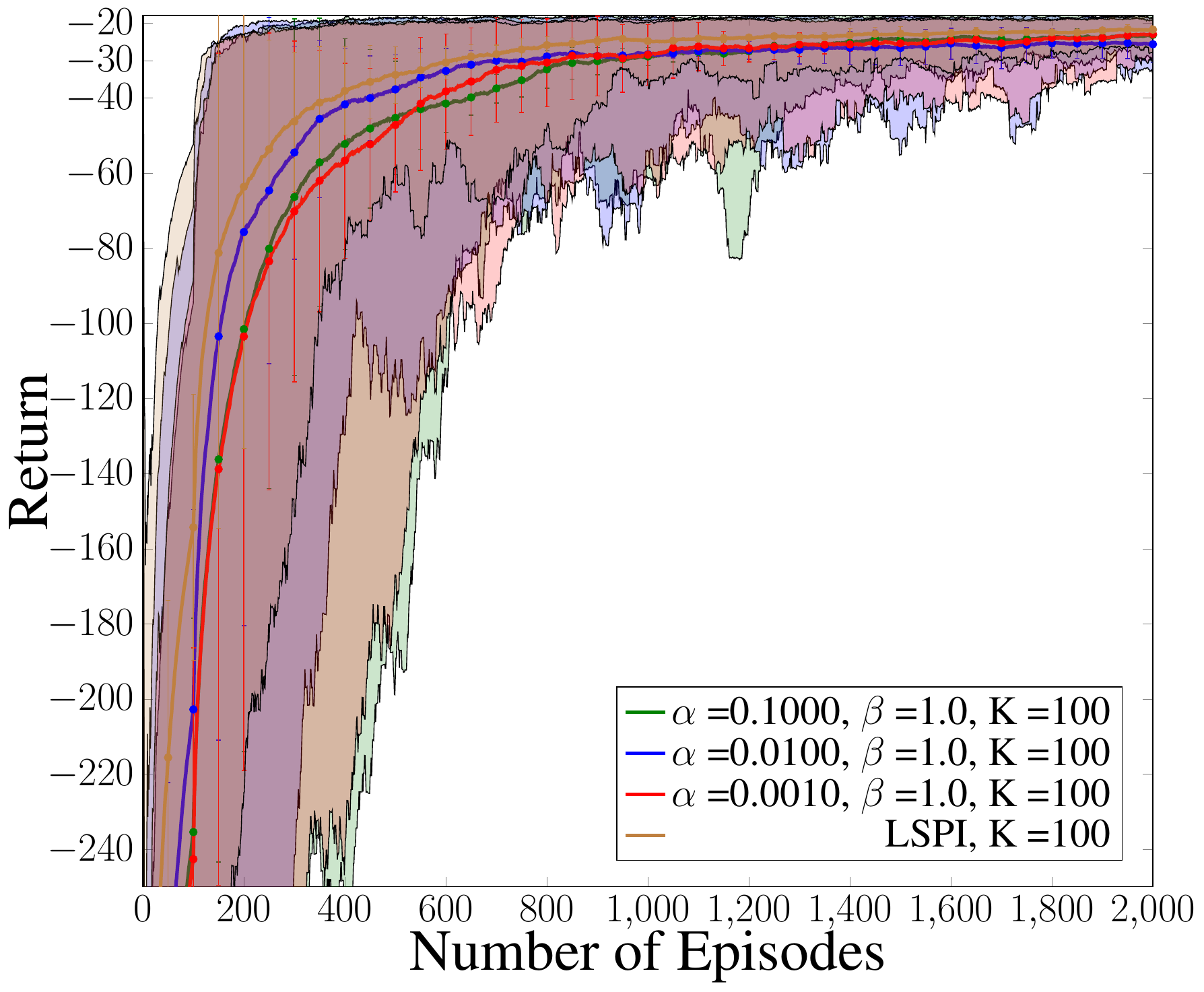}}
      \scalebox{.21}{\includegraphics{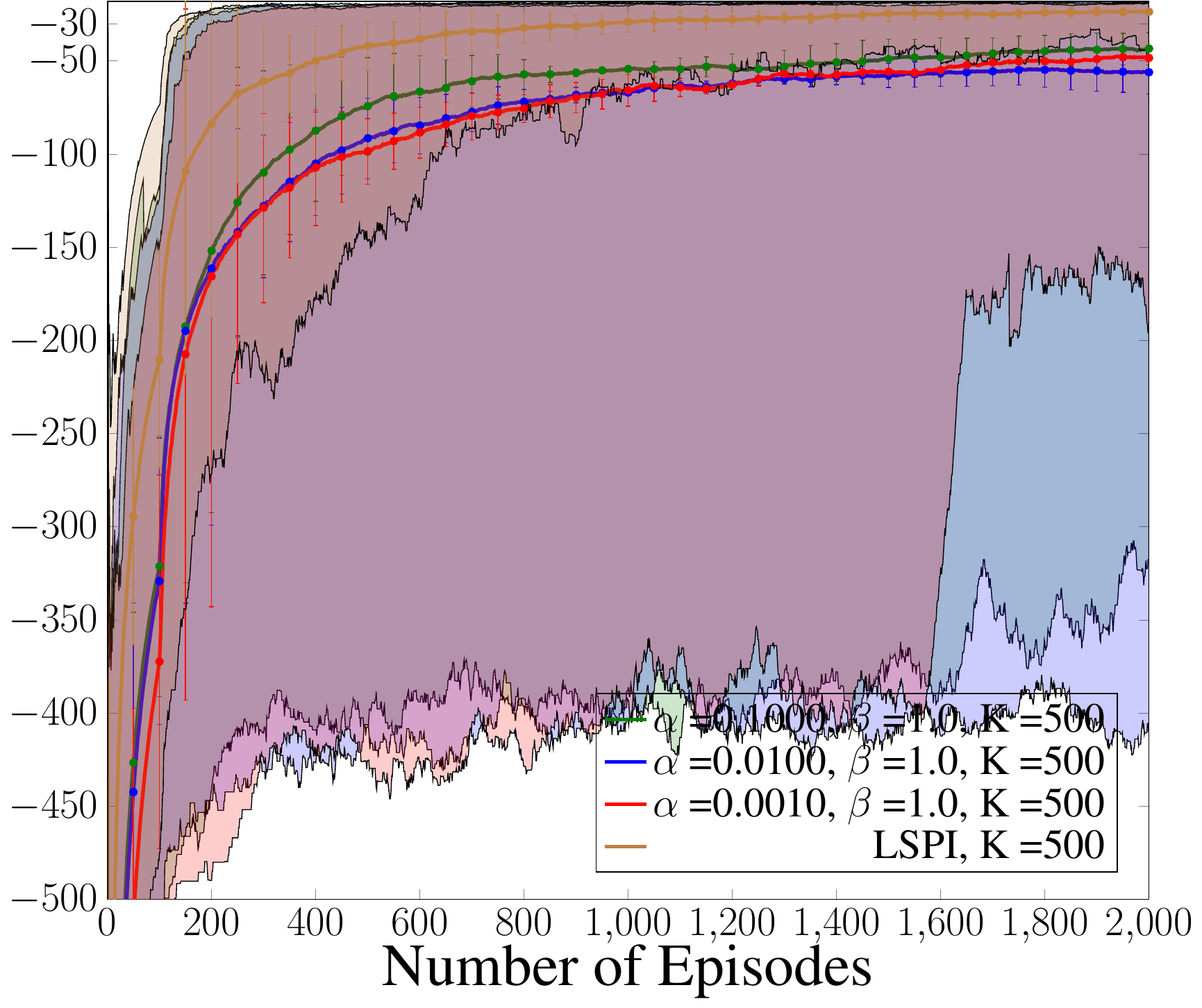}}
    \end{tabular}
    \caption{RBLSPI hyperparameter $\beta = 1.0$}
    \label{fig:puddleworld1}
  \end{subfigure}
  \caption{Performance of RBLSPI and online LSPI on the {\bf puddle world} environment for varying parameter $K$. The performance of the RBLSPI algorithm is also presented for varying hyperparameters $\alpha$ and $\beta$.}
  \label{fig:puddleworld}
\end{figure*}

In the puddle world environment, we consider the average undiscounted return received by our agent at each episode, in contrast to the previous domains where we considered  the total number of steps per episode.
Our results for puddle world are presented in Figure~\ref{fig:puddleworld}.
Actually, Figure~\ref{fig:puddleworld01} shows the performance of RBLSPI in the case where $\beta=0.1$, while Figure~\ref{fig:puddleworld1} corresponds to $\beta=1.0$.
Our results show that frequent improvement of our policies increase the performance of the proposed RBLSPI algorithm.
This happens due to the fact that after each policy improvement step, we sample a new exploration policy (followed to select actions thereafter) though the sampling of posterior distribution over model's parameters (see Eq.\ref{eq:posterior}).
In this way, we reinforce the exploration behavior of our agent.
It worths also to be mentioned that the performance of the online LSPI is quite good in contrast to its performance on the other domains that have been examined previously.
Actually, the onpolicy variant of online LSPI algorithm performs quite close or even better compared to RBLSPI algorithm.
We should also mention that the performance of the original offpolicy (online) LSPI algorithm \citep{Busoniu10OnlineLSPI} is poor and it cannot  discover a good policy even after a huge number of episodes.
The main conclusion of our empirical analysis is that deep exploration is necessary especially in the case where the reward signal is sparse or the target cannot be easily reached by the agent. 

\end{document}